\documentclass{article}

     \PassOptionsToPackage{numbers, compress}{natbib}

\usepackage[preprint]{neurips_2022}

\usepackage[utf8]{inputenc} 
\usepackage[T1]{fontenc}    
\usepackage{hyperref}       
\usepackage{url}            
\usepackage{booktabs}       
\usepackage{amsfonts}       
\usepackage{nicefrac}       
\usepackage{microtype}      
\usepackage{xcolor}         
\usepackage{enumitem}
\usepackage{amsmath}
\usepackage{algorithmic}
\usepackage{algorithm}
\usepackage{graphicx}
\usepackage{subfig}
\usepackage{color}
\usepackage{multirow}
\usepackage{array}
\usepackage{placeins}
\usepackage{textcase}
\usepackage[font=small]{caption}
\usepackage{wrapfig}

\usepackage{paralist}
\setdefaultleftmargin{2em}{}{}{}{.5em}{.5em}
\usepackage{amssymb}
\usepackage{graphicx}
\usepackage{arydshln}

\newcommand{\D}{{\cal D}}
\newcommand{\iid}{\emph{i.i.d.}}
\newcommand{\s}[1]{^{(#1)}}

\newcommand{\mahaldist}[2]{||#1||^2_{#2}}

\newcommand{\abc}[1]{\textcolor{black}{#1}}
\newcommand{\abcn}[1]{\textcolor{black}{#1}}
\newcommand{\zxy}[1]{\textcolor{black}{#1}}
\newcommand{\dzy}[1]{\textcolor{black}{#1}}
\newcommand{\CUT}[1]{}

\title{Pareto Optimization for Active Learning under Out-of-Distribution Data Scenarios}

\author{%
  Xueying Zhan\footnotemark[1] \\
  City University of Hong Kong\\
  \texttt{xyzhan2-c@my.cityu.edu.hk} \\
  \And
  Zeyu Dai\footnotemark[1] \\
  The Hong Kong Polytechnic University \\
  \texttt{ze-yu.dai@connect.polyu.hk}\\
  \And
  Qingzhong Wang \\
  Baidu Research \\
  \texttt{wangqingzhong@baidu.com} \\
  \And
  Qing Li \\
  The Hong Kong Polytechnic University \\
  \texttt{csqli@comp.polyu.edu.hk} \\
  \And
  Haoyi Xiong \\
  Baidu Research \\
  \texttt{xionghaoyi@baidu.com}\\
  \And
  Dejing Dou \\
  Baidu Research \\
  \texttt{doudejing@baidu.com}\\
  \And
  Antoni B.~Chan \\
  City University of Hong Kong\\
  \texttt{abchan@cityu.edu.hk}
}

\begin{document}
\renewcommand*{\thefootnote}{\fnsymbol{footnote}}
\footnotetext[1]{Work is completed while Xueying Zhan is at Baidu Research. The first two authors are equally contributed.} 
\setcounter{footnote}{0} 
\renewcommand*{\thefootnote}{\arabic{footnote}}

\maketitle

\begin{abstract}
Pool-based Active Learning (AL) has achieved great success in minimizing labeling cost by sequentially selecting 
informative unlabeled samples from a large unlabeled data pool and querying their labels from oracle/annotators. However, existing AL sampling strategies might not work well in out-of-distribution (OOD) data scenarios, where the unlabeled data pool contains some data samples that do not belong to the classes of the target task.
Achieving good AL performance under OOD data scenarios is a challenging task due to the natural conflict between AL sampling strategies and OOD sample detection -- AL 
selects data that are hard to be classified by the current basic classifier \abc{(e.g., samples whose predicted class probabilities have high entropy)}, while OOD samples tend to have \abc{more uniform predicted class probabilities (i.e., high entropy)} than in-distribution (ID) data. In this paper, we propose a sampling scheme, Monte-Carlo Pareto Optimization for Active Learning (\textbf{POAL}), which selects optimal subsets of unlabeled samples with fixed batch size from the unlabeled data pool. \abc{We cast the AL sampling task as a multi-objective optimization problem, and thus} we utilize Pareto optimization based on two conflicting objectives: (1) the normal AL data sampling scheme (e.g., maximum entropy), and  (2) 
the confidence of not being an OOD sample.
Experimental results show its effectiveness on both classical Machine Learning (ML) and Deep Learning (DL) tasks.
\end{abstract}

\section{Introduction}
\label{intro}
In real-life applications, large amounts of 
unlabeled data are easy to be obtained, but labeling them is expensive and time-consuming \citep{shen2004multi}. AL is an effective way to solve this problem -- it achieves greater accuracy with less training data by selecting the most informative or representative instances from which it learns, and then querying their labels  
from oracles/annotators 
\citep{zhan2021comp}. Current AL sampling strategies have been tested in relatively simple and clean well-studied datasets \citep{kothawade2021similar} like \emph{MNIST} \citep{deng2012mnist} and \emph{CIFAR10} \citep{krizhevsky2009learning} datasets. 
However, in realistic scenarios, when collecting unlabeled data samples, unrelated data samples (i.e., out-of-domain data) might be mixed in with the task-related data, e.g., images of letters when the task is to classify images of digits \citep{du2021contrastive}.
%
%
Most AL methods are not robust to OOD data scenarios. For instance, \citet{karamcheti2021mind} has demonstrated \zxy{empirically}
that collective outliers hurt AL performances under Visual Question Answering (VQA) tasks. 
\abc{Meanwhile, selecting and querying OOD samples that are invalid for the target model will waste the labeling cost \citep{du2021contrastive}, and make the AL sampling process less effective.}

There is a natural conflict between uncertainty-based AL sampling strategy and OOD data detection -- most AL sampling schemes, especially uncertainty-based measures, prefer selecting data that are hardest to be classified by the current basic learner/classifier, 
but these data might also 
\begin{wrapfigure}{r}{4cm}
\centering
\includegraphics[width=0.3\textwidth]{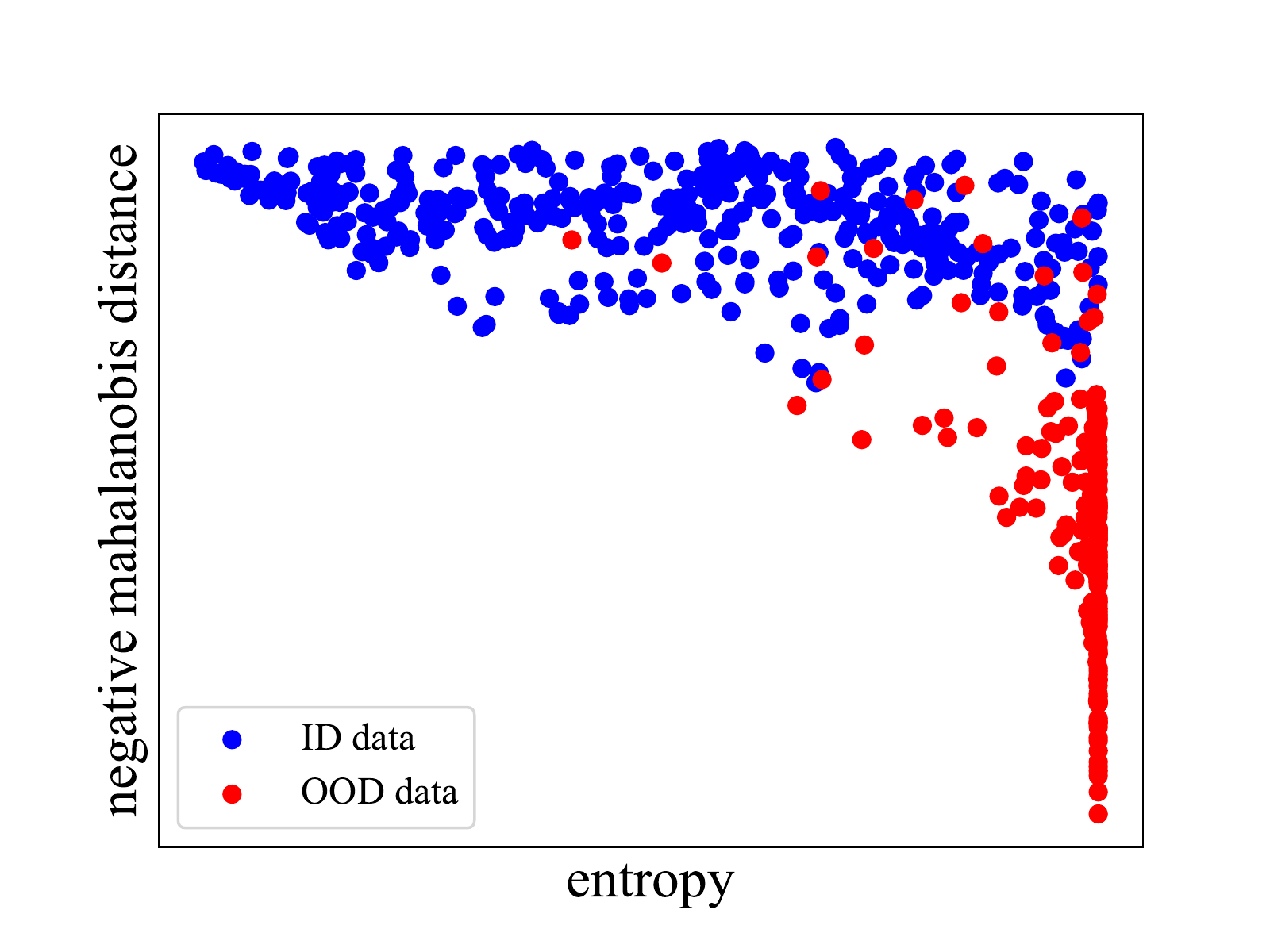}
\caption{Scatter plot of the AL score (entropy) and ID confidence score (negative Mahalanobis distance) of unlabeled data from \emph{EX8} dataset. \abcn{AL will select high entropy data first, which includes most of the OOD data. Larger ID score indicates the data is more likely to be ID.}
}
\label{relation}
\end{wrapfigure}
be OOD samples. For example, a typical uncertainty-based AL method is Maximum Entropy (\textbf{ENT}) \citep{lewis1994heterogeneous, shannon2001mathematical}, which selects data 
whose predicted class probabilities have largest entropy.
However, \textbf{ENT} is also a typical OOD detection method -- high entropy of the predicted class distribution (high predictive uncertainty) suggests that the input may be OOD \citep{ren2019likelihood}. Therefore, when naively applying uncertainty-based AL in OOD data scenarios, OOD samples are more likely to be selected for labeling than ID data.
\abcn{Fig.~\ref{relation} shows an example on the \emph{EX8} dataset \citep{andrew2008stan}, which illustrates this conflict.}
%
%
Moreover, almost all of the OOD data has 
high \textbf{ENT}, \abc{and thus the AL algorithm is more likely to select OOD data for labeling, thus wasting the labeling budget}. 

Although the OOD problem has been demonstrated to affect AL in real-life applications \citep{karamcheti2021mind}, there are only a few studies on it \citep{kothawade2021similar,du2021contrastive}. 
In \citep{kothawade2021similar}, OOD is only considered as a sub-task \zxy{and its sub-modular mutual information based sampling scheme is both time and memory consuming.} 
\citep{du2021contrastive} needs to pre-train extra self-supervised models like SimCLR \citep{chen2020simple}, and also introduces hyper-parameters to trade-off between semantic and distinctive scores, whose values 
affect the final performance (see Section 4.3 in \citep{du2021contrastive}). These two factors limit the range of its application. \zxy{We compared our work with \citep{du2021contrastive}, \citep{chen2020simple} in detail in Appendix A.4.1.}

In this paper, to address the above issues, we advocate simultaneously considering both the AL criterion and ID confidence when designing AL sampling strategies. Since these two objectives are conflict, we define the AL sampling process under OOD data scenarios 
as a multi-objective optimization problem \citep{seferlis2004integration}. Unlike traditional methods for handling multiple-criteria based AL, such as weighed-sum optimization \citep{zhan2022comparative} or \abcn{two-stage optimization} (first select a subset based on one objective then make final decisions based on the other objective) \citep{shen2004multi}, we propose a novel and flexible batch-mode \textbf{P}areto \textbf{O}ptimization \textbf{A}ctive \textbf{L}earning (\textbf{POAL}) framework with fixed batch size. The contributions of this paper are 
as follows:

\begin{compactenum} 
\item 
We propose AL under OOD data scenarios within the framework of a multi-objective optimization problem. 

\item Our framework is flexible and can accommodate 
different combinations of AL methods and OOD detection methods according to various target tasks. In this paper, we use \textbf{ENT} as the AL objective and Mahalanobis distance based ID confidence scores.
\item \abc{Naively applying Pareto optimization to AL will result in a Pareto Front with non-fixed size, which can introduce large computational cost.}
To enable efficient Pareto optimization, we propose a \emph{Monte-Carlo (MC) Pareto optimization algorithm for fixed-size batch-mode AL}.
\item Our framework works well on both classical ML and DL tasks, and we propose pre-selecting and early-stopping techniques to reduce the computational cost on large-scale datasets. 

\item Our framework has no trade-off hyper-parameters for balancing the AL and OOD objectives.
This is crucial since \zxy{i) AL is data-insufficient, and there might be no extra validation set for tuning hyper-parameters; or ii) hyper-parameter tuning in AL can be label expensive, since every change of the hyper-parameters causes AL to label new examples, 
%
%
thus provoking substantial labeling inefficiency \citep{ash2020deep}.}
\end{compactenum}

\section{Related Work}
\paragraph{Pool-based Active Learning.}
Pool-based AL has been well-studied in previous years \citep{settles2009active, zhan2021comp, ren2021survey, zhan2022comparative} and widely adopted in various tasks like natural language processing \citep{dor2020active}, semantic parsing \citep{duong2018active}, object detection \citep{haussmann2020scalable}, image classification/segmentation \citep{gal2017deep, yoo2019learning}, etc. Most pool-based AL sampling schemes rely on fixed heuristic sampling strategies, 
which follow two main branches:
uncertainty-based measures and representative/diversity-based measures \citep{ren2021survey, zhan2022comparative}. Uncertainty-based approaches aim to select data that maximally reduce the uncertainty of target basic learner 
\citep{ash2020deep}. Typical uncertainty-based measures that perform well on classical ML tasks (e.g.,  Query-by-Committee \citep{seung1992query}, \textbf{ENT} \citep{lewis1994heterogeneous}, Bayesian Active Learning by Disagreement \textbf{BALD} \citep{houlsby2011bayesian})
have also been generalized to DL tasks \citep{wang2014new, gal2017deep, beluch2018power, zhan2022comparative}.
Representative/diversity-based methods select batches of unlabeled data that are most representative of the unlabeled set. Typical methods include Hierarchical sampling \citep{dasgupta2008hierarchical}, $k$-Means, and Core-Set approach \citep{sener2018active}. However, 
these methods measure representativeness/diversity using pair-wise distances/similarities calculated across the whole data pool, which is computationally expensive. 
%
\zxy{Both of these two branches of AL methods might not work well under OOD data scenarios. As discussed in Section~\ref{intro}, uncertainty-based measure like \textbf{ENT} are not robust to OOD samples, since OOD samples are naturally difficult to be classified. Representative/diversity-based methods are also not robust to OOD scenarios, since outliers/OOD samples that are far from other data are more likely to be selected first as representative data. To address these issues, we proposed our \textbf{POAL} framework, which can be adopted to various AL sampling scheme and OOD data scenarios.}
\vspace{-0.3cm}
\paragraph{Out-of-Distribution.} 
Detecting OOD data is of vital importance in ensuring the reliability and safety of ML systems in real-life applications \citep{yang2021generalized}, since the OOD problem severely influences real-world predictions/decisions. For example, in medical diagnosis, the trained classifier could wrongly classify  a healthy \abc{OOD} sample as pathogenic \citep{ren2019likelihood}. Existing methods compute ID  confidence scores based on 
the predictions of (ensembles of) classifiers trained on ID data, e.g., the ID confidence can be the entropy of the predictive class distribution \citep{ren2019likelihood}. \citet{hendrycks2017baseline} observed that a well-trained neural network assigns higher softmax scores to ID data than OOD data.  Follow-up work \textbf{ODIN} \citep{liang2017enhancing} amplifies the effectiveness of OOD detection with softmax score 
by considering temperature scaling and input pre-processing.
Others \citep{lee2018simple,cui2020accelerating} proposed a simple and effective method for detecting both OOD and adversarial data samples based on Mahalanobis distance. 
Due to its superior performance compared with other OOD strategies (see Table 1 in \citep{ren2019likelihood}), 
our \textbf{POAL} framework uses Mahalanobis distance as the basic criterion for calculating the ID confidence score.
\abc{Nonetheless, our framework is general and any ID confidence score could be adopted.}
\vspace{-0.3cm}
\paragraph{\abc{Multi-objective} optimization.} Many real-life applications require optimizing multiple objectives that are conflicting with each other. 
For example, in the sensor placement problem, the goal is to maximize sensor coverage while minimizing \zxy{deployment} 
costs \citep{watson2004multiple}. Since there is no single solution that simultaneously optimizes each objective, Pareto optimization can be used to find a set of ``\emph{Pareto optimal}'' solutions with optimal trade-offs of the objectives \citep{miettinen2012nonlinear}. A decision maker can then select a final solution based on their requirements. Compared with traditional optimization methods for multiple objectives (e.g., weighted-sum \citep{marler2010weighted}), Pareto optimization algorithms are designed via different meta-heuristics \emph{without any trade-off parameters} \citep{zhou2011multiobjective,liu2021hydratext}.

Besides optimization of the two conflicting objectives (AL score and ID confidence), we also need to perform batch-mode subset selection.
In previous work on subset selection, 
Pareto Optimization Subset Selection (\textbf{POSS}) \citep{qian2015subset} solves the subset selection problem by optimizing two objectives simultaneously: maximizing the criterion function, and minimizing the subset size. However, \textbf{POSS} does not support \emph{more than one} criterion and cannot work with fixed subset size, \abc{and thus is not suitable for our task.}
Therefore, we propose Monte-Carlo \textbf{POAL}, which achieves: i) optimization of multiple objectives; 2) no extra trade-off parameters for tuning; iii) subset selection with fixed-size.

\vspace{-0.3cm}
\section{Methodology}
\vspace{-0.3cm}
In this section, we introduce our definition of pool-based AL under OOD data scenarios, the overview of \textbf{POAL}, and its detailed implementations.
\subsection{Problem Definition}
We consider a general pool-based AL process for a $K$-class classification task with feature space $\mathcal{X}$, label space $\mathcal{Y} \in \{1,...,K\}$ under OOD data scenarios.
We assume that the oracle can provide a fixed number of labels, 
and when queried with an OOD sample the oracle will return an ``OOD'' label\footnote{In our implementation, the OOD label is -1.} to represent data outside of the specified classification task.
Let $B$ be the number of labels that the oracle can provide, i.e., the budget. Our goal is to select the most informative $B$ instances with fewer OOD samples so as to obtain the highest classification accuracy.  To reduce the computational cost, we consider batch-mode AL where 
batches of samples with fixed size $b$ are selected and queried.
We denote the AL acquisition function as $\alpha(\mathbf{x}; \mathcal{A})$, where $\mathcal{A}$ refers to AL sampling strategy, and the basic learner/classifier as $f(\theta)$.
We denote the current labeled set as
$\D_l=\{(\mathbf{x}_i,y_i)\}_{i=1}^{N}$
and the large unlabeled data pool as $\D_u=\{\mathbf{x}_i\}_{i=1}^{M}$.
The labeled data are sampled $\iid$ over data space $\D$, i.e., $\D_l \in \D$, and $N \ll M$. Under OOD data scenarios, active learners may query OOD samples whose labels are not in $\mathcal{Y}$. 
To simplify the problem settings, only ID samples are added to $\D_l$.
\subsection{Pareto Optimization for Active Learning}
\begin{figure}
\centering
\includegraphics[width=1.0\textwidth]{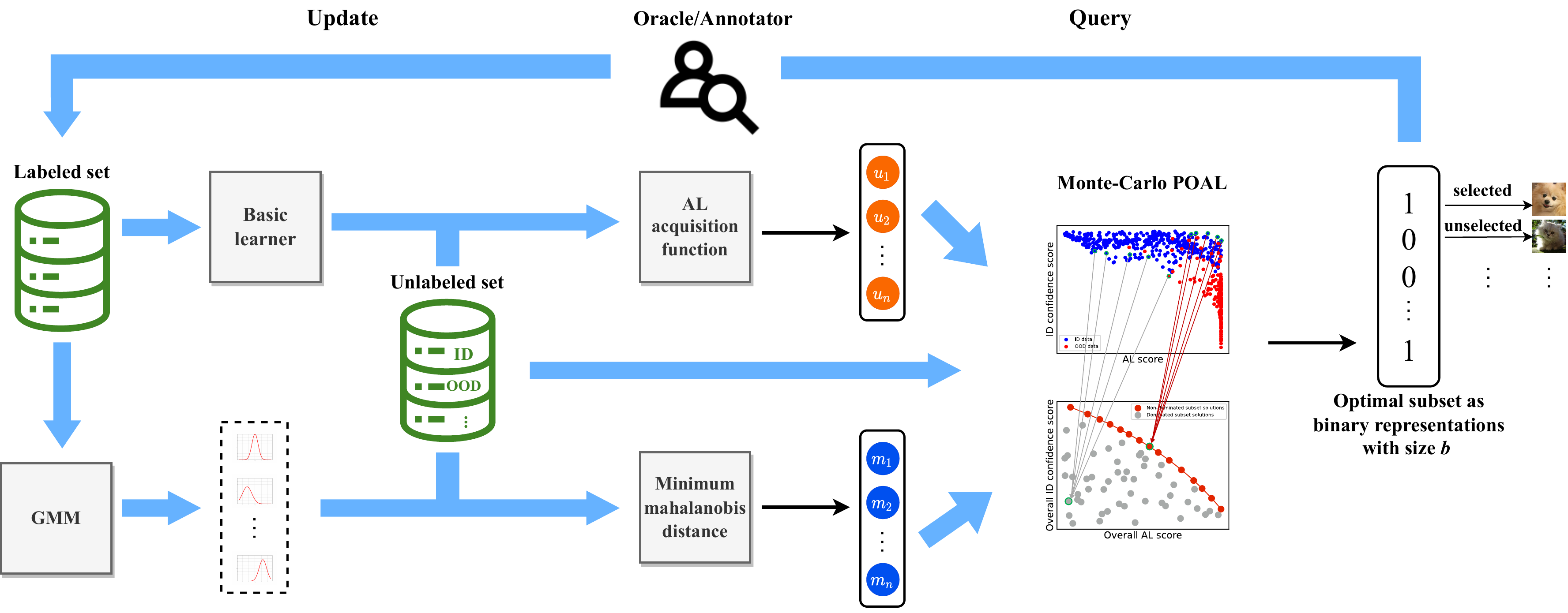}
\caption{Our proposed Monte-Carlo \textbf{POAL} framework. \zxy{In each AL iteration, we train a basic learner and estimate \textbf{GMM}s (in DL tasks, we estimate multivariate Gaussian distributions with low- and high- feature levels) to calculate the AL score and the Mahalanobis distance-based ID confidence score for each unlabeled sample.
We then construct Monte Carlo \textbf{POAL} by randomly generating candidate solutions with fixed batch size $b$ and updating the Pareto set iteratively to get an optimal subset solution.} 
}
\label{overview}
\end{figure}

In this section we introduce our proposed 
\textbf{POAL} framework as outlined in Figure~\ref{overview}. In each AL iteration, we firstly utilize the clean labeled set that only contains ID data for training a basic classifier $\theta$ and constructing class-conditional Gaussian Mixture Models (\textbf{GMM}s) for detecting OOD samples in $\D_u$. Based on classifier $\theta$, AL acquisition function $\alpha(\mathbf{x};\mathcal{A})$ and \textbf{GMM}s, we calculate the informativeness/uncertainty score \abc{$\mathcal{U}(\mathbf{x}_i)$} and
ID confidence score 
\abc{$\mathcal{M}(\mathbf{x}_i)$} for each unlabeled sample $\mathbf{x}_i$ in $\D_u$. 
We then apply our proposed \emph{Pareto optimization algorithm for fixed-size subset selection}, which outputs a batch of $b$ unlabeled samples with both higher informativeness/uncertainty score and higher ID confidence score. 
We next introduce each component of our \textbf{POAL} framework.

\subsection{AL sampling strategy}
\label{al_sec}
Typical pool-based AL sampling methods 
select data by maximizing the acquisition function $a(\mathbf{x};\mathcal{A})$, i.e., $\mathbf{x}^* = \arg\max_{\mathbf{x} \in \mathcal{D}_u} a(\mathbf{x};\mathcal{A})$. 
Different from the typical approach, in our work we use the AL scores for the whole unlabeled pool $\mathcal{D}_u$ for the subsequent Pareto optimization.
Thus, we convert the acquisition function to a querying density function via $\mathcal{U}(\mathbf{x}_i) = \tfrac{\alpha(\mathbf{x}_i;\mathcal{A}) }{\sum\nolimits_{j=1}^{M} \alpha(\mathbf{x}_j; \mathcal{A}) }$. 
In this paper we use entropy (\textbf{ENT}) as our basic AL sampling strategy. 
\textbf{ENT} selects data samples whose predicted class probabilities have largest entropy, and its acquisition function is 
$\alpha_{\textbf{ENT}}(\mathbf{x}) = - \sum \nolimits_{k=1}^{K}p_f(y=k|\mathbf{x};\theta)\log p_f(y=k|\mathbf{x};\theta))$, where $p_f(y|\mathbf{x}; \theta)$ is the posterior class probability using classifier $f(\theta)$.

Our framework is flexible due to its capability to easily incorporate other AL sampling strategies. A basic requirement is that its acquisition function  can be easily converted to a querying density function. 
In general, AL methods that explicitly provide per-sample scores (e.g., class prediction uncertainty) 
or inherently provide pair-wise rankings among the unlabeled pool (e.g., \zxy{\textbf{$k$-Means}}) 
can be easily converted to a querying density. 
Thus, many uncertainty-based measure like Query-by-Committee \citep{seung1992query}, Bayesian Active Learning by Disagreements (\textbf{BALD}) \citep{houlsby2011bayesian, gal2017deep}, and Loss Prediction Loss (\textbf{LPL}) \citep{yoo2019learning} are applicable since they explicitly provide uncertainty information per sample.  Also, $k$-Means and Core-Set \citep{sener2018active} approaches  
are applicable since they provide pair-wise similarity information for ranking. 
On the other hand, AL approaches that only provide overall scores for the candidate subsets, such as 
Determinantal Point Processes \citep{biyik2019batch, zhan2021multiple},
are unable to be used in our framework.


\subsection{ID confidence score via Mahalanobis distance}
\label{maha_sec}
The motivation of employing Mahalanobis distance for the ID  confidence score in AL is from \citep{lee2018simple}. Intuitively,  ID unlabeled data can be distinguished from OOD unlabeled data since the ID unlabeled data should be closer to the ID labeled data ($\D_l$) in feature space. One possible solution is to calculate the minimum distance between an unlabeled sample and the labeled data of its predicted label (pseudo label provided by classifier $\theta$) as in \citep{du2021contrastive}, and if this minimum distance 
is large then the unlabeled sample is  likely to be OOD, and vice versa.
However, calculating pair-wise distances between all labeled and unlabeled data is computationally expensive. 
\abc{A more efficient method is to summarize the ID labeled data via a data distribution (probability density function) and then compute distances of the unlabeled data to the ID distribution. Since we adopt GMMs to represent the data distribution, then our ID confidence score is based on Mahalanobis distance.}


AL for classical ML models focuses  on training basic learner $\theta$ with a fixed feature representations, while AL for DL jointly optimizes the feature representation $\mathcal{X}$ and classifier $\theta$ 
\citep{zhan2021comp, ren2021survey}. 
Considering that the differences between AL for ML and DL can influence the estimation of the data distributions, and thus the Mahalanobis distance calculation, we adopt different settings 
as follows.
\paragraph{Classical ML tasks.}
We represent the labeled data distribution as a \textbf{GMM}, which is estimated from $\D_l$.
Specifically, since we have labels in $\D_l$, we represent each class $k$ with its own class-conditional \textbf{GMM},
\begin{equation}
\label{gmm}
\small
p_{\text{GMM}}(\mathbf{x}|y = k) = \sum\nolimits_{c = 1}^{C_k} \pi_{c}\s{k} \mathcal{N}(\mathbf{x}|\mu_{c}\s{k}, \Sigma_{c}\s{k}),
\end{equation}
where $(\pi_c\s{k}, \mu_{c}\s{k}, \Sigma\s{k}_{c})$ are the mixing coefficient, mean vector and covariance matrix of the $c$-th Gaussian component of the $k$-th class, and $\mathcal{N}(\mathbf{x}|\mu,\Sigma)$ is a multivariate Gaussian distribution with mean $\mu$ and covariance $\Sigma$. 
The parameters of the class-conditional \textbf{GMM} for class $k$ are estimated from all the labeled data for class $k$, $\D\s{k}_l = \{\mathbf{x}_i | y_i=k\}$ using 
a maximum likelihood estimation (MLE) \citep{reynolds2009gaussian} and the 
Expectation-Maximization (EM) algorithm 
\citep{dempster1977maximum}.
Finally, inspired by \citep{lee2018simple}, we define the ID confidence score $\mathcal{M}(\mathbf{x})$ via the Mahalanobis distance between the unlabeled sample $\mathbf{x}$ and its closest Gaussian component in any class,
\begin{equation}
\label{maha_ml}
\small
\mathcal{M}_{\text{ML}}(\mathbf{x}) = \max_k \max_c 
- \mahaldist{\mathbf{x} - \mu_c\s{k}}{\Sigma_c\s{k}},
\end{equation}
where the Mahalanobis distance is $\mahaldist{\mathbf{x}-\mu}{\Sigma} = (\mathbf{x}-\mu)^T\Sigma^{-1}(\mathbf{x}-\mu)$.

\paragraph{DL tasks.} For DL, we follow \citep{lee2018simple} to calculate $\mathcal{M}(\mathbf{x})$, since it makes good use of both low- and high-level features generated by the deep neural network (DNN) and also applies calibration techniques, e.g., input-preprocessing \citep{liang2018enhancing} and feature ensemble \cite{lee2018simple} to improve the OOD detection capability. 
Specifically, denote $f_{\ell}(\mathbf{x})$ as the output of the $\ell$-th layer of a DNN for input $\mathbf{x}$.
For feature ensemble, class-conditional Gaussian distributions with shared covariance are estimated for each layer and for each class $k$  by calculating the empirical mean and shared covariance $(\hat{\mu}\s{k}_{\ell}, \hat{\Sigma}_{\ell})$:
 \begin{equation}
\label{mean_cov_dl}
\small
\hat{\mu}_{\ell}\s{k} = \tfrac{1}{|\D_l^{(k)}|}\sum_{\mathbf{x} \in \D_l^{(k)}} f_{\ell}(\mathbf{x}),
\quad
\hat{\Sigma}_{\ell} = \tfrac{1}{|\D_l|}
\sum_{k}\sum_{\mathbf{x}\in \D_l^{(k)}}(f_{\ell}(\mathbf{x}) - \hat{\mu}_{\ell}\s{k})(f_{\ell}(\mathbf{x}) - \hat{\mu}_{\ell}\s{k})^T
.
\end{equation}
%
Next we find the closest class to an $\mathbf{x}$ for each layer $\ell$  via $\hat{k}_{\ell} = \arg\min_k \mahaldist{f_{\ell}(\mathbf{x}) - \hat{\mu}_{\ell}\s{k}}{\Sigma_{\ell}}.$
An input pre-processing step is applied to make the ID/OOD data more separable by adding a small perturbation to $\mathbf{x}$: $\hat{\mathbf{x}} = \mathbf{x} + \varepsilon \text{sign}(\nabla_{\mathbf{x}} \mahaldist{f_{\ell}(\mathbf{x}) - \hat{\mu}_{\ell}\s{\hat{k}}}{\hat{\Sigma}_{\ell}})$,
where $\varepsilon$ controls the perturbation magnitude\footnote{In our experiments we follow \citep{lee2018simple} and set $\varepsilon = 0.01$.}. Finally, the ID confidence score is obtained by averaging the Mahalanobis distances for $\hat{\mathbf{x}}$ over the $L$ layers,
\begin{equation}
\label{maha_dl}
\small
\mathcal{M}_{\text{DL}}(\mathbf{x}) = \frac{1}{L} \sum_{\ell=1}^L \max_k - \mahaldist{f_{\ell}(\mathbf{\hat{x}}) - \hat{\mu}_{\ell}\s{k}}{\hat{\Sigma}_{\ell}}.
\end{equation}

\subsection{Monte-Carlo Pareto Optimization Active Learning}
\label{mc-poal-sec}
\abc{Our framework for AL for OOD data contains two criteria, the AL informativeness score and the ID confidence score.}
In normal AL, there are two typical strategies to perform sampling with multiple AL criteria: 
i) \emph{weighted-sum optimization} and ii) \emph{two-stage optimization} \citep{zhan2022comparative}. However, both these optimization strategies have 
drawbacks in our AL for OOD scenario.  
In \emph{weighted-sum optimization}, the objective functions are summed up with weight $\eta$: $\alpha_{\textbf{WeightedSum}}(\mathbf{x}) = \eta \mathcal{U}(\mathbf{x}) + (1- \eta) \mathcal{M}(\mathbf{x})$. However, this introduces an extra hyper-parameter for tuning, but  
the true proportion of OOD data samples in $\D_u$ is unknown, and thus, we cannot tell which criterion is more important. Furthermore, due to the lack of data, there is not enough validation data for properly tuning $\eta$. For \emph{two-stage optimization},  a subset of possible ID samples is first selected with threshold $\mathcal{M}(\mathbf{x}) < \delta$, and then the most informative $b$ samples in the subset are selected by maximizing $\mathcal{U}(\mathbf{x})$. 
This yields a similar problem as 
weighted-sum optimization: if $\delta$ has not been properly selected, we might i) select OOD samples in the first stage (when $\delta$ is too large), and these OOD samples will be more likely to be selected first due to their higher informativeness score in the second stage; ii) select ID samples \abc{that are  close to existing samples} (when $\delta$ is too small), and 
due to the natural conflict between the two criteria, 
the resulting subset 
will be non-informative, thus influencing the AL performance.

Considering this peculiar contradiction between AL and OOD criteria, 
developing a combined optimization method that does not require manual tuning of this tradeoff is of vital importance. Therefore, we propose \textbf{POAL} for balancing $\mathcal{U}(\mathbf{x})$ and $\mathcal{M}(\mathbf{x})$ automatically without requiring hyper-parameters. Consider one iteration of the AL process, where we need to select $b$ samples from $\D_u$.
The size of the search space is the number of combinations $M$ choose $b$, $\mathcal{C}(M,b)$, and the search is an  NP-hard problem in general.
 
 Inspired by \citep{qian2015subset}, we use a binary vector  to represent a candidate subset: $\mathbf{s} \in \{0,1\}^{M}$, where $s_i = 1$ represents $i$-th sample is selected, and $s_i = 0$ otherwise.
 \abc{For the unlabeled set $\D_u$, we denote the vector of AL scores as $\mathbf{u}=[\mathcal{U}(\mathbf{x}_i)]_{i=1}^M$ and the vector of ID confidence scores as $\mathbf{m}=[\mathcal{M}(\mathbf{x}_i)]_{i=1}^M$.  The two criteria scores for the subset $\mathbf{s}$ are then computed as $o_U(\mathbf{s}) = \mathbf{s}^T\mathbf{u}$ and $o_M(\mathbf{s}) = \mathbf{s}^T\mathbf{m}$.\footnote{\abc{To keep the two criteria consistent, we normalize the ID confidence scores: $\mathbf{m} \leftarrow \max(\mathbf{m}) - \mathbf{m}$.}}
 We next define the following rank relationships between two candidate subsets $\mathbf{s}$ and $\mathbf{s}'$ \citep{qian2015subset}:}
 \begin{compactitem}
 \item  $\mathbf{s}' \preceq \mathbf{s}$ denotes that \zxy{$\mathbf{s}'$ is \emph{dominated} by $\mathbf{s}$}, such that both scores for $\mathbf{s}'$ are no better than those of $\mathbf{s}$, i.e., 
 $o_U(\mathbf{s}') \leq o_U(\mathbf{s})$ and 
 $o_M(\mathbf{s}') \leq o_M(\mathbf{s})$.
 \item $\mathbf{s}' \prec \mathbf{s}$ denotes that \zxy{$\mathbf{s}'$ is \emph{strictly dominated} by $\mathbf{s}$}, such that 
 $\mathbf{s}'$ has one strictly smaller score (e.g., $o_U(\mathbf{s}') < o_U(\mathbf{s})$) and one score that is not better (e.g.,  $o_M(\mathbf{s}') \leq o_M(\mathbf{s})$).
 \item $\mathbf{s}$ and $\mathbf{s}'$ are \emph{incomparable} if \dzy{both} $\mathbf{s}$ is not \dzy{\emph{dominated} by} $\mathbf{s}'$ and $\mathbf{s}'$ is not \dzy{\emph{dominated} by} $\mathbf{s}$.
 \end{compactitem}
 \zxy{Our goal is to find optimal subset solution(s) $\mathbf{s}$, with $\sum_i s_i = b$, that dominates the remaining subset solutions also satisfy $\sum s'_i = b$.}
 Due to the large searching space, 
 it is impossible to traverse all possible subset solutions. Thus we propose Monte-Carlo \textbf{POAL} for fixed-size subset selection. Monte-Carlo \textbf{POAL} iteratively 
 generates a candidate solution $\mathbf{s}$ at random, and checks it against the current Pareto set $\mathcal{P}=\{\mathbf{s}_1,\mathbf{s}_2,\cdots\}$.
 If there is no candidate solution in $\mathcal{P}$ that \emph{strictly dominates} $\mathbf{s}$, 
 then $\mathbf{s}$ is added to $\mathcal{P}$ and all candidate solutions in $\mathcal{P}$ that are \emph{dominated} by $\mathbf{s}$ are removed.

Our proposed Monte-Carlo \textbf{POAL} method is inspired by \textbf{POSS} \citep{qian2015subset}, but there are essential differences. \textbf{POSS} only supports one criterion (e.g., just $\mathcal{U}(\mathbf{x})$), while the other criterion is that the subset size does not exceed $b$ (i.e., $\sum_i s_i \leq b$).  
This results in different subset sizes for the candidate solutions in $\mathcal{P}$ when using \textbf{POSS}, and \textbf{POSS} can finally pick one solution from $\mathcal{P}$ by maximizing/minimizing the only criterion. Moreover, in \textbf{POSS}, in each iteration, the generated solution is not \zxy{completely} random, as it is related to candidate solutions in $\mathcal{P}$. 
\textbf{POSS} first randomly selects one solution in $\mathcal{P}$, then generates a new solution by randomly flipping each bit of selected solution with equal probability\zxy{, which may results in the candidate subset size that is not $b$}. However, in our task, we need fixed-size subset solutions and we have two conflict criteria, and thus  \textbf{POSS} cannot satisfy our requirements.
Due to these differences in how candidate solutions are generated and the number of criteria
, the theoretical bound on the number of iterations $T$ needed for convergence of \textbf{POSS} 
($E[T]\leq 2 e b^2 M$) 
\citep{qian2015subset} cannot be applied to our 
Monte-Carlo \textbf{POAL} method. Thus, when should the iterations of our Monte-Carlo \textbf{POAL} terminate? 
We noticed in previous research \citep{qian2017optimizing} 
that the Pareto set empirically converges much faster than $E[T]$ (see Fig.~2 in \citep{qian2017optimizing}). Therefore, we propose an \emph{early-stopping} technique. Detailed comparisons between \textbf{POAL} and \textbf{POSS} are in Appendix A.4.2. 
\vspace{-0.3cm}
\paragraph{Early-stopping.} The Monte-Carlo \textbf{POAL} should terminate when there is no significant change in the Pareto set $\mathcal{P}$ after many successive iterations, which indicates that a randomly generated $\mathbf{s}$ has little probability to change $\mathcal{P}$ since most non-dominated solutions are already included in $\mathcal{P}$. We propose an early-stopping strategy with reference to automatic termination techniques in Multi-objective Evolutionary Algorithms \citep{saxena2016entropy}. Firstly, we need to define the difference between two Pareto sets, $\mathcal{P}$ and $\mathcal{P}'$.
We represent each candidate solution $\mathbf{s}$ as a 2-dimensional feature vector, $v(\mathbf{s}) = (o_U(\mathbf{s}), o_M(\mathbf{s}))$. The two Pareto sets can then be compared via their feature vector distributions, 
$\mathbf{V}(\mathcal{P}) = \{v(\mathbf{s})\}_{\mathbf{s}\in \mathcal{P}}$ and
$\mathbf{V}(\mathcal{P}') = \{v(\mathbf{s}')\}_{\mathbf{s}'\in \mathcal{P}'}$.
%
In our paper, we employed maximum mean discrepancy (MMD) \citep{gretton2006kernel} to measure the difference:
\begin{equation}\label{mmd}
\small
 {\rm MMD}(\mathcal{P},\mathcal{P}') = 
 \Big\|\tfrac{1}{|\mathcal{P}|} \sum_{\mathbf{s}\in\mathcal{P}}  v(\mathbf{s}) - \tfrac{1}{|\mathcal{P}'|} \sum_{\mathbf{s}'\in\mathcal{P}'}  v(\mathbf{s}')\Big\|_{\mathcal{H}}.
\end{equation}
The distance is based on the notion of embedding probabilities in a reproducing kernel Hilbert space $\mathcal{H}$. In our work, we utilize the RBF kernel with various bandwidths. Second, at iteration $t$, we compute the mean and standard deviation (SD) of the MMD scores in a sliding window of the previous $s_w$ iterations, 
\begin{equation}\label{mean_sd}
\small
M_{t} = \frac{1}{s_w}\sum_{i=t-s_w+1}^{t} {\rm MMD}(\mathcal{P}_{i-1}, \mathcal{P}_{i}), \quad 
S_{t} = \frac{1}{s_w}\sum_{i=t-s_w+1}^{t} ({\rm MMD}(\mathcal{P}_{i-1}, \mathcal{P}_{i}) - M_{t})^2.
\end{equation}
If there is no significant difference in the mean and SD after several iterations, (e.g., $M_t$ and $S_t$ do not change within 2 decimal points), then 
we assume that $\mathcal{P}$ has converged and we can stop iterating.


Finally, after obtaining the converged $\mathcal{P}$, we need to pick one non-dominated solution in $\mathcal{P}$ as the final subset selection. 
We denote the final selection as
$\mathbf{s}^* = \arg\max\nolimits_{\mathbf{s}\in \mathcal{P}} \mathcal{F}(\mathbf{s})$,  where $\mathcal{F}(\mathbf{s})$ is the final selection criteria.
In our work, we set 
\abcn{$\mathcal{F}(\mathbf{s}) = \sum\nolimits_{\mathbf{s}' \in \mathcal{P}} \mathbf{s}^T\mathbf{s}'$.}
That is, we select the subset with maximum intersection with other non-dominated solutions. The maximum intersection operation implies a weighting process, \abc{where the final selected subset contains samples that were commonly used in other non-dominated solutions.}
The whole process of our Monte-Carlo \textbf{POAL} with early stopping is detailed in Appendix Algorithm A1.
\vspace{-0.3cm}
\paragraph{Pre-selection for large-scale data.} 
We next consider using \textbf{POAL} on large datasets. The searching space of size $\mathcal{C}(M,b)$ is extremely large for a large unlabeled data pool.
We propose an \textbf{optional} pre-selecting technique to decrease the searching space. 
Our goal is to select an optimistic subset $\D_\text{sub}$ from $\D_u$. We conduct \zxy{naive} Pareto optimization on the \zxy{original unlabeled data pool.} 
Firstly, an optimal Pareto front $\mathcal{P}_{\text{naive}}$, a set of non-dominated data points, is selected by normal Pareto optimization with objectives $\mathcal{U}(\mathbf{x})$ and $\mathcal{M}(\mathbf{x})$. Since the size of $\mathcal{P}_{\text{naive}}$ is not fixed and might not meet our requirement of the minimum pre-selected subset size $s_m$, \zxy{we iteratively update $\D_\text{sub}$ by firstly adding the data points in $\mathcal{P}_{\text{naive}}$ to $\D_\text{sub}$, and then excluding $\D_\text{sub}$ from $\D_u$ for the next round of naive Pareto optimization.
This iteration terminates when $|\D_{\text{sub}}| \geq s_m$.} 
The pseudo code of pre-selecting process is presented in Appendix Algorithm A2. The pre-selected subset size $s_m$ is set according to personal requirements (i.e., the computing resources and time budget).
In our paper, we set $s_m = 6b$.

\section{Experiment}
\label{experiment}
In this section, we evaluate the effectiveness of our proposed Monte-Carlo \textbf{POAL} on both classical ML and DL tasks. 
\vspace{-0.2cm}
\subsection{Experimental Design}
\vspace{-0.2cm}
\paragraph{Datasets.} For classical ML, we use pre-processed data from LIBSVM\citep{chang2011libsvm}:
\begin{compactitem}
\item Synthetic dataset \emph{EX8} uses \emph{EX8a} as ID data 
and \emph{EX8b} as OOD data \citep{andrew2008stan}. 
\item \emph{Vowel}  \citep{asuncion2007uci, aggarwal2015theoretical} has $11$ classes, and we use $7$ classes as ID data and the remaining for OOD data.
\item \emph{Letter} \citep{frey1991letter, asuncion2007uci} has $26$ classes. We use 10 classes (\emph{a-j}) as ID and the remaining 16 classes (\emph{k-z}) as OOD. We construct $16$ datasets with increasing OOD classes gradually: \emph{letter(a-k)}, \emph{letter(a-l)},..., \emph{letter(a-z)}.
\end{compactitem}
For DL, we adopt the following image datasets:
\begin{compactitem}
\item \emph{CIFAR10} \citep{krizhevsky2009learning} has $10$ classes, and we construct two datasets: \emph{CIFAR10-04} splits the classes according to ID:OOD ratio of 6:4, and \emph{CIFAR10-06} splits data with ratio as 4:6.
\item \emph{CIFAR100} \citep{krizhevsky2009learning} has $100$ classes, and we construct \emph{CIFAR100-04} and \emph{CIFAR100-06} using the same ratios as \emph{CIFAR10}.
\end{compactitem}

\paragraph{Baselines.} Our work helps existing AL methods select more informative ID data samples while preventing OOD data selection. Thus, in our experiments, we adopt \textbf{ENT} as our basic AL sampling strategy for our \textbf{POAL} framework.
We compare against baseline methods without OOD detection:
1) normal \textbf{ENT}; 2) random sampling (\textbf{RAND}); 
3) Mahalanobis distance as sampling strategy (\textbf{MAHA}).
To show the effectiveness of our MC Pareto optimization, we compare against 2 alternative combination strategies:
1) weighted sum optimization (\textbf{WeightedSum}) using weights $\eta = \{0.2, 1.0, 5.0\}$; 2) two-stage optimization (\textbf{TwoStage}) using threshold $\delta = \texttt{mean}(\mathbf{m})$.
We also compare against several widely adopted AL techniques, $k$\textbf{-Means}, \textbf{LPL} \citep{yoo2019learning} and \textbf{BALD} \citep{gal2017deep}, to observe the performance under OOD data scenarios.
Finally, we report the oracle results of \textbf{ENT} where only ID data is selected first (denoted as \textbf{IDEAL-ENT}), which serves as an upper-bound of our \textbf{POAL}.

\paragraph{Implementation details.} \zxy{The training/test split of the datasets are fixed in the experiments},  
while the initial label set and unlabeled set are randomly generated from the training set. Experiments are repeated  $100$ times in classical ML tasks and $3$ times for DL tasks.
We use the same basic classifier for each dataset (details in Appendix A.2.2).
To evaluate model performance, we measure accuracy and plot accuracy vs.~budget curves to present the performance change with increasing labeled samples. To calculate average performance, we compute area under the accuracy-budget curve (AUBC) \citep{zhan2021comp}, 
with higher values reflecting better performance under varying budgets.
More details about the experiments are in Appendix A.2, including dataset splits, $\text{ID}:\text{OOD}$ ratios, basic learner settings, and AL hyper-parameters (budget $B$, batch size $b$, and initial label set size), etc. 

\subsection{Results on classical ML tasks}

The overall performance on various datasets are shown in Figure~\ref{classical_ml}.
We first compare the performance of \textbf{POAL} against each single component score, \textbf{ENT} and \textbf{MAHA}. \textbf{POAL} is closest to the oracle performance of \textbf{IDEAL-ENT}, and both \textbf{ENT} and \textbf{MAHA} are not effective when used as a single selection strategy under OOD scenarios.  
Next we compare \textbf{POAL} against other combination methods using the same $\mathcal{U}$ and $\mathcal{M}$ scores.
\abc{On all datasets, \textbf{POAL} is superior to \textbf{WeightedSum} and \textbf{TwoStage}, demonstrating that joint Pareto optimization better considers the two conflicting criteria.}



\begin{figure*} [htb]
\centering
\subfloat[EX8]{\includegraphics[width=0.2\linewidth]{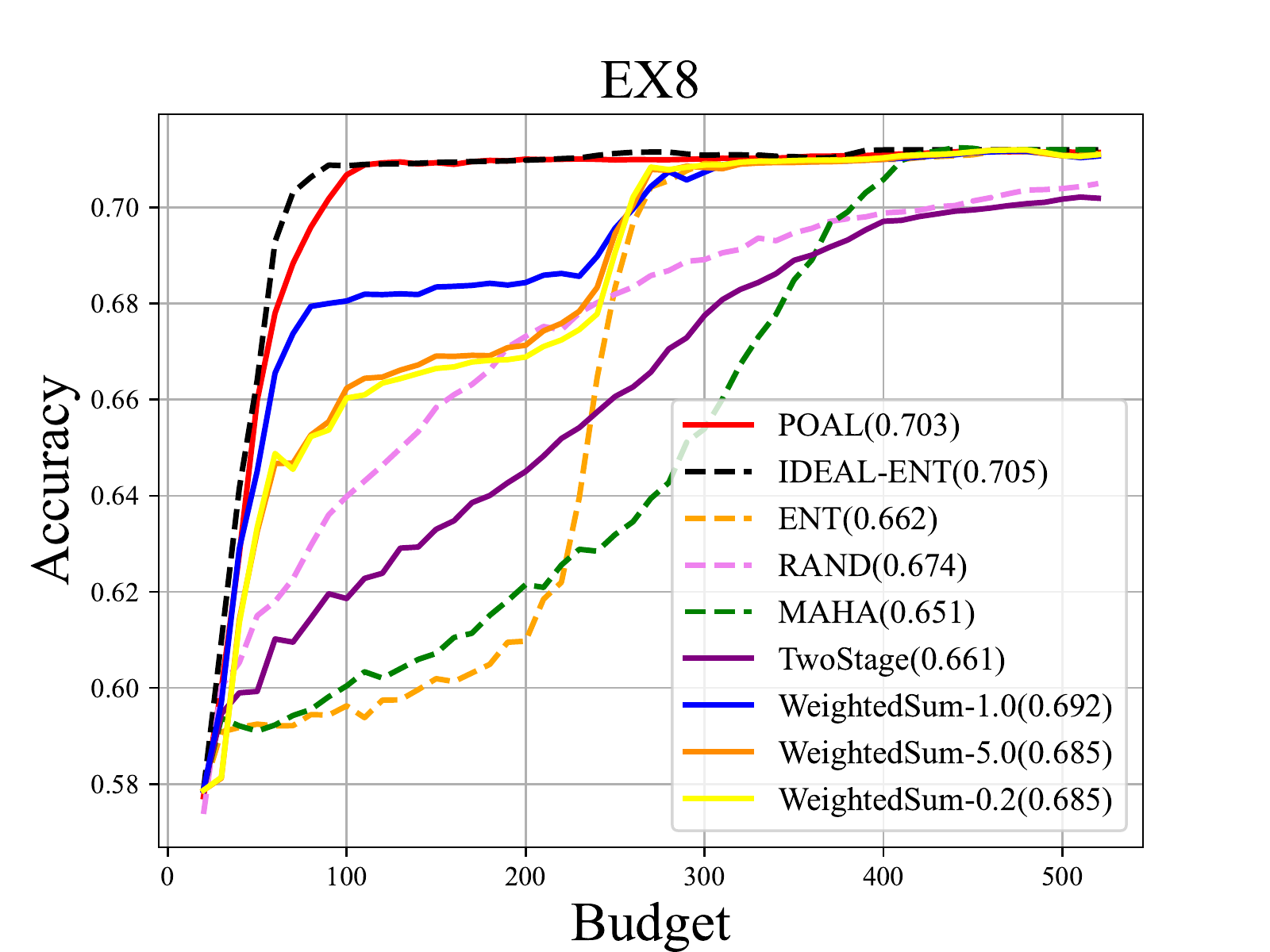}}
\subfloat[vowel]{\includegraphics[width=0.2\linewidth]{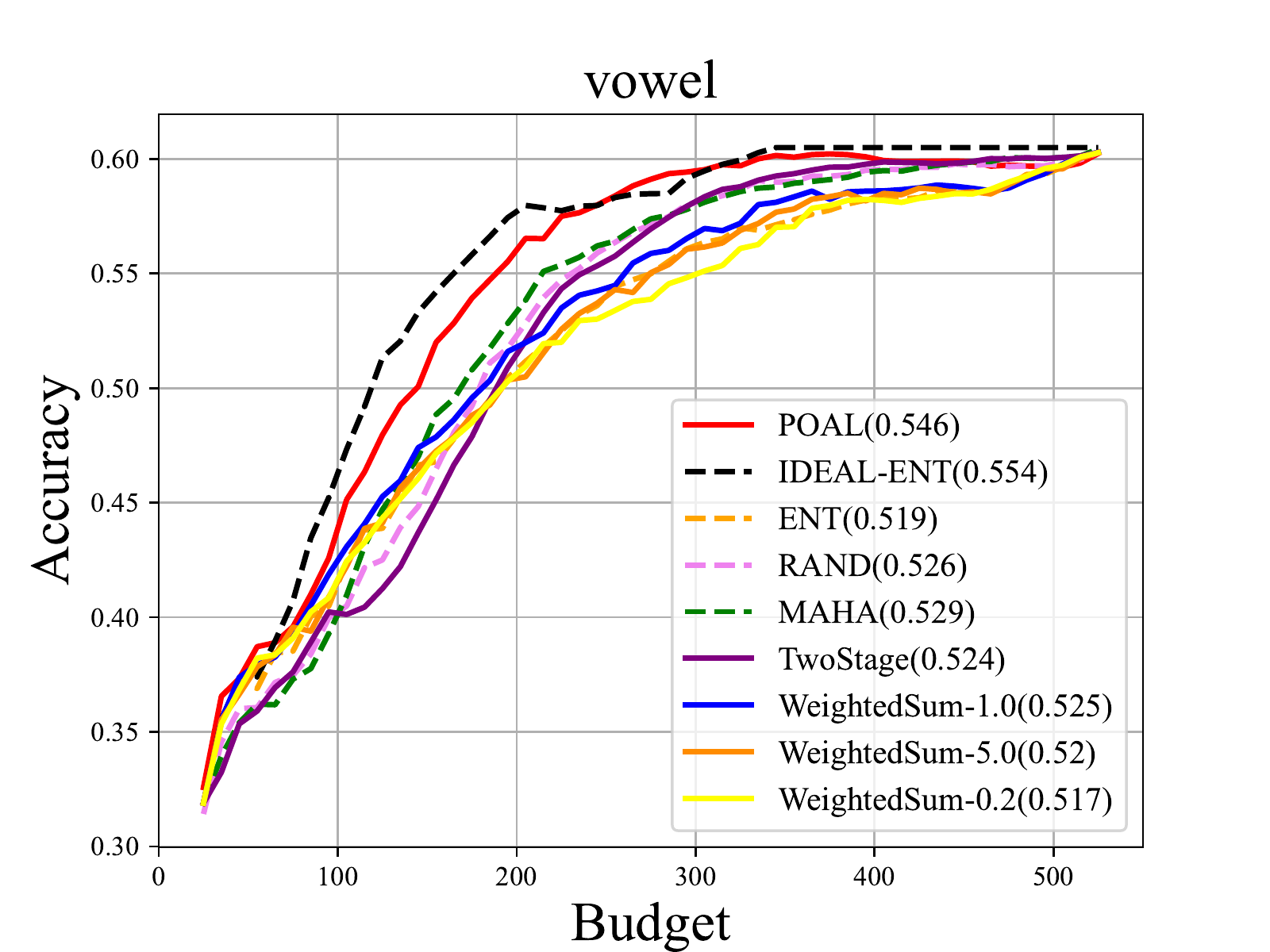}}
\subfloat[letter(a-k)]{\includegraphics[width=0.2\linewidth]{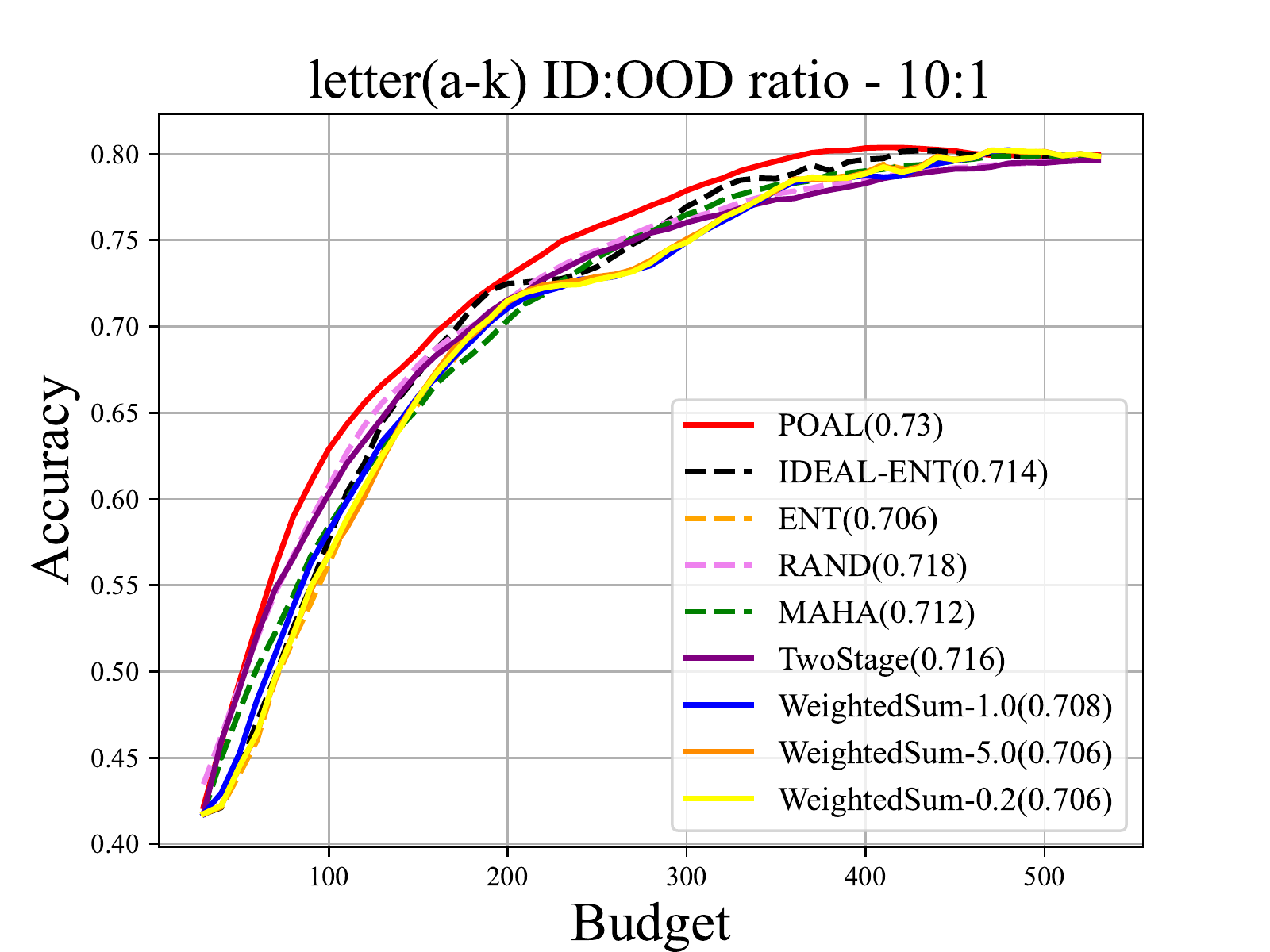}}
\subfloat[letter(a-z)]{\includegraphics[width=0.2\linewidth]{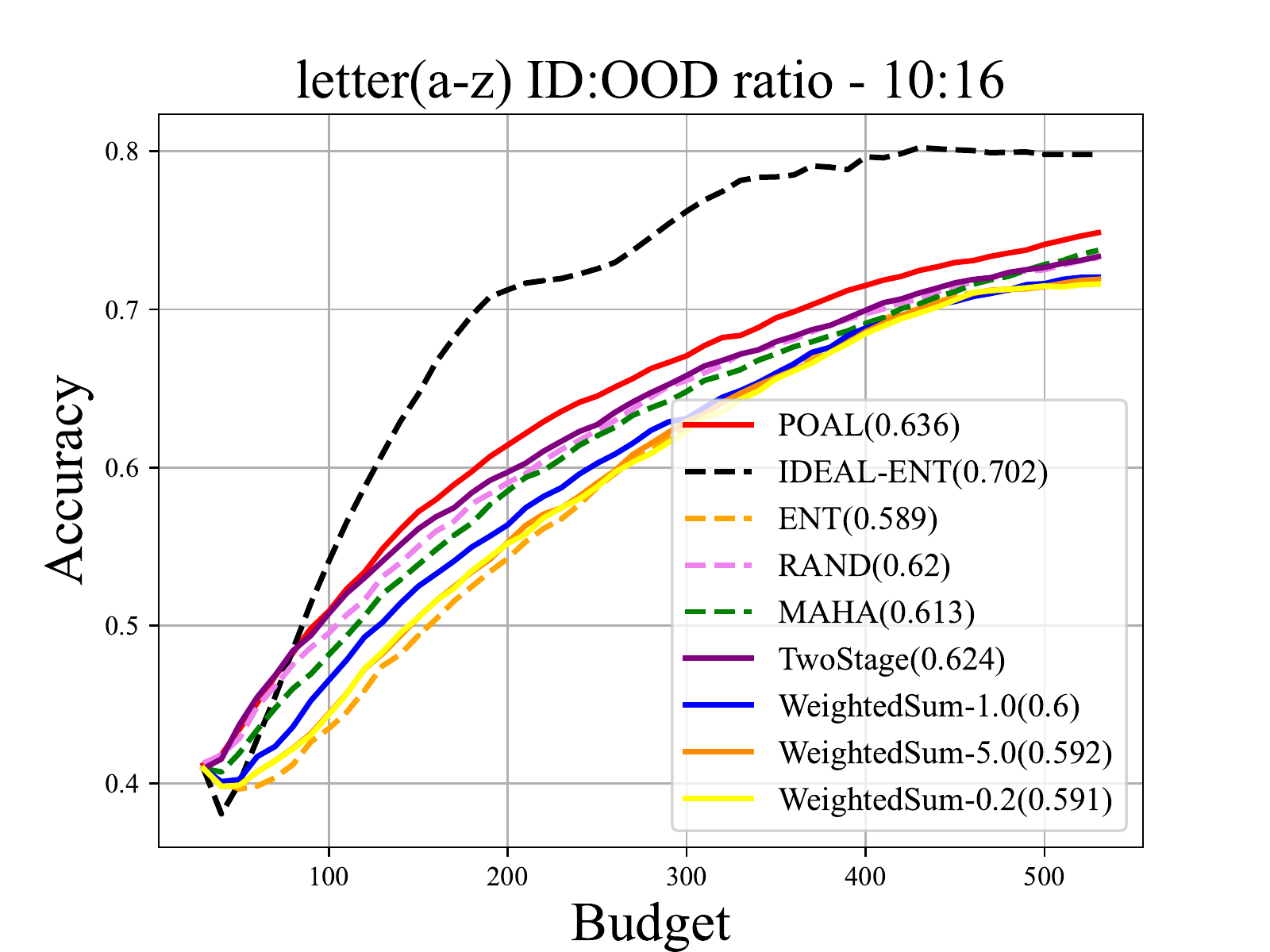}}
\subfloat[letter(AUBC)]{\includegraphics[width=0.2\linewidth]{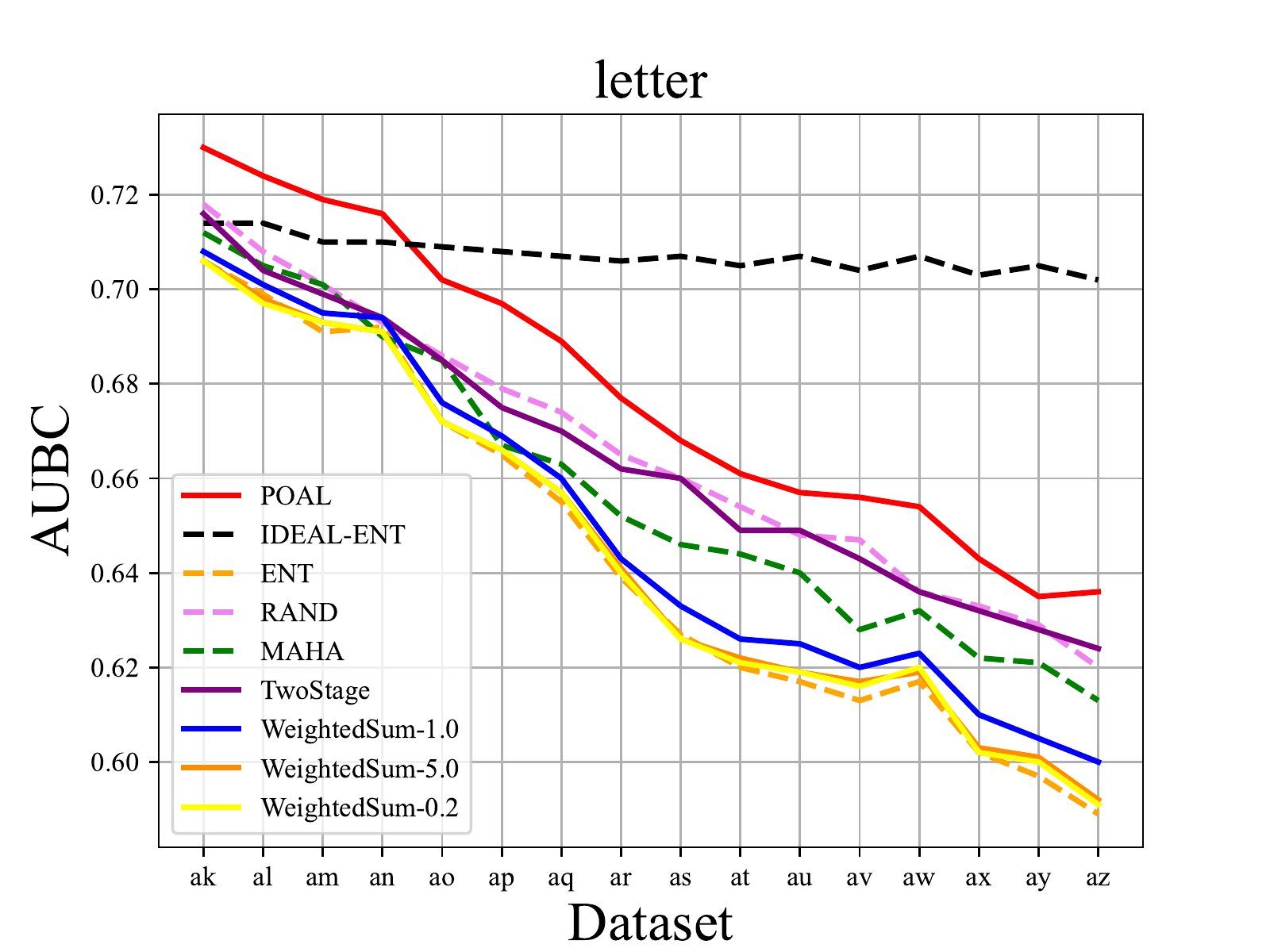}}
\caption{\zxy{(a)-(d) are accuracy vs. budget curves} for classical ML tasks. The AUBC performances are shown in parentheses in the legend.  \zxy{To observe the effect of increasing ID:OOD ratio on \emph{letter} datasets, we plot AUBC vs. dataset curves in (e). The complete figures (more \emph{letter} datasets) are shown in Appendix Figure A2.}
}
\label{classical_ml}
\end{figure*}
We next analyze the results for each dataset. The synthetic dataset \emph{EX8} has two properties: the ID data is non-linearly separable and the OOD data is far from ID data (see data distribution in Appendix Figure A1-a). These properties make the calculation of $\mathcal{M}$ more accurate. \zxy{Meanwhile, both OOD and ID data close to the decision boundaries have high entropy score} 
(see Figure~\ref{relation}). As shown in Figure~\ref{classical_ml}-a, \textbf{ENT} always selects OOD data with higher entropy until all OOD data are selected (210 OOD data are in \emph{EX8}), and it performs even worse than \textbf{RAND}. \textbf{MAHA} prefers to select ID samples first (see Appendix Figure A3), but the data samples that have smaller Mahalanobis distance are also \abc{easily classified} and far from decision boundaries, and thus not informative.
This confirms the claim that 
both informativeness/uncertainty-based AL measure and ID confidence could not be directly used in AL under OOD scenarios. The curve of our \textbf{POAL} is  close to the oracle \textbf{IDEAL-ENT}, which indicates that \textbf{POAL} selects the most informative ID data,  while excluding the OOD samples. 

For real-life datasets, \textbf{POAL} also demonstrates its superiority on \emph{vowel} (see Figure~\ref{classical_ml}-b), a multi-class classification task. In \emph{letter}, we fixed the ID data (letters \emph{a-j}) and gradually increase the OOD data (letters \emph{k-z}).
From Figure~\ref{classical_ml}e, we observe that \zxy{except for \textbf{IDEAL-ENT}, all methods are influenced by increasing OOD data, e.g., the AUBC of \textbf{RAND} is $0.718$ in Figure~\ref{classical_ml}c, $0.62$ in Figure~\ref{classical_ml}d. Although our method is also influenced by increasing OOD data, it is still superior to all baselines.}
\textbf{MAHA} performs better on distinct ID-/OOD- distribution data scenarios. However, if there is overlap between the two distributions,
\textbf{MAHA} does not perform very well due to the overlap of ID/OOD distributions (see appendix Figures A1 and A3). Similar conclusions were also reported  in \citep{ren2019likelihood}, where \textbf{MAHA} does not perform well on genomic datasets, since OOD bacteria classes are interlaced under the same taxonomy. 



\subsection{Results on DL tasks}
\label{overall_perform_dl}
\begin{figure*} [htb]
\centering
\subfloat[CIFAR10-04]{\includegraphics[width=0.25\linewidth]{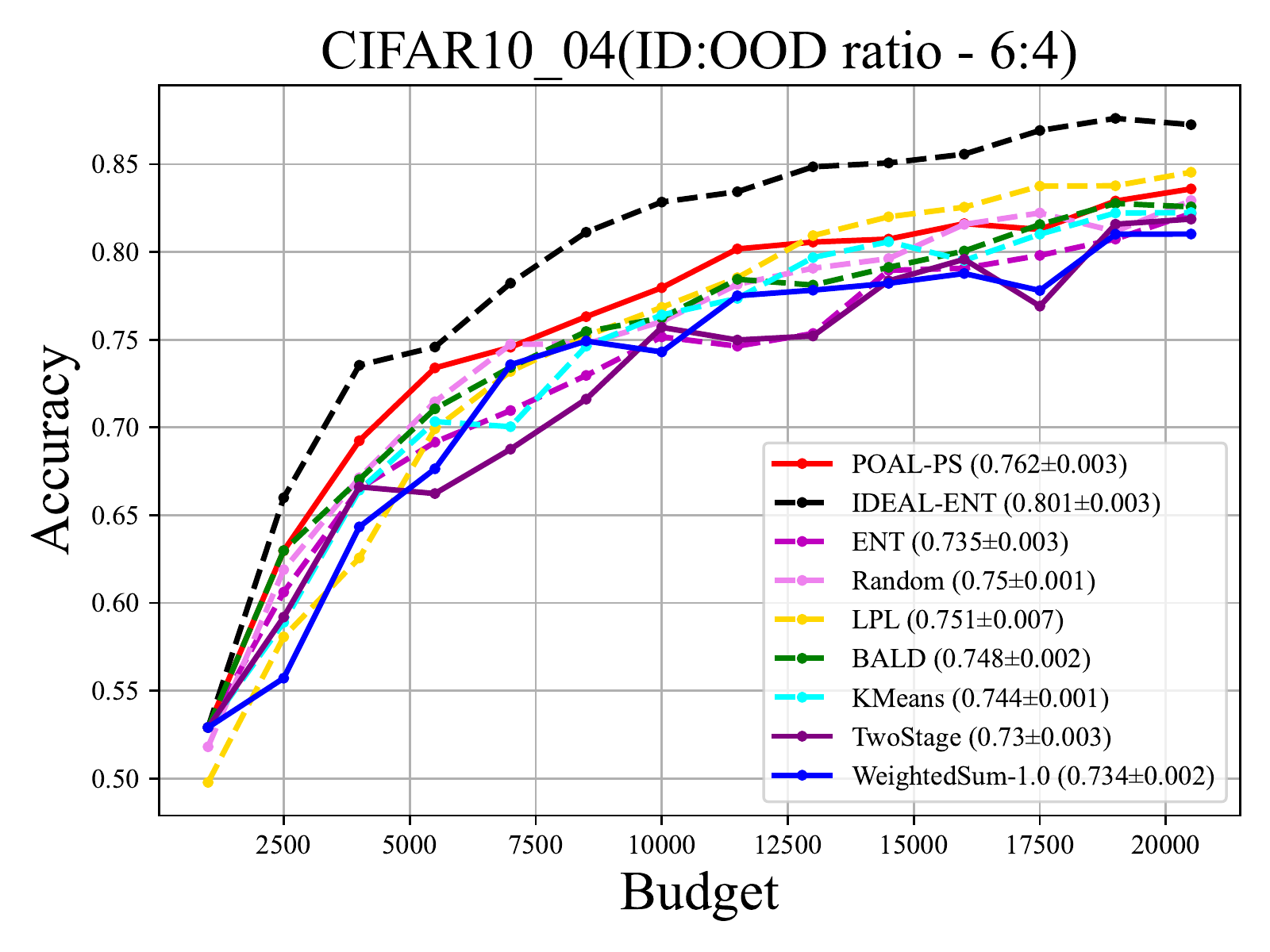}}
\subfloat[CIFAR10-06]{\includegraphics[width=0.25\linewidth]{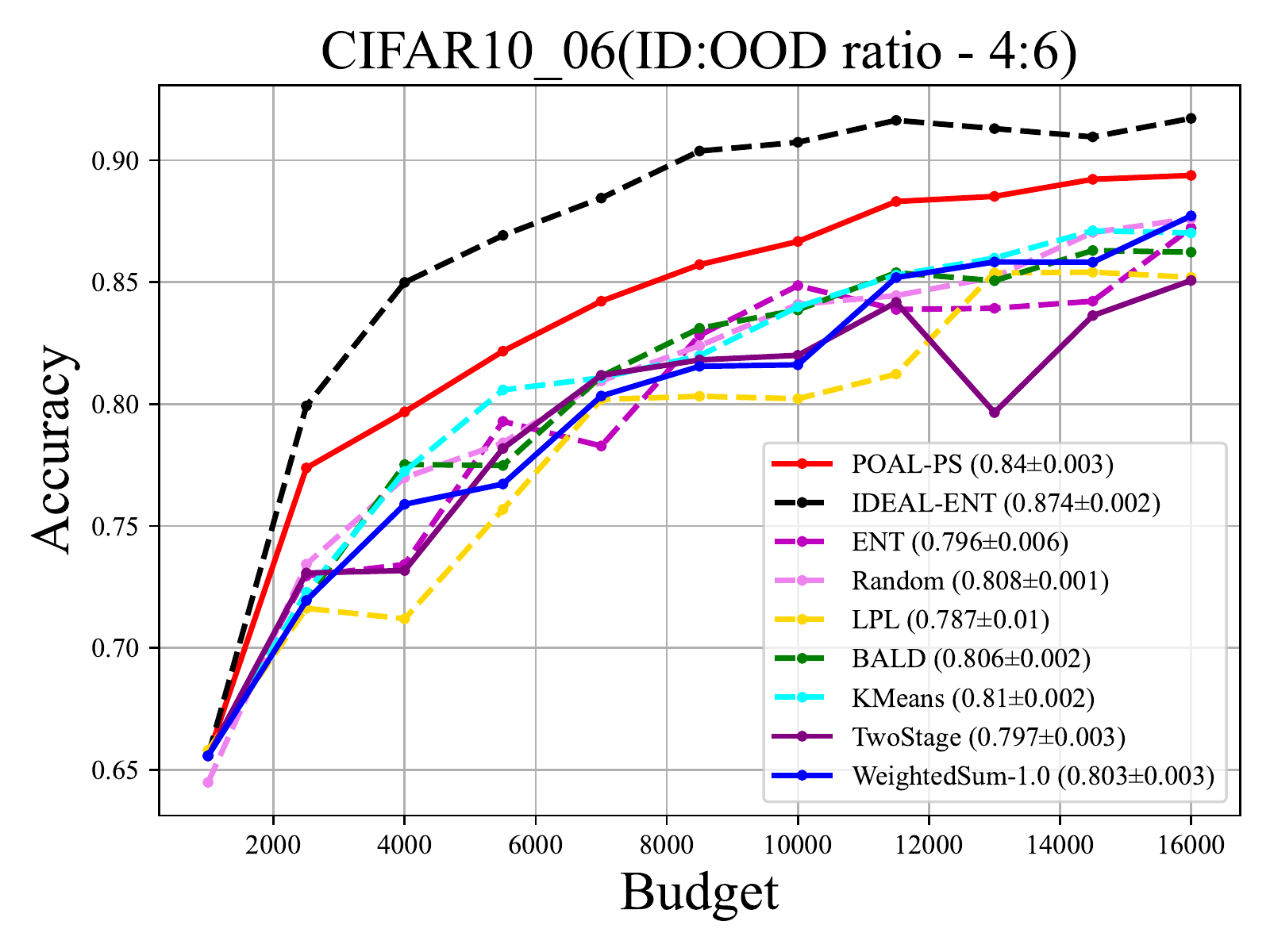}}
\subfloat[CIFAR100-04]{\includegraphics[width=0.25\linewidth]{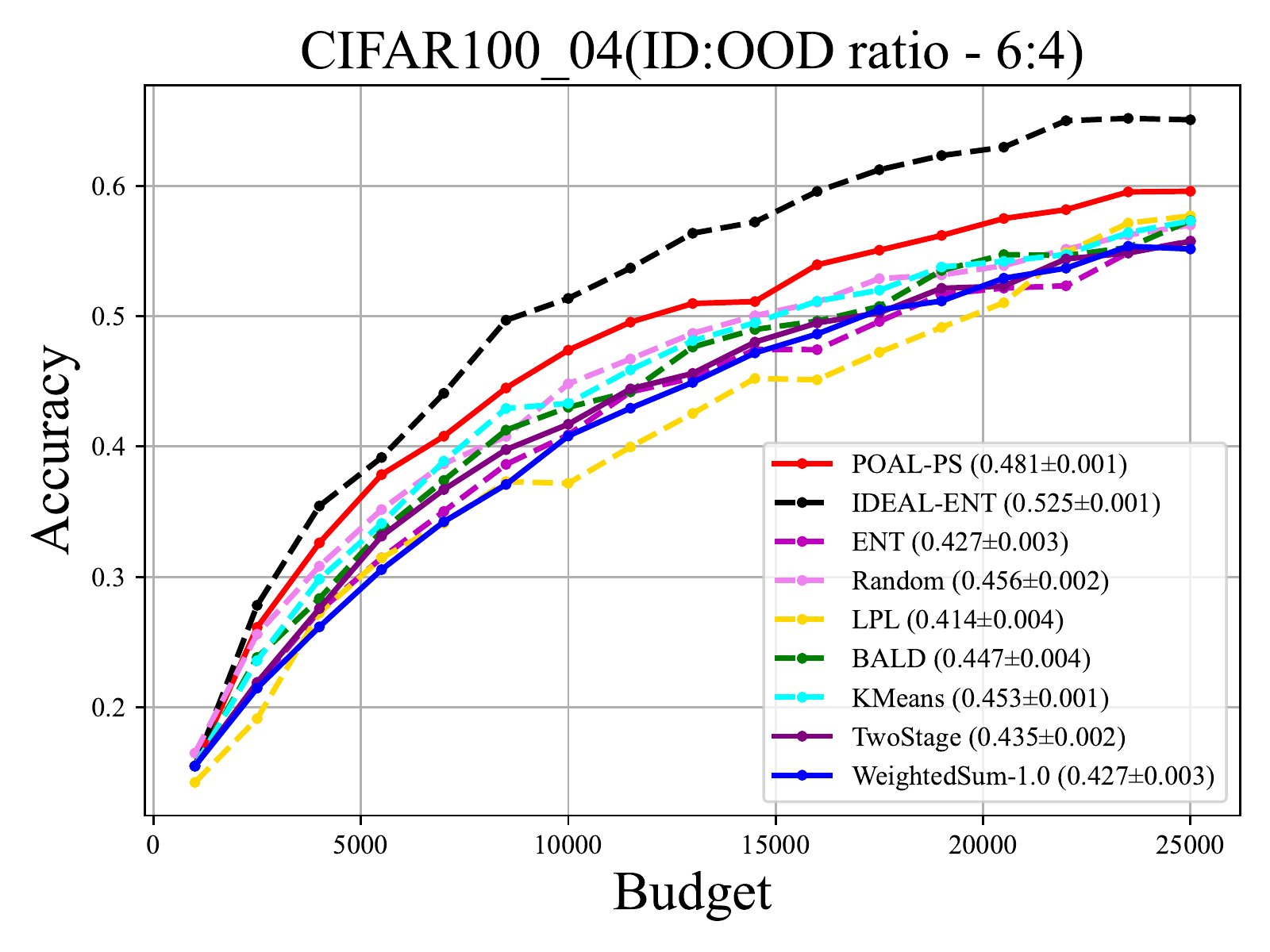}}
\subfloat[CIFAR100-06]{\includegraphics[width=0.25\linewidth]{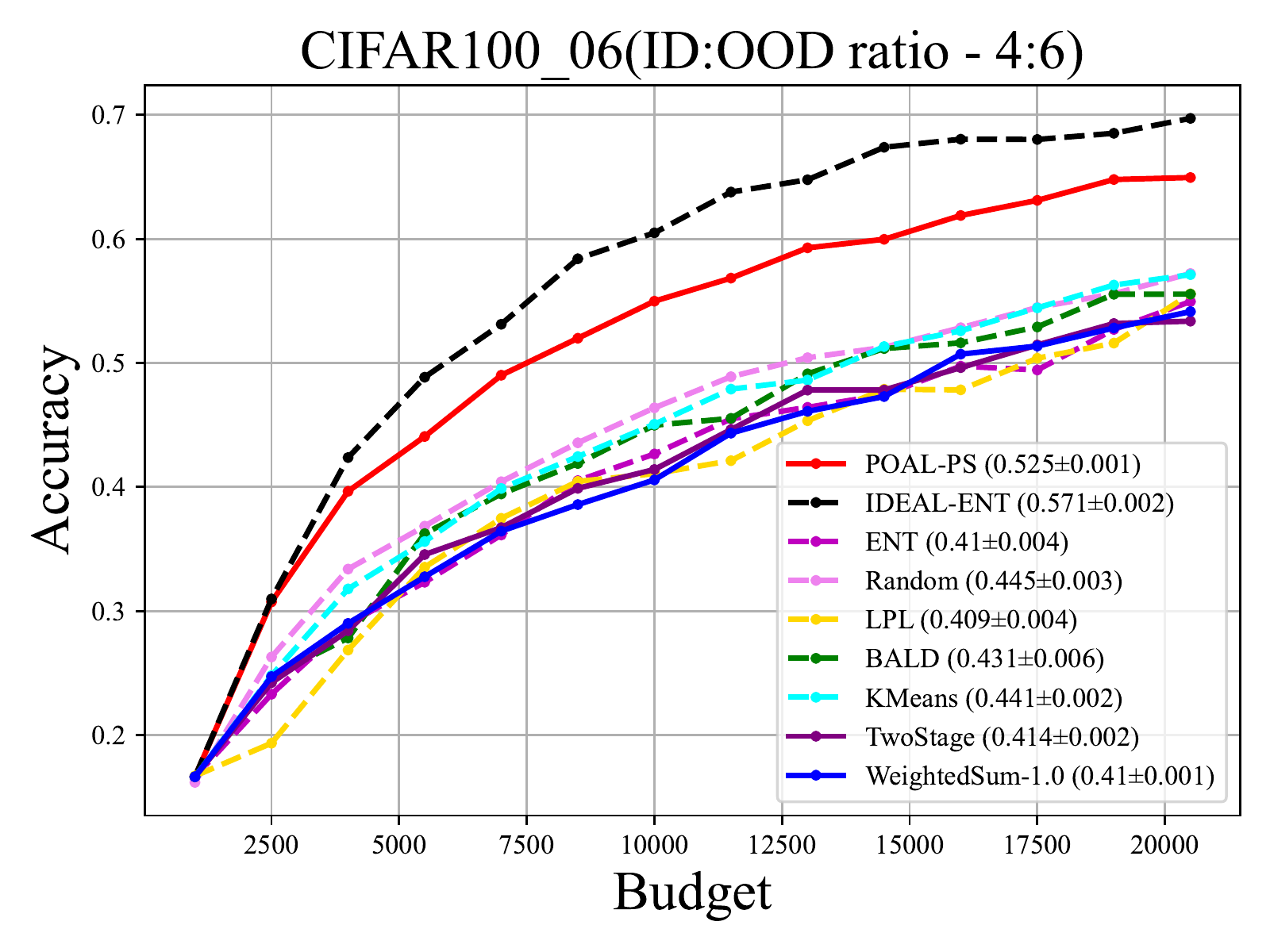}}

\subfloat[CIFAR10-04 OOD]{\includegraphics[width=0.25\linewidth]{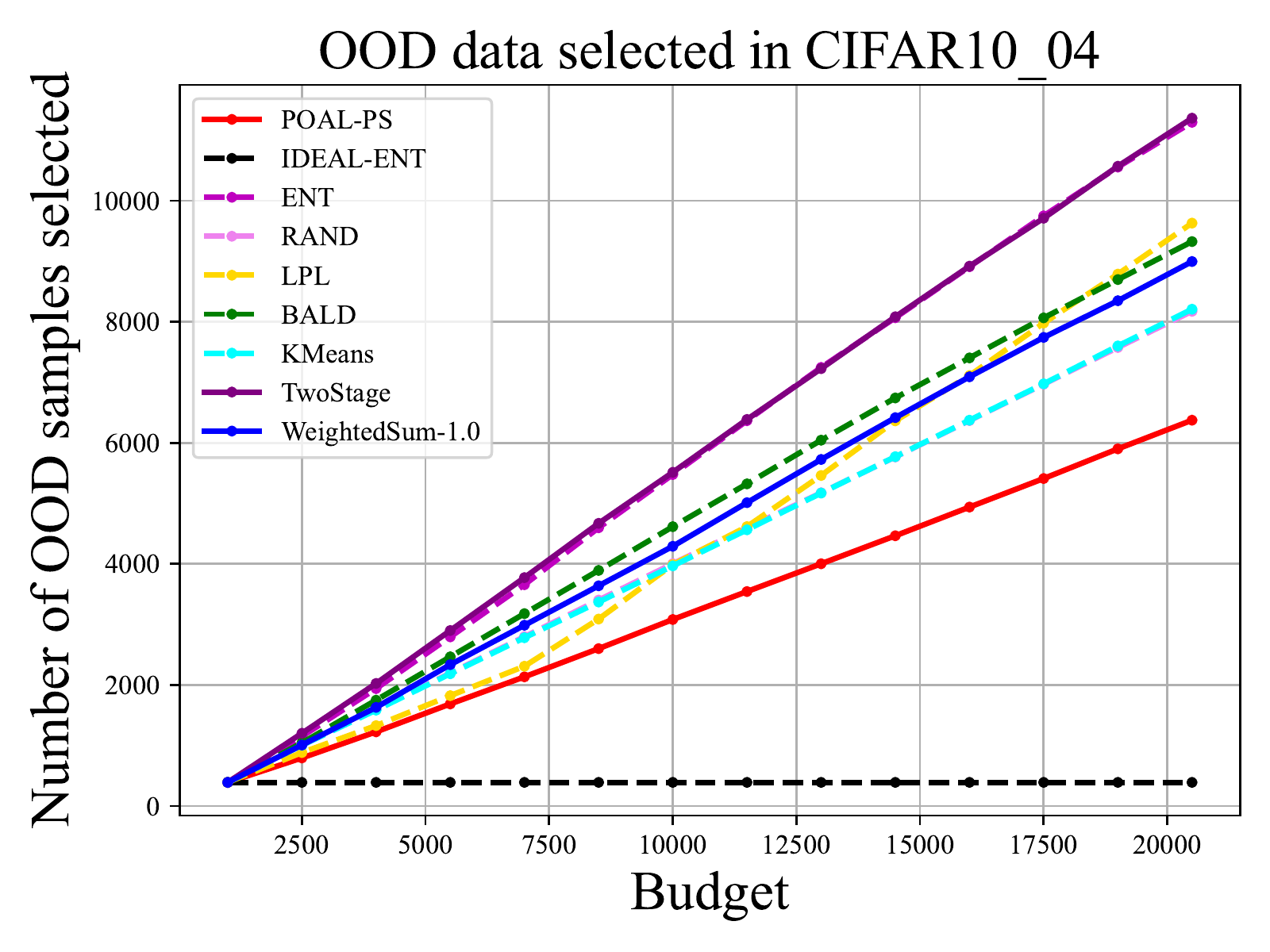}}
\subfloat[CIFAR10-06 OOD]{\includegraphics[width=0.25\linewidth]{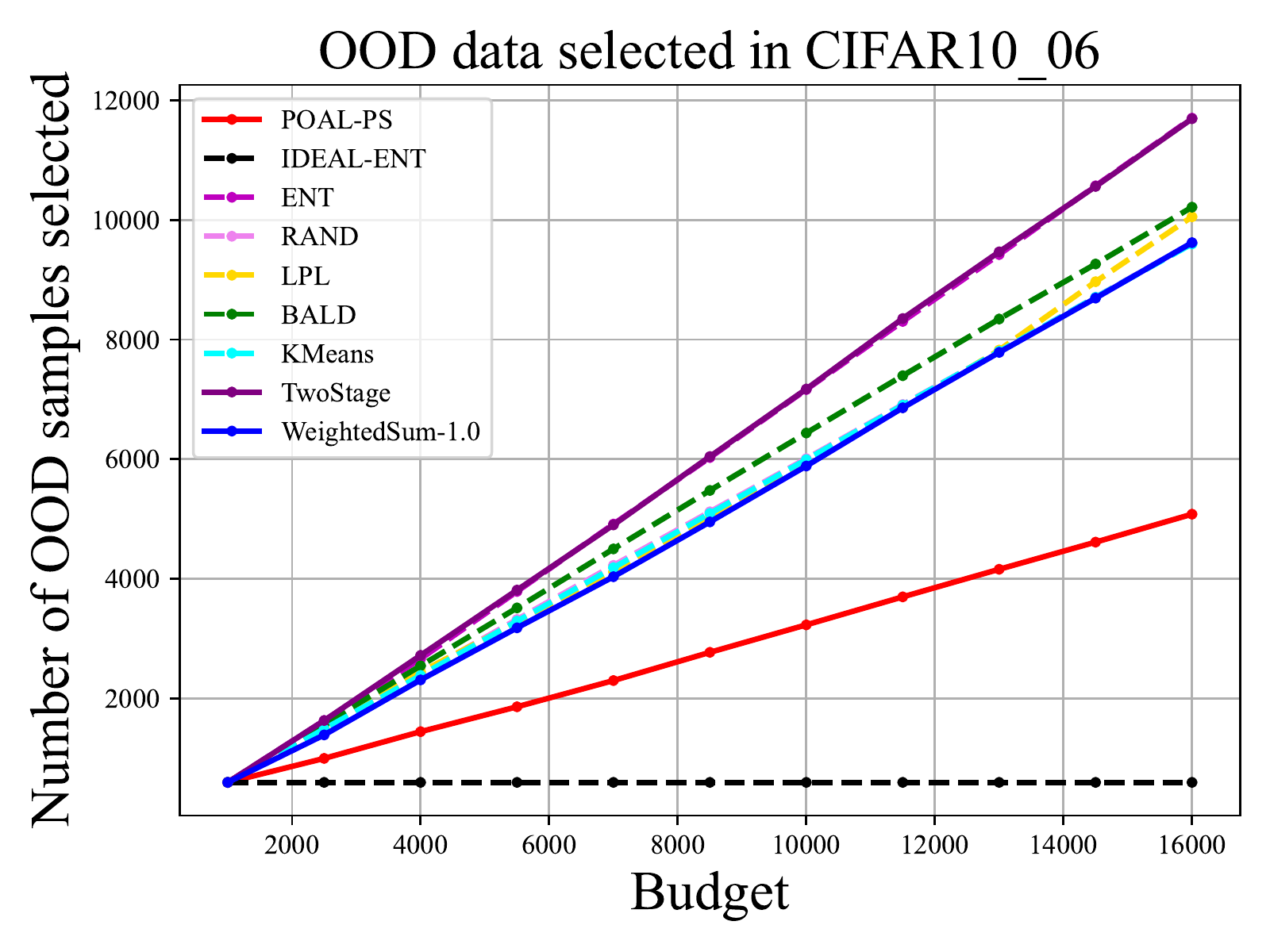}}
\subfloat[CIFAR100-04 OOD]{\includegraphics[width=0.25\linewidth]{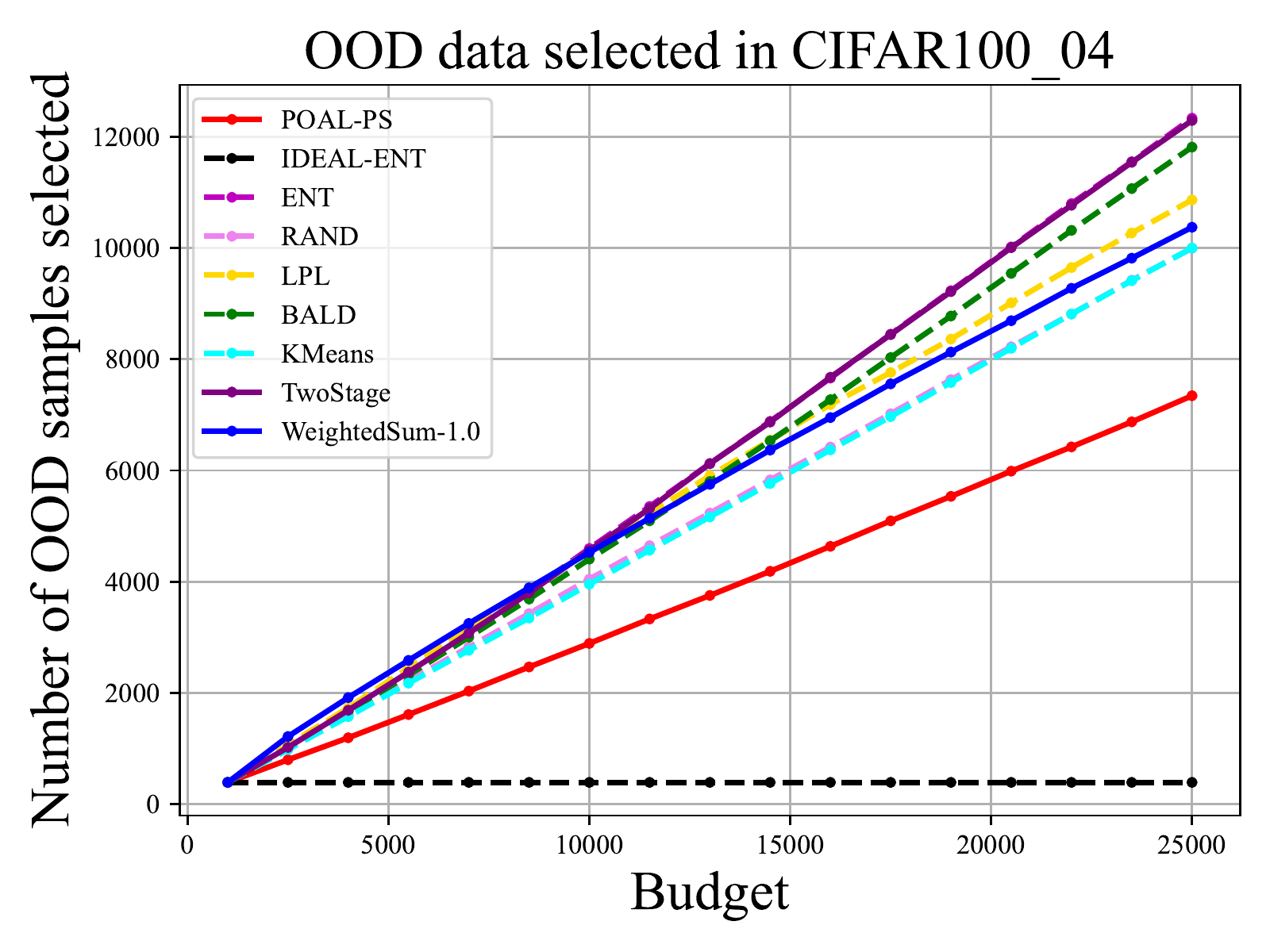}}
\subfloat[CIFAR100-06 OOD]{\includegraphics[width=0.25\linewidth]{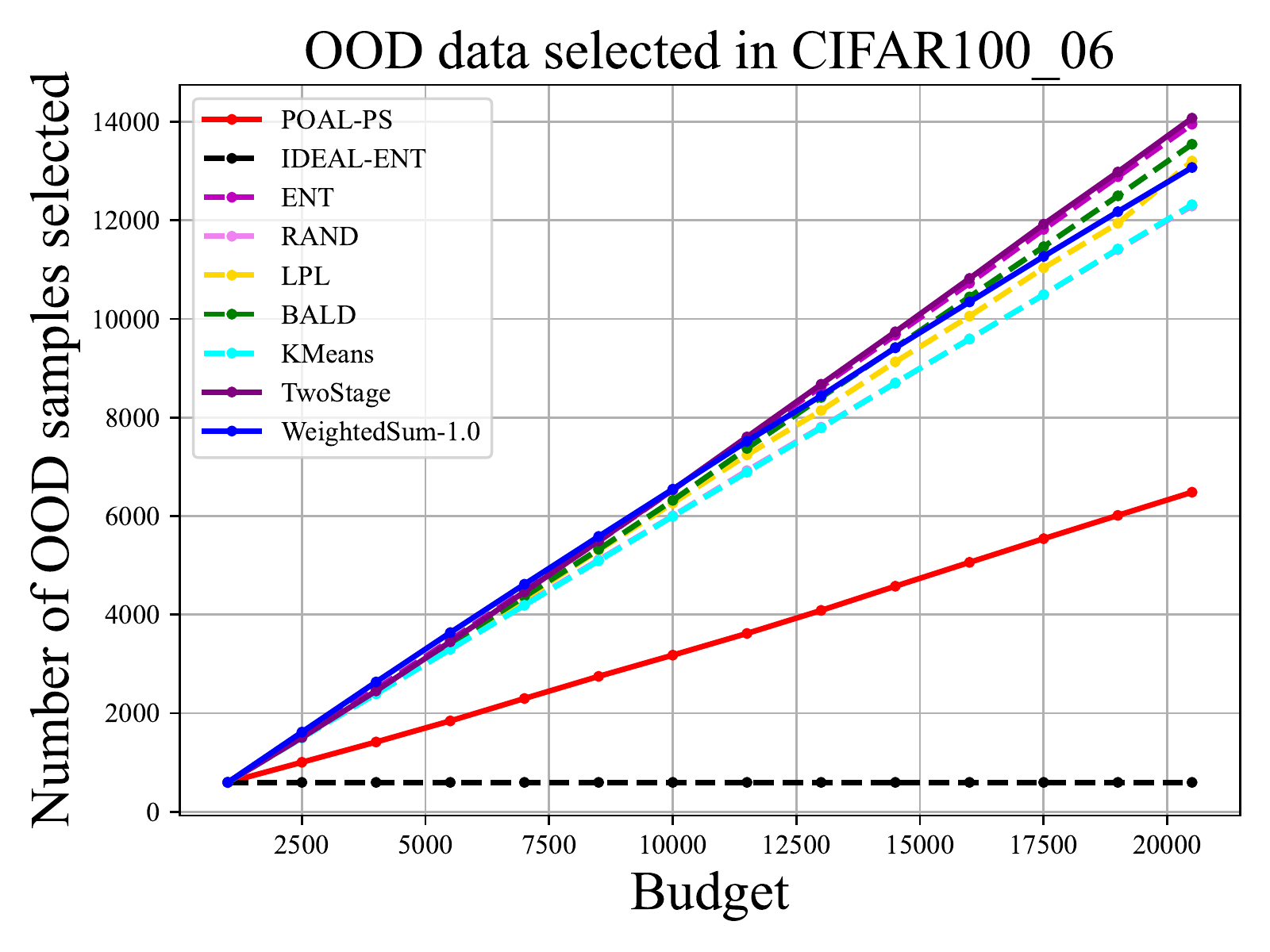}}
\caption{Results of \textbf{POAL} with \textbf{P}re-\textbf{S}electing (\textbf{POAL-PS}) on DL datasets. (a-d) are accuracy vs. budget curves. We present the mean and standard deviation (in parentehsis) of AUBC. (e-h) plot the number of OOD samples selected. More baseline comparisons like \citep{ash2020deep}, \citep{du2021contrastive} and \citep{kothawade2021similar} are shown in Appendix Figure A4.}
\label{dl}
\vspace{-0.5cm}
\end{figure*}

We next present experiments showing that our \textbf{POAL} also works on large-scale datasets together with deep AL.
%
The accuracy-budget curves for \emph{CIFAR10} and \emph{CIFAR100} with ID:OOD ratios 6:4 and 4:6, together with their number of OOD samples selected during the AL processes, are presented in Figure~\ref{dl}. 
\textbf{POAL} outperforms both uncertainty- and representative/diversity-based baselines like \textbf{LPL}, \textbf{BALD} and \textbf{$k$-Means} on all tasks. Although \textbf{LPL} performs fairly well on normal \emph{CIFAR10} and \emph{CIFAR100} datasets \citep{yoo2019learning, zhan2022comparative}, under OOD data scenarios, OOD samples have larger predicted loss values, and thus influence the selection decision. Similar phenomenon is observed for \textbf{BALD} and \textbf{ENT}. The performances of \textbf{$k$-Means} are close to \textbf{RAND}, since both  methods sampled ID/OOD data with the same ratio. Although \textbf{$k$-Means} will not select OOD samples first, like uncertainty-based measures, it still failed to provide comparable performances. \textbf{POAL} performs well by selecting less OOD samples during the AL iterations, as shown in Figure~\ref{dl}(e-h). Although the pre-selecting strategy in \textbf{POAL} makes the search space smaller, i.e., changes the global solutions to local solution, \textbf{POAL} still performs significantly better than baseline AL methods.
\abcn{Finally, \textbf{POAL} outperforms the other combination strategies, \textbf{WeightedSum} and \textbf{TwoStage}, by large margins.}


\section{Conclusion, Discussion and Future Work}
\label{conclusion}
\vspace{-0.3cm}
In this paper, we proposed an effective and flexible Monte-Carlo \textbf{POAL} framework for improving AL when encountering OOD data. Our \textbf{POAL} framework is able to i) incorporate various AL sampling strategies; ii) handle large-scale datasets; iii) support batch-mode AL with fixed batch size; iv) prevent trade-off hyper-parameter tuning; and v) work well on both classical ML and DL tasks.
Our work has two limitations: i) 
for the ID confidence score calculation, 
Mahalanobis distance only performs well on distinct ID-/OOD- distributions and thus influence \textbf{POAL} performance;
ii) \zxy{we need many iterations for the Pareto set converge based on MC sampling (see Appendix A.3.2 for timing cost comparison).} 
Future work could improve OOD detection criterion by introducing OOD-specific feature representations and other techniques for better distinguishing ID/OOD distributions. Although MC sampling requires more iterations, compared with other sampling schemes like importance sampling and Metropolis-Hasting, it is efficient and keeps the method free from initialization problems. 
We will try to find more efficient sampling schemes while reducing the impact of initialization. 

\small
\bibliography{neurips_2022}
\bibliographystyle{plainnat}

\appendix

\section{Appendix}

\subsection{Methodology}

\subsubsection{Pseudo code of Monte-Carlo POAL framework.}
Our Monte-Carlo \textbf{POAL} with early stopping is detailed in  Algorithm~\ref{mcpoal}. The implementation of MMD is the RBF kernel with various bandwidths with open-source implementation\footnote{The implementation of ${\rm MMD}$ is: \url{https://github.com/easezyc/deep-transfer-learning/blob/master/MUDA/MFSAN/MFSAN_3src/mmd.py}}.

\begin{algorithm}[tb]
 \caption{Monte-Carlo \textbf{POAL} with early stopping under OOD data scenarios.}
 \label{mcpoal}
 \begin{algorithmic}[1]
  \REQUIRE Unlabeled data pool $\mathcal{D}_u$ with size $M$, criteria $\mathcal{U}(\mathbf{x})$ and $\mathcal{M}(\mathbf{x})$, batch size $b$, maximum repeat time $T$, population interval $p_\text{inv}$, sliding window size $s_w$ and final decision function $\mathcal{F}$.
  \ENSURE A subset $\mathcal{D}_s$ of $\mathcal{D}_u$ where $|\mathcal{D}_s| = b$.
  \STATE Let $\mathbf{s} = \{0\}^{M}$, $\mathcal{P} = \{\mathbf{s}\}$ and $t=1$.
  \FOR {$t$ in $1,...,T$}
   \STATE Generate a random solution $\mathbf{s}$ with condition $\sum s_i = b$.
   \IF {$\nexists\mathbf{z} \in \mathcal{P}$ such that $\mathbf{s} \prec \mathbf{z}$}
   \STATE $Q = \{\mathbf{z} \in \mathcal{P} | \mathbf{z} \preceq \mathbf{s}\}$.
    \STATE $\mathcal{P} = (\mathcal{P} \setminus Q) \cup \{\mathbf{s}\}$.
   \ENDIF \\
   \COMMENT{\texttt{Terminate the loop early if the pareto set $\mathcal{P}$ converged.}}
   \IF {$t\ne 0$}
    \STATE Calculate ${\rm MMD}_t$ according to Equation 5.
    \IF {$\texttt{MOD}(t, p_\text{inv}) = 0$}
     \STATE Calculate $M_t$ and $S_t$ with interval $p_\text{inv}$ according to Equation 6 respectively.
     \STATE $\hat{M}_t = \texttt{ROUND}(M_t, 2)$, $\hat{S}_t = \texttt{ROUND}(S_t, 2)$.
     \IF {$\hat{M}_{t}=\hat{M}_{t-1}=...=\hat{M}_{t-s_w}$ and $\hat{S}_{t}=\hat{S}_{t-1}=...=\hat{S}_{t-s_w}$}
      \STATE \textbf{Break}
     \ENDIF
    \ENDIF
   \ENDIF
  \ENDFOR
  \RETURN $\arg\max_{\mathbf{s} \in \mathcal{P}} \mathcal{F}(\mathbf{s})$.
 \end{algorithmic}
\end{algorithm}

\subsubsection{Pseudo code of pre-selecting technique.}
The detailed pre-selecting process is presented in Algorithm~\ref{pre-select}.

\begin{algorithm}[htb]
\caption{Pre-selecting technique on large-scale datasets.}
\label{pre-select}
\begin{algorithmic}[1]
\REQUIRE Unlabeled data pool $\D_u$,  criteria $\mathcal{U}$ and $\mathcal{M}$, minimum size of subset $S_m$ of pre-selection.
\ENSURE A pre-selected subset $\D_{\text{sub}}$ of $\D_u$ with $|\D_{\text{sub}}| \geq S_m$.
\STATE Let $\D_{\text{sub}} = \emptyset$. $i = 0$.
\WHILE {$|\D_{\text{sub}}| < s_m$}
\STATE \dzy{Let $\D_{p} = \emptyset$.}
\FOR {$i$ in $0,..., |\D_u|$}
\IF {$\nexists \mathbf{x} \in \D_{p}$ such that $[(\mathcal{U}(\D^i_u) \leq \mathcal{U}(\mathbf{x})) \wedge (\mathcal{M}(\D^i_u) < \mathcal{M}(\mathbf{x}))]  \text{~or~} [(\mathcal{U}(\D^i_u) < \mathcal{U}(\mathbf{x})) \wedge (\mathcal{M}(\D^i_u) \leq \mathcal{M}(\mathbf{x}))]$}
\STATE $Q = \{\mathbf{x} \in \D_{p} | ~\ \mathcal{U}(\mathbf{x}) \leq \mathcal{U}(\D^i_u)  \wedge \mathcal{M}(\mathbf{x}) \leq \mathcal{M}(\D^i_u) \}$.
\STATE $\D_{p} = (\D_{p} \setminus Q) \cup \{\D^i_u$\}. 
\ENDIF   
\ENDFOR
\STATE \dzy{$\D_u = \D_u \setminus \D_{p}$.}
\STATE \dzy{$\D_{\text{sub}} = \D_{\text{sub}} \cup \D_{p}$.}
\ENDWHILE
\RETURN $\D_{\text{sub}}$. 
\end{algorithmic}
\end{algorithm}

\subsection{Experimental settings}

\subsubsection{Datasets}

We summarized the datasets we adopted in our experiments in Table~\ref{d}, including how to split the datasets (initial labeled data, unlabeled data pool and test set), dataset information (number of categories, number of feature dimensions) and task-concerned information (maximum budget, batch size, numbers of repeated trials and basic learners adopted in each task). Especially, we recorded the ID $:$ OOD ratio in each task. We down-sampled \emph{letter} dataset and control each category only have 50 data samples in training set.
 
\begin{figure*} [htb]
\centering
\subfloat[EX8]{\includegraphics[width=0.25\linewidth]{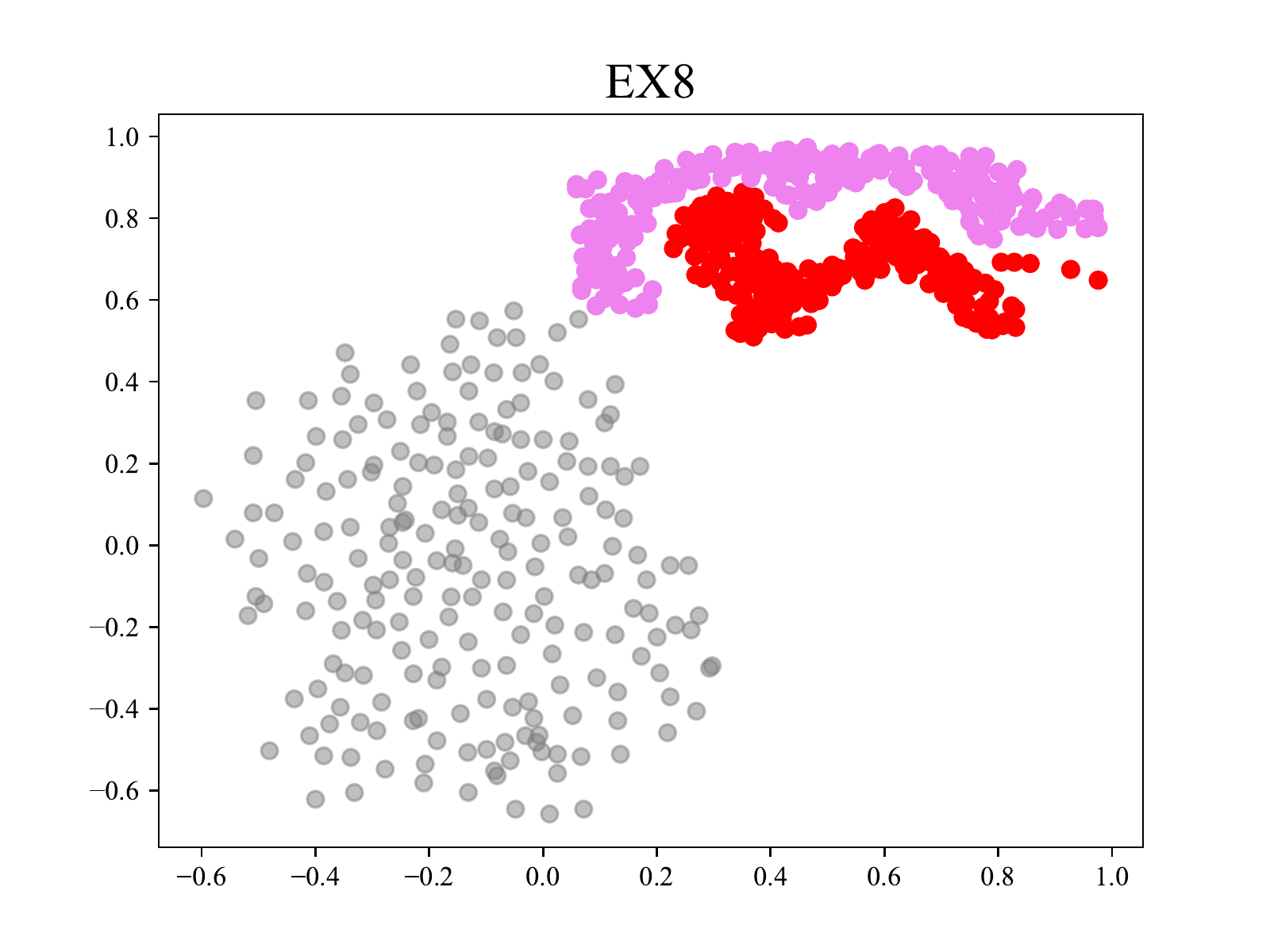}}
\subfloat[vowel]{\includegraphics[width=0.25\linewidth]{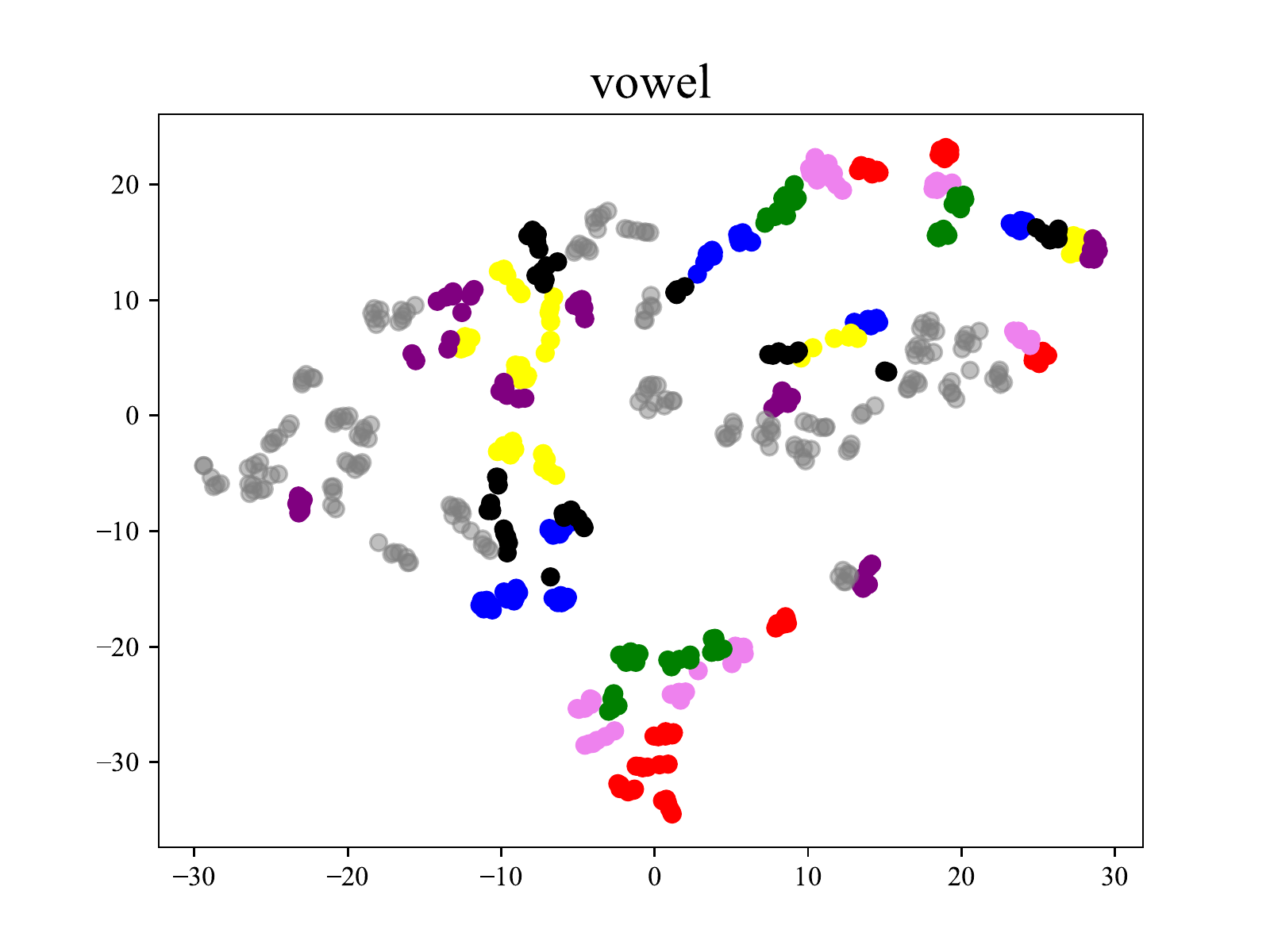}}
\subfloat[letter(a-k)]{\includegraphics[width=0.25\linewidth]{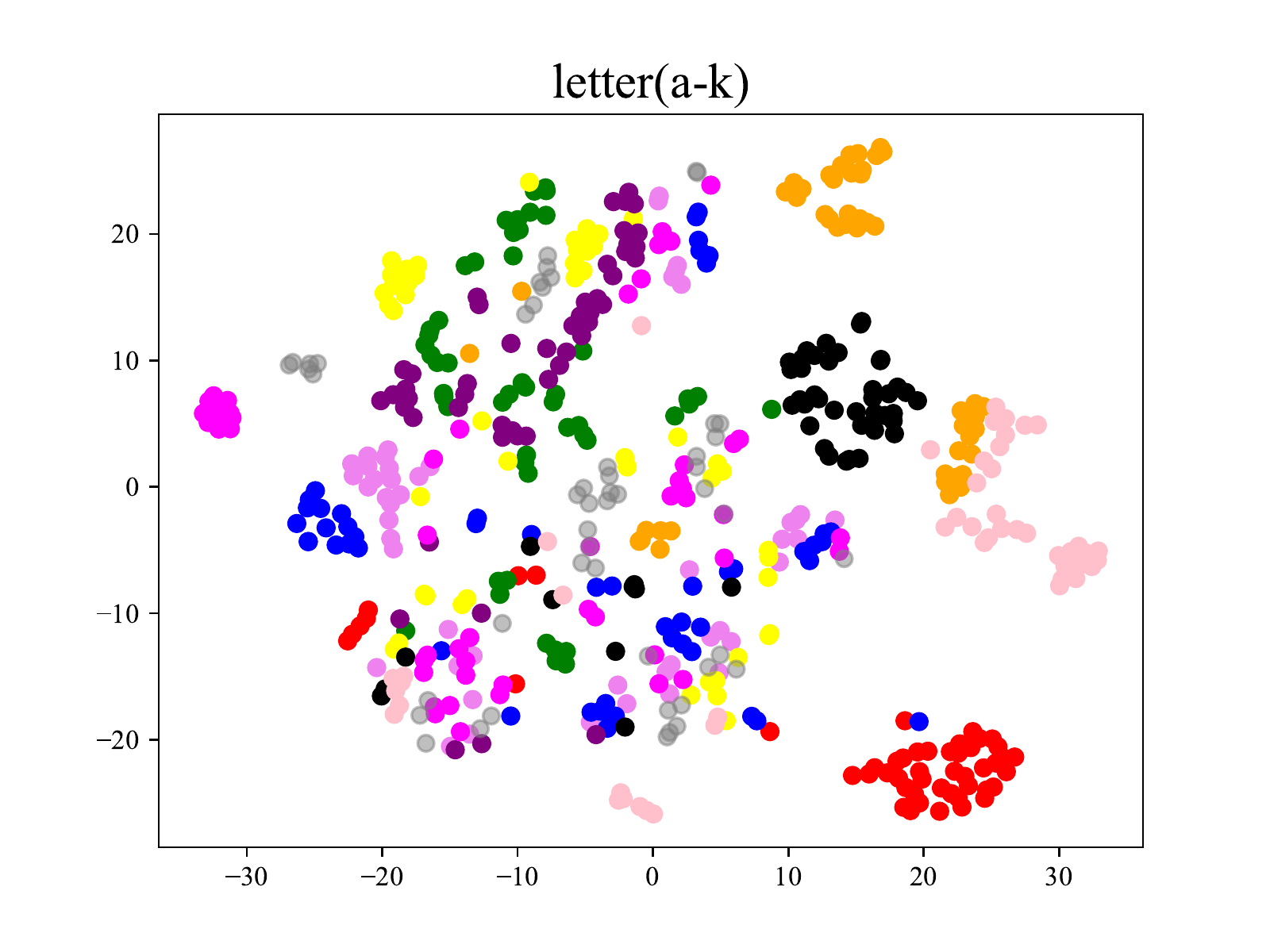}}
\subfloat[letter(a-l)]{\includegraphics[width=0.25\linewidth]{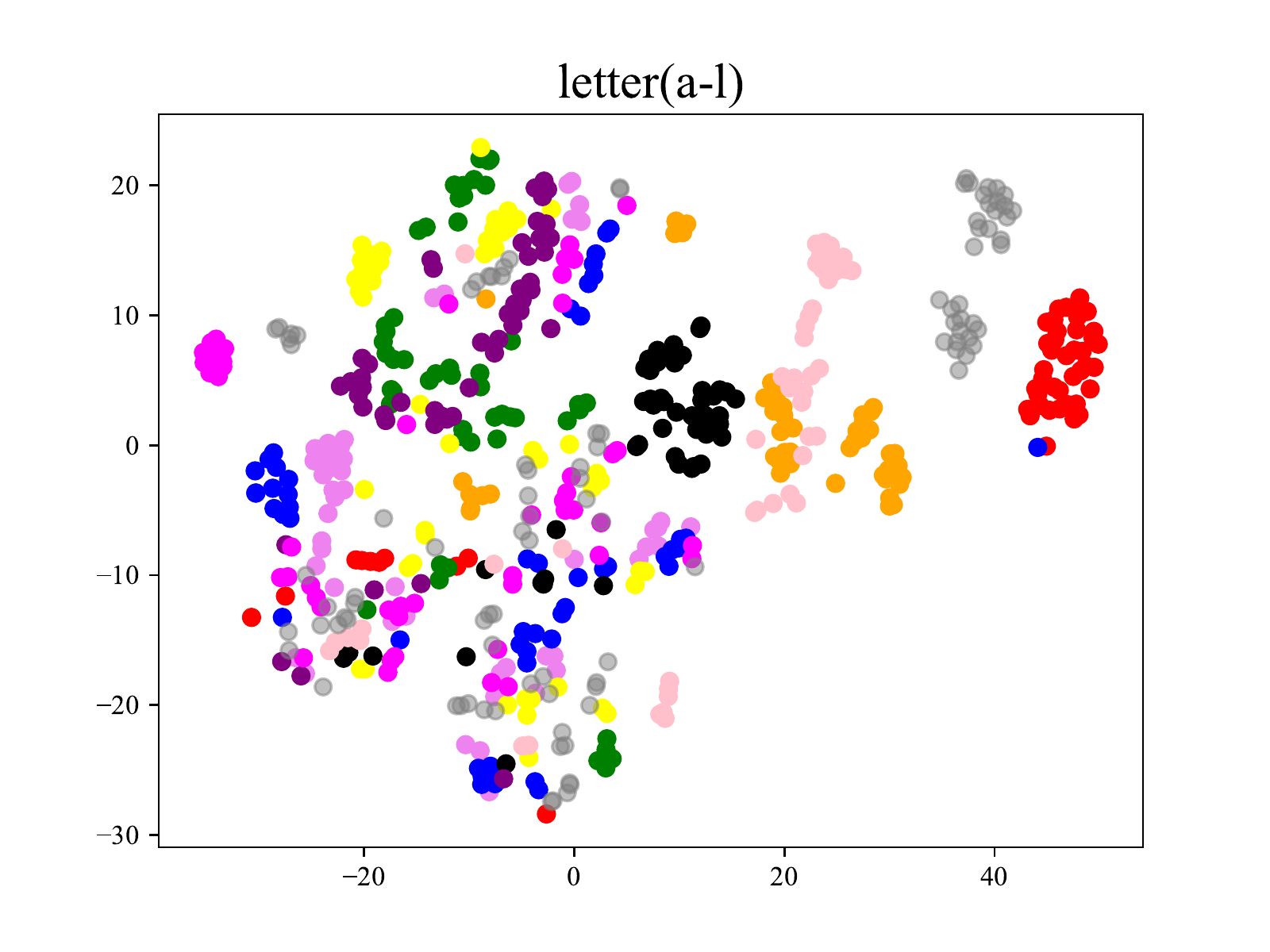}}

\subfloat[letter(a-m)]{\includegraphics[width=0.25\linewidth]{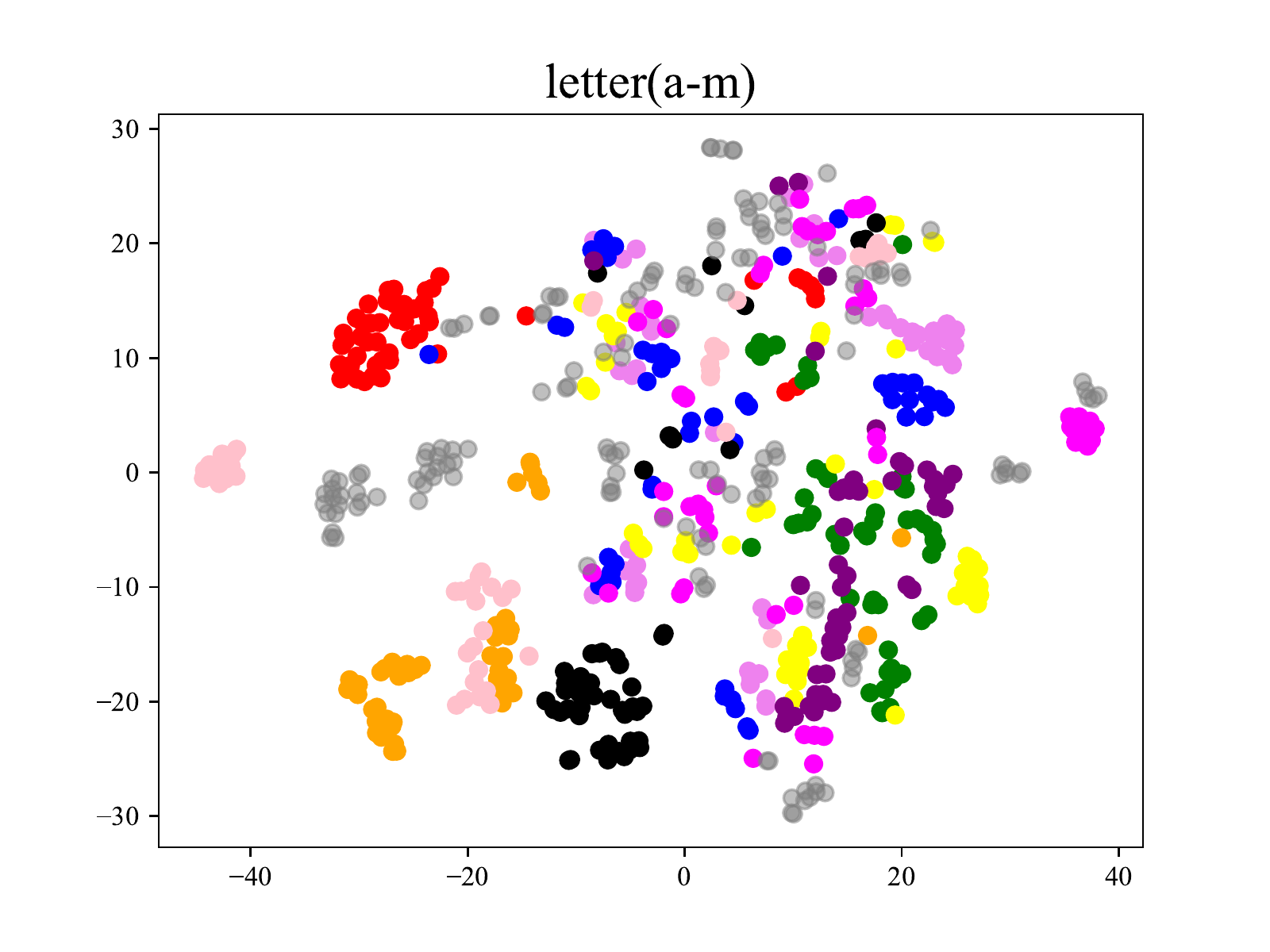}}
\subfloat[letter(a-n)]{\includegraphics[width=0.25\linewidth]{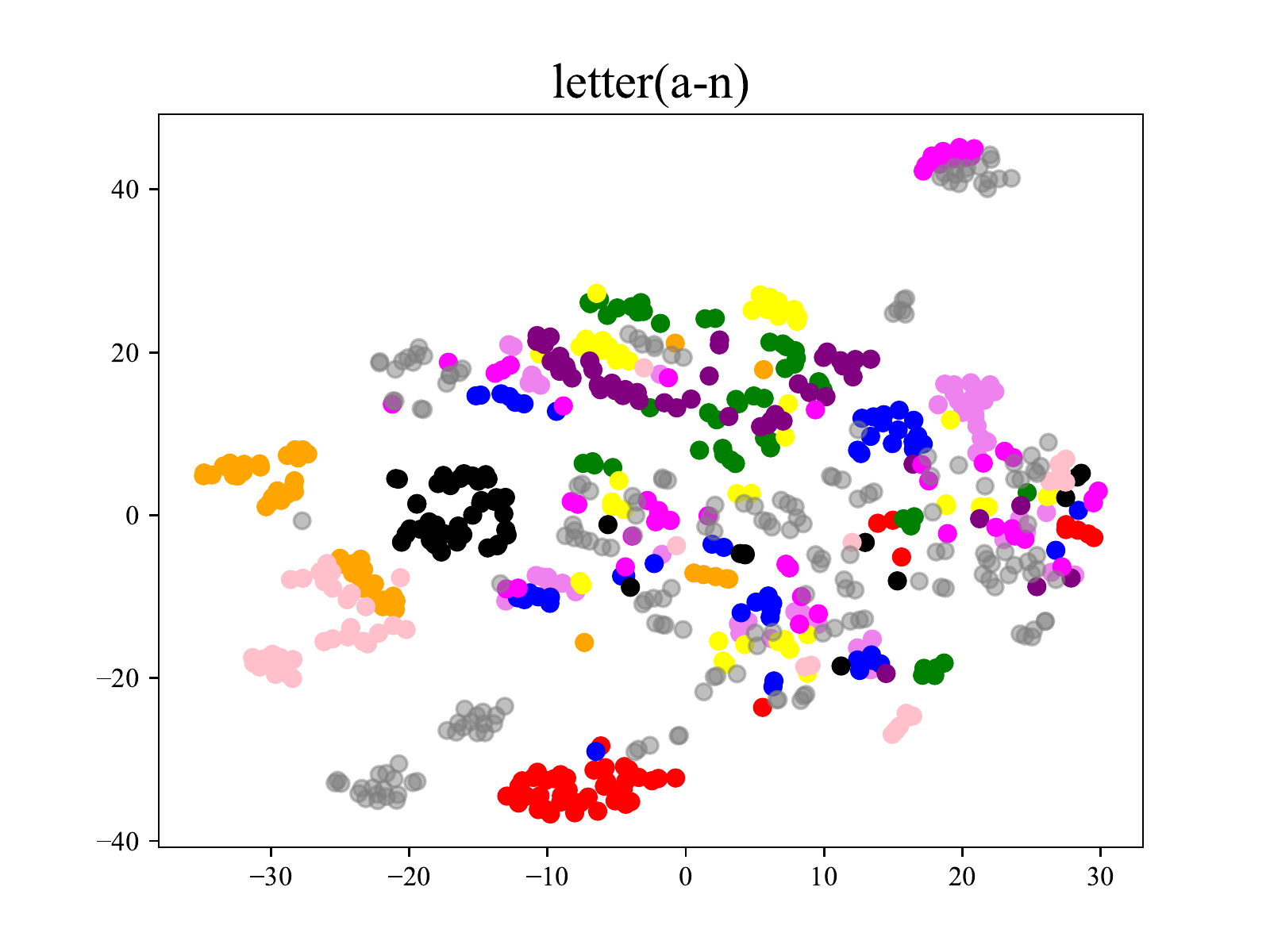}}
\subfloat[letter(a-o)]{\includegraphics[width=0.25\linewidth]{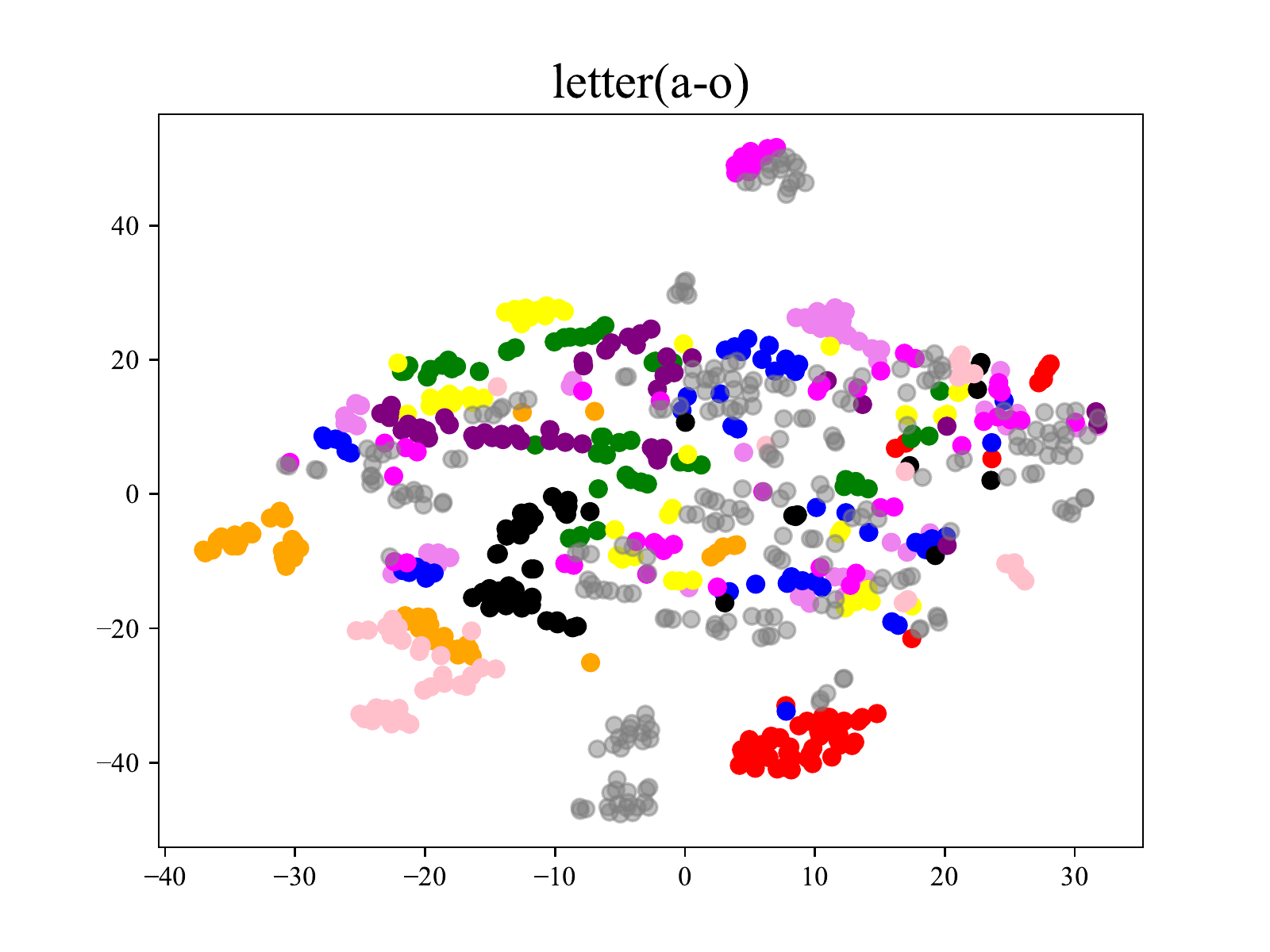}}
\subfloat[letter(a-p)]{\includegraphics[width=0.25\linewidth]{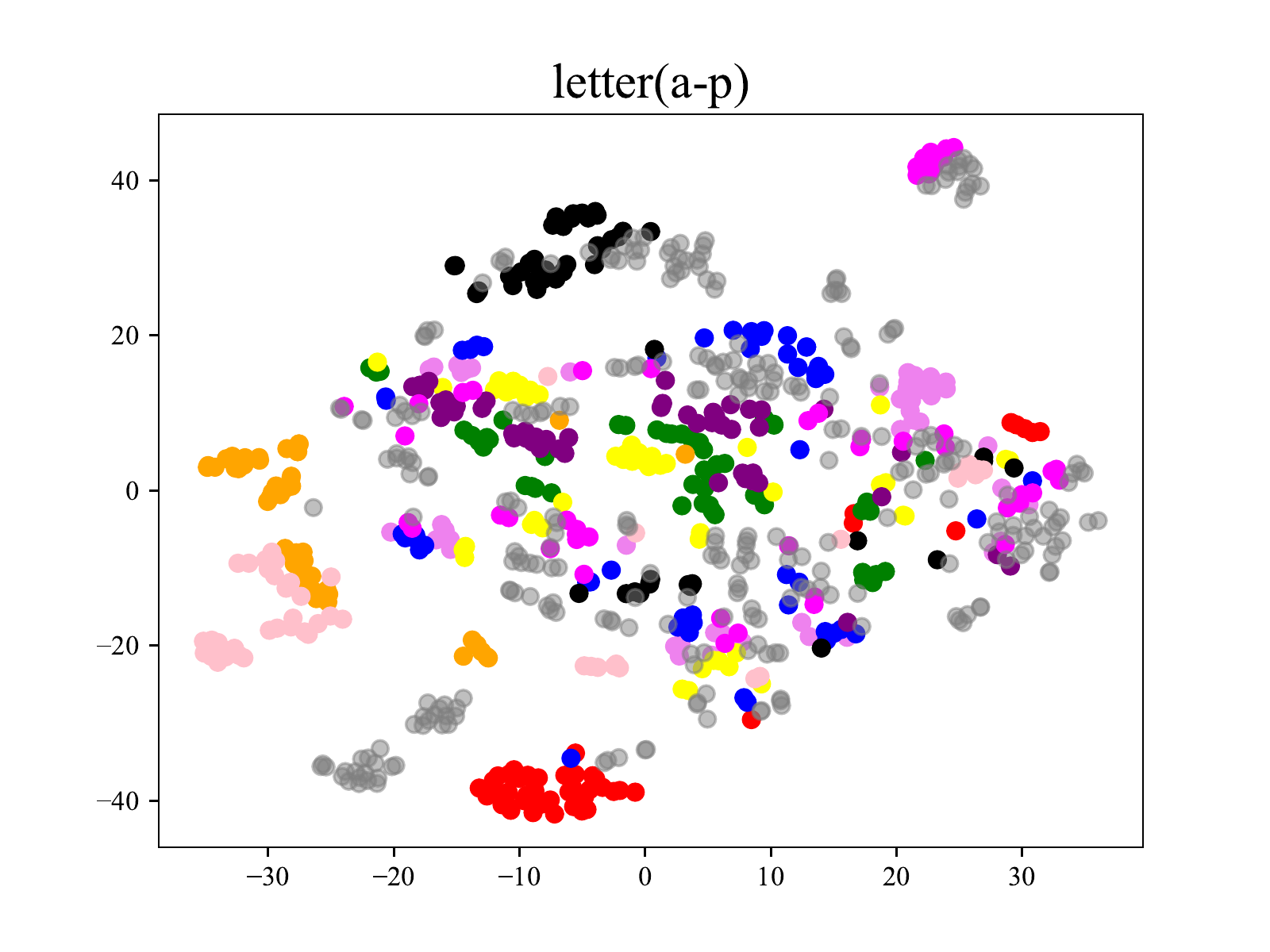}}

\subfloat[letter(a-q)]{\includegraphics[width=0.25\linewidth]{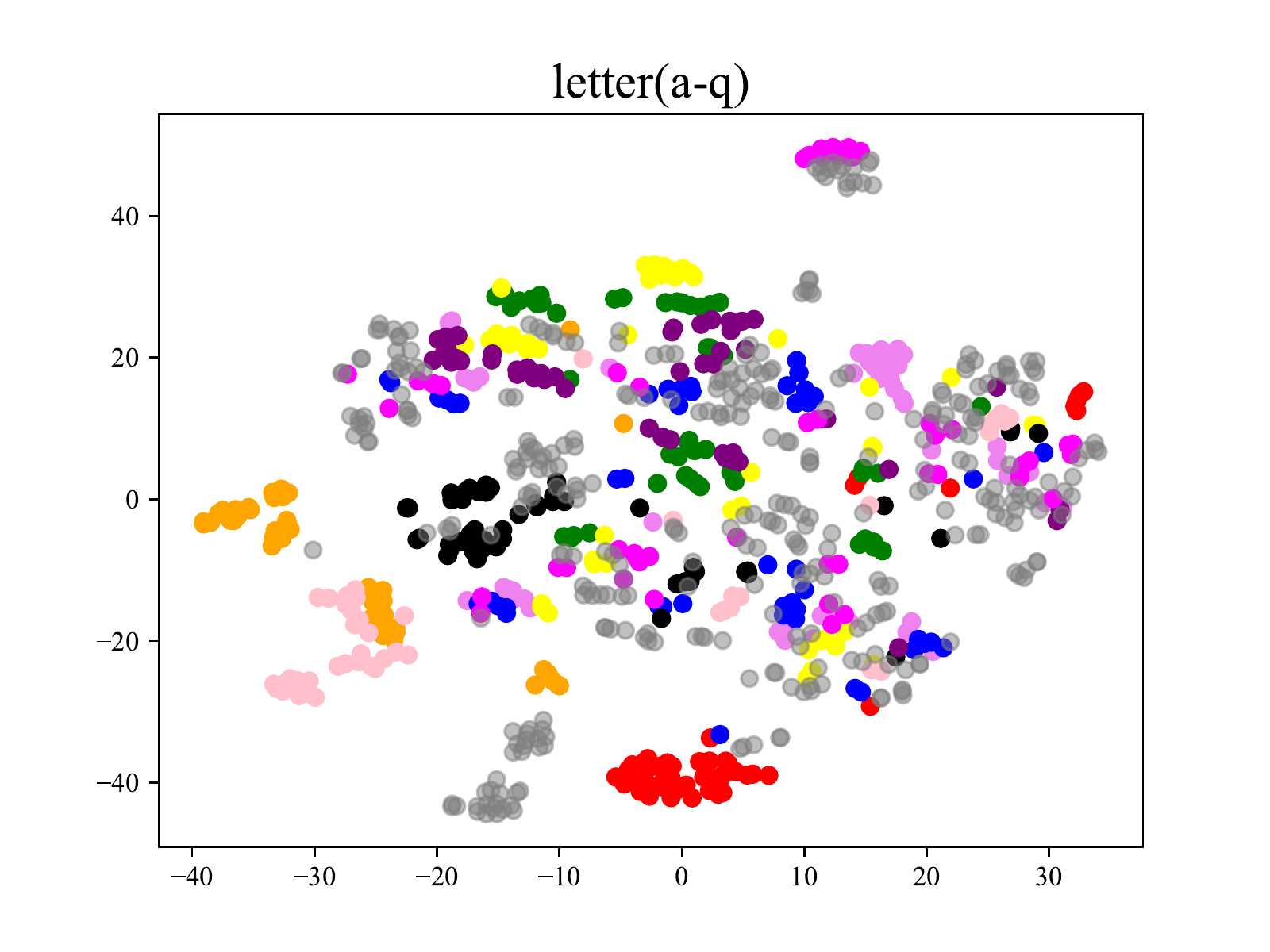}}
\subfloat[letter(a-r)]{\includegraphics[width=0.25\linewidth]{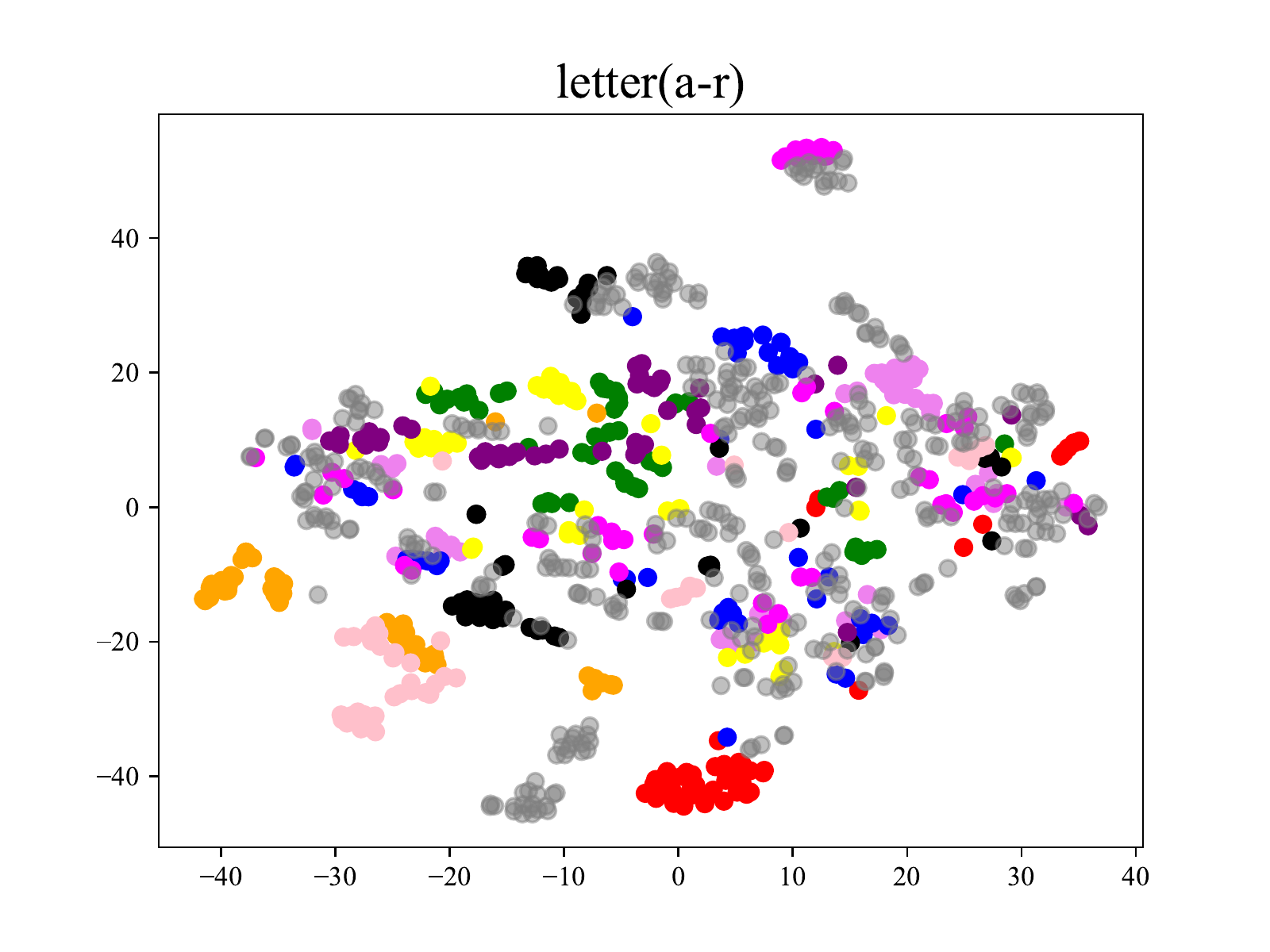}}
\subfloat[letter(a-s)]{\includegraphics[width=0.25\linewidth]{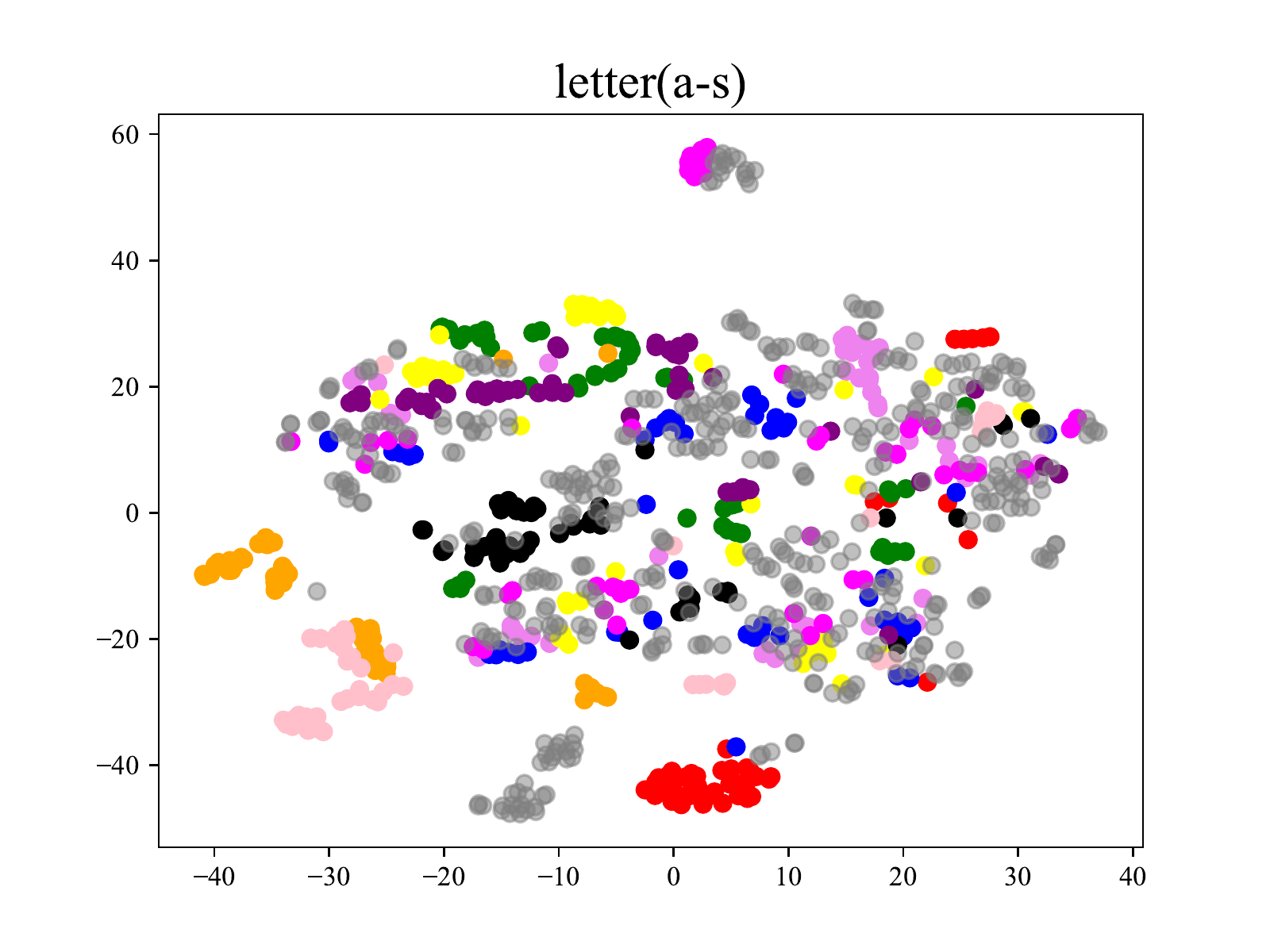}}
\subfloat[letter(a-t)]{\includegraphics[width=0.25\linewidth]{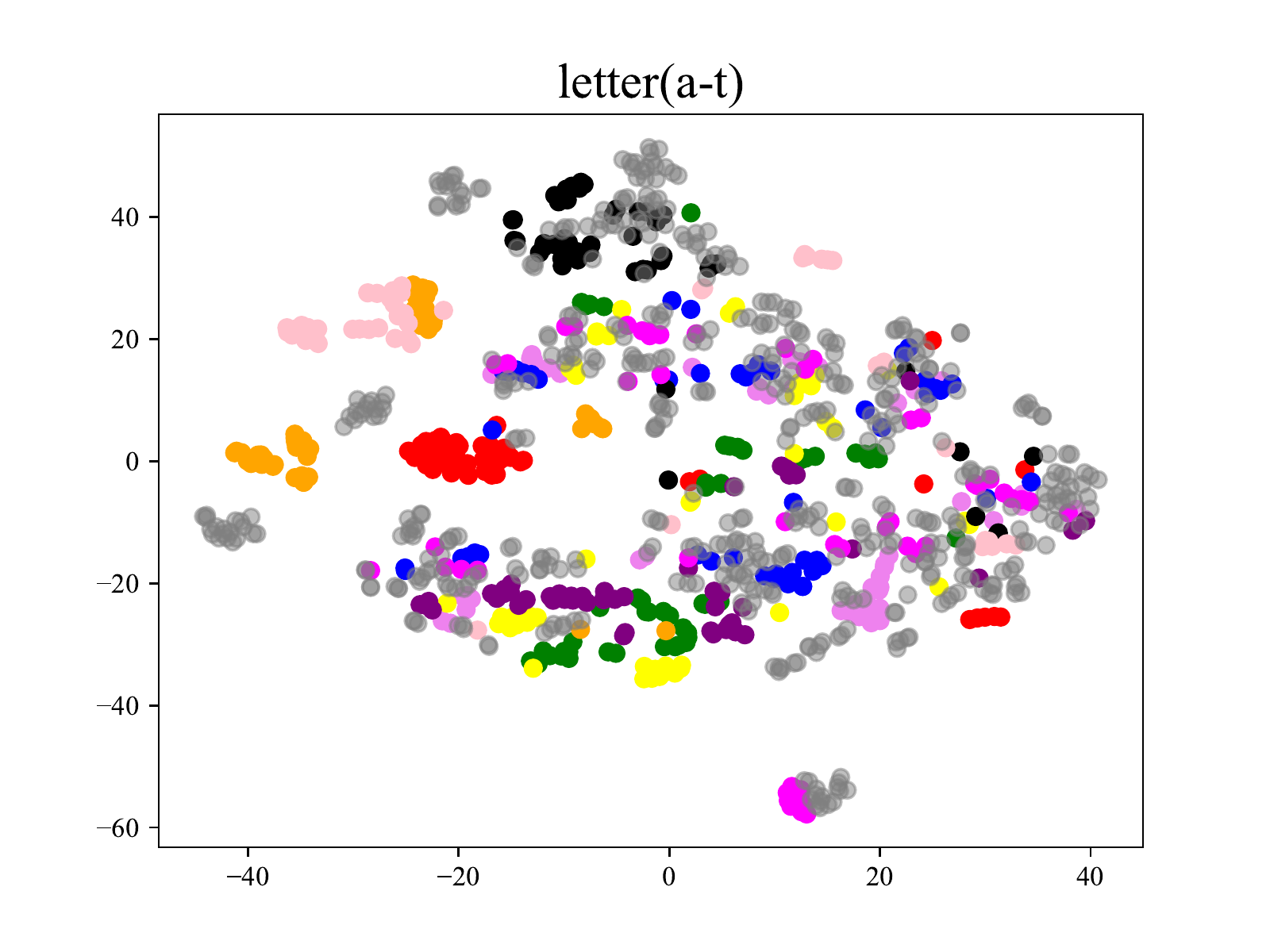}}

\subfloat[letter(a-u)]{\includegraphics[width=0.25\linewidth]{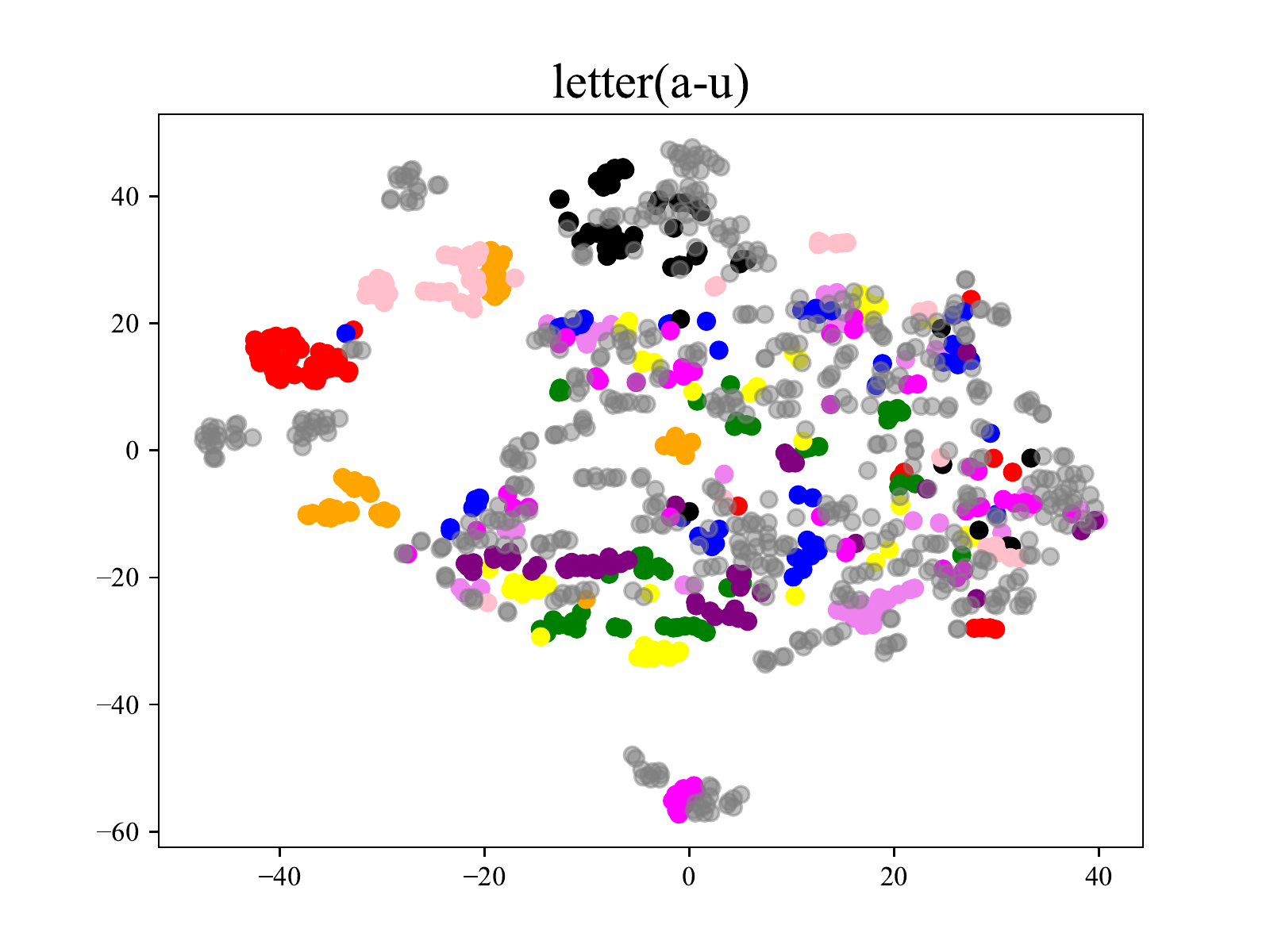}}
\subfloat[letter(a-v)]{\includegraphics[width=0.25\linewidth]{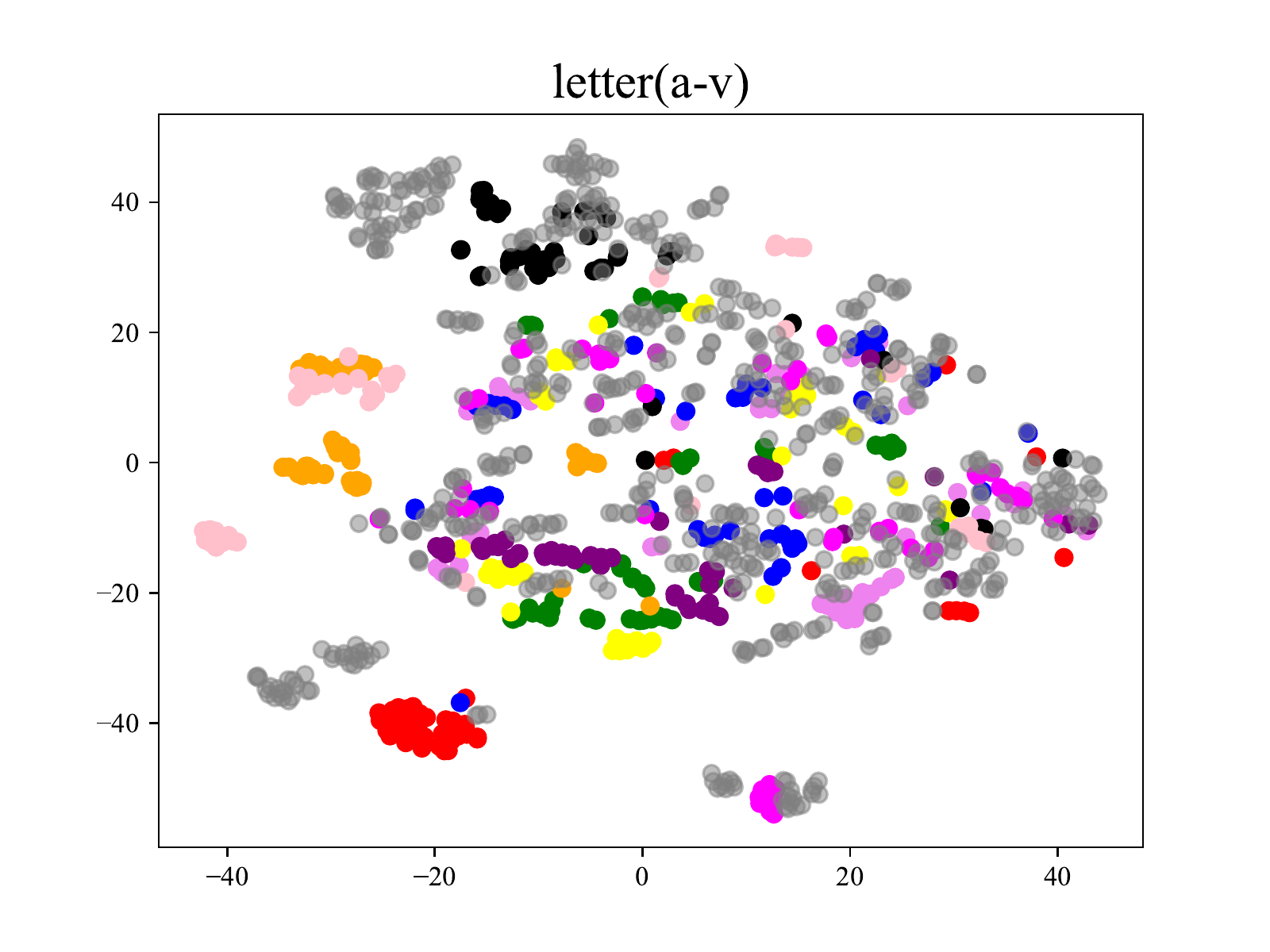}}
\subfloat[letter(a-w)]{\includegraphics[width=0.25\linewidth]{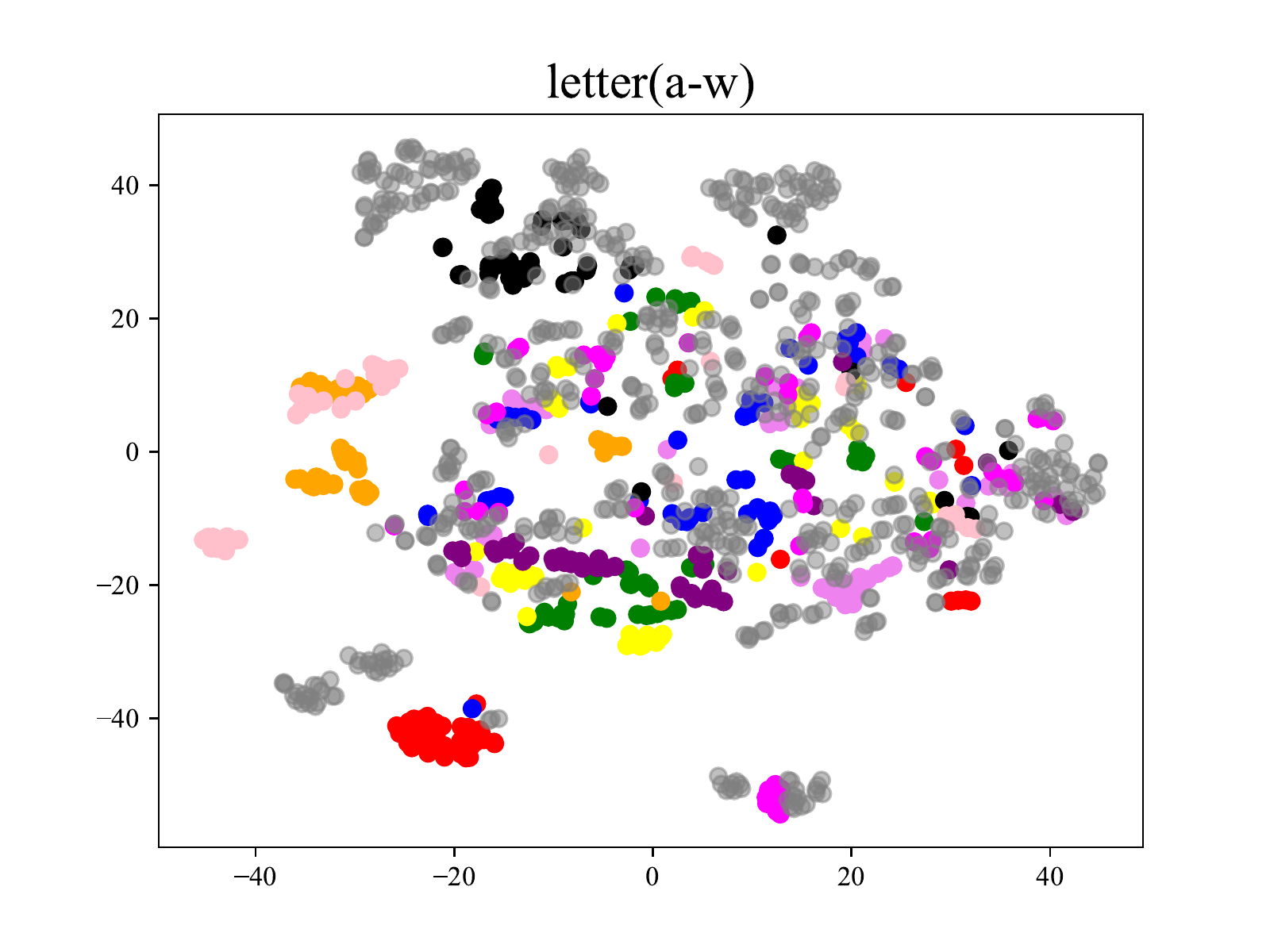}}
\subfloat[letter(a-x)]{\includegraphics[width=0.25\linewidth]{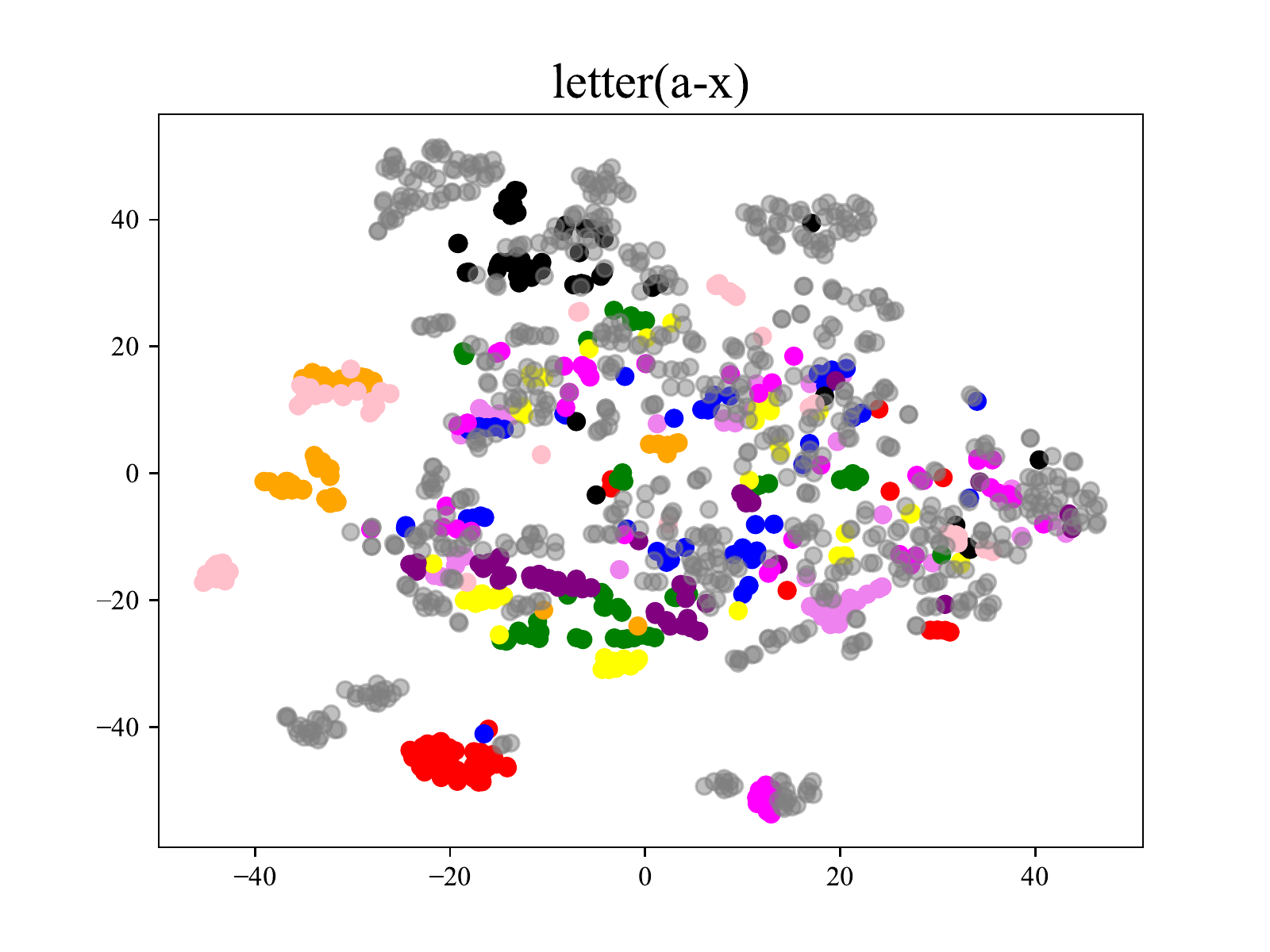}}

\subfloat[letter(a-y)]{\includegraphics[width=0.25\linewidth]{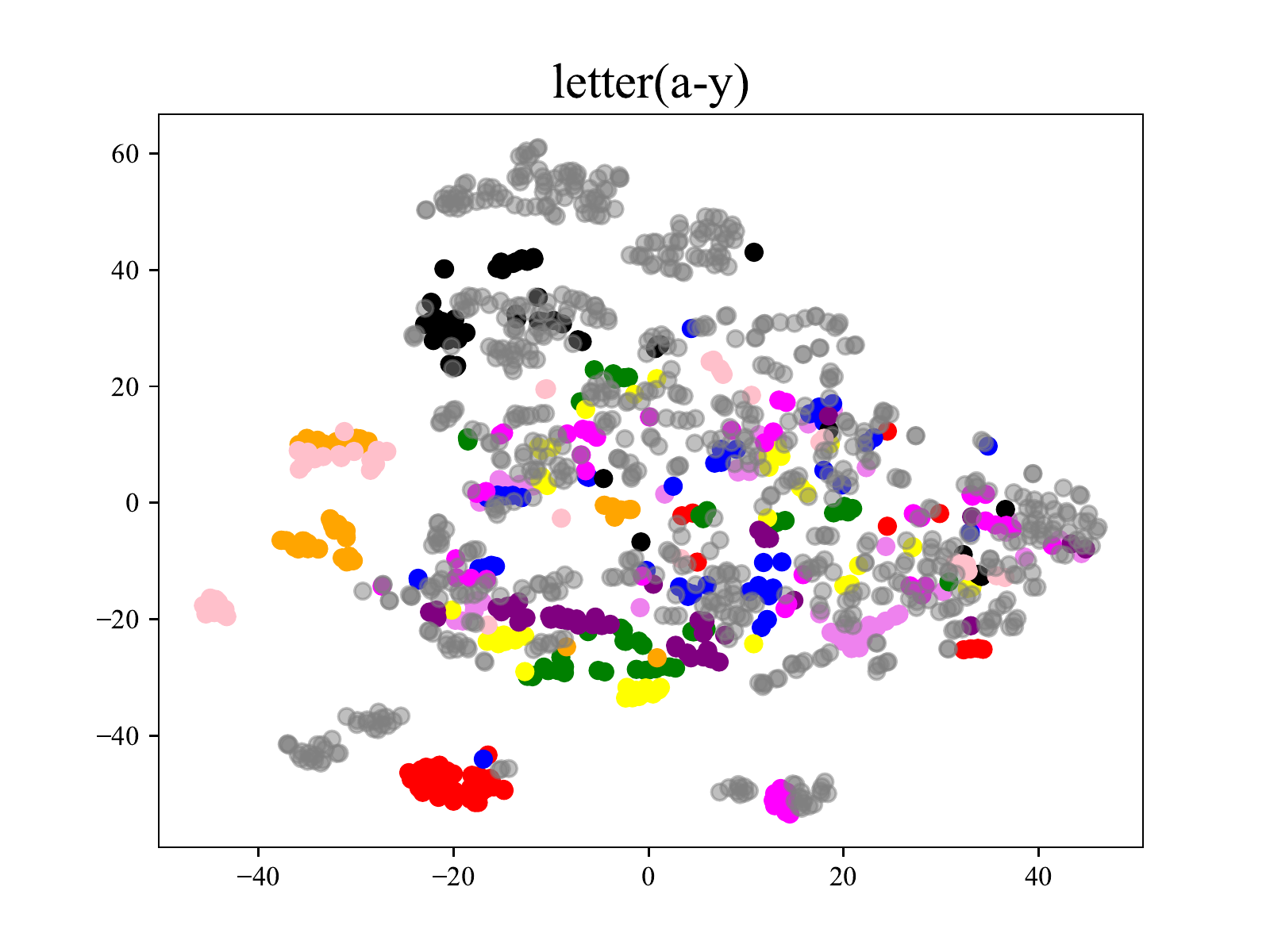}}
\subfloat[letter(a-z)]{\includegraphics[width=0.25\linewidth]{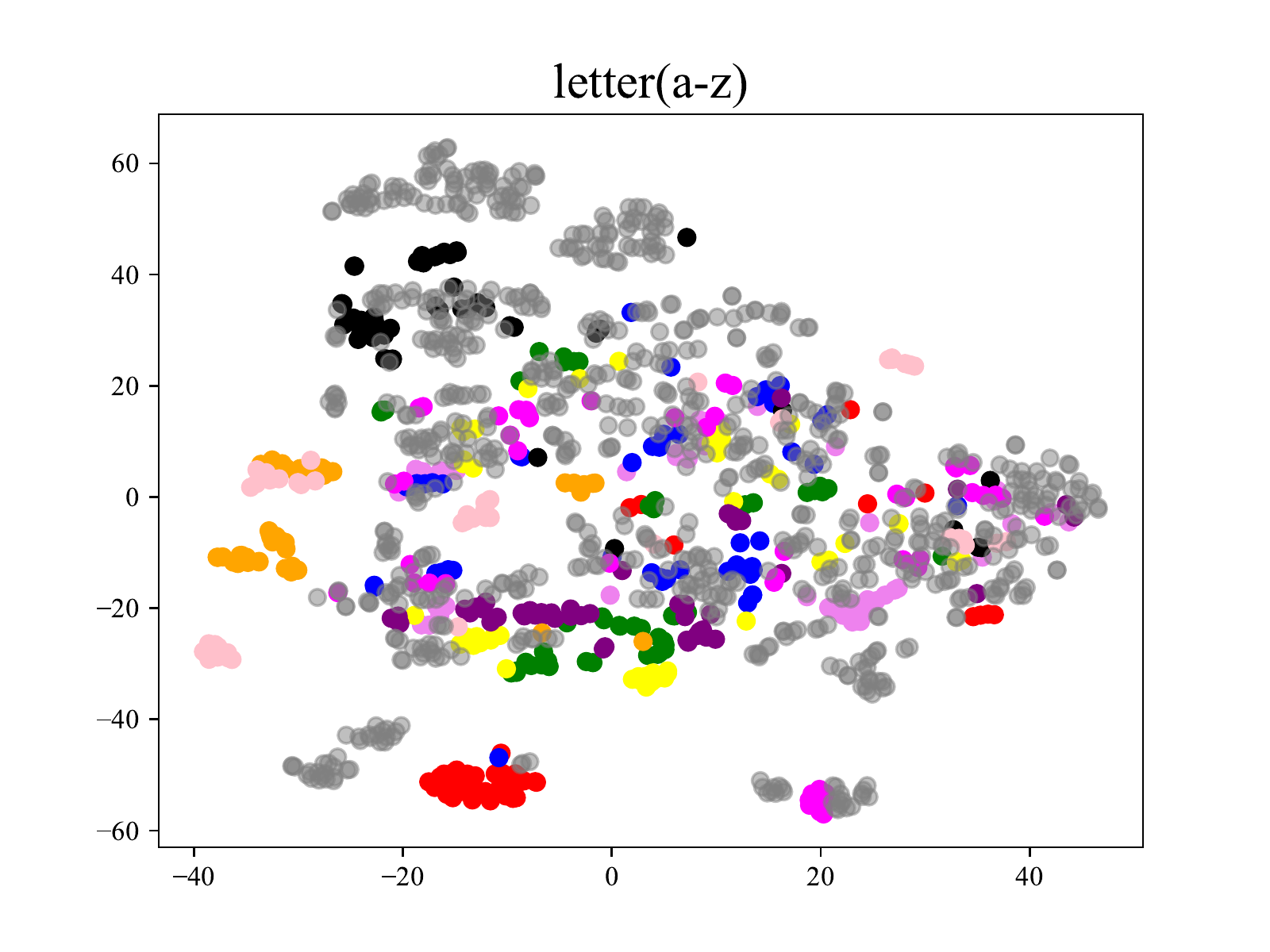}}
\caption{Visualization of data sets in classical ML tasks by t-SNE. We set the semitransparent grey dots to represent OOD data samples, the remaining colorful dots are ID data samples.}
\label{dataset}
\end{figure*}

We visualized the datasets of classical ML tasks by t-Distributed Stochastic Neighbor Embedding (t-SNE)\footnote{\url{https://scikit-learn.org/stable/modules/generated/sklearn.manifold.TSNE.html}}, as shown in Figure~\ref{dataset}. We split ID/OOD data by setting OOD data samples as  semitransparent grey dots to better observe the distinction between ID-/OOD- data distributions.

\begin{table}[htb]
\tiny
\centering
\caption{Datasets used in the experiments.}
\label{d}
{\begin{tabular}{@{}l|p{0.7cm}<\centering p{1.1cm}<\centering p{0.9cm}<\centering p{0.9cm}<\centering p{0.9cm}<\centering p{0.9cm}<\centering p{1.0cm}<\centering p{0.6cm}<\centering p{0.6cm}<\centering p{0.8cm}<\centering}
\hline
Dataset & $\#$ of ID classes & $\#$ of feature dimension & $\#$ of initial labeled set & $\#$  of unlabeled pool & $\#$  of test set & $\#$ of Maximum Budget &  $\text{ID}:\text{OOD}$ Ratio & batch size $b$ & $\#$ of repeat trials & basic learner\\
\hline
\emph{EX8} & 2 & 2 & 20 & 650 & 306 & 500 & $46:21$ & 10 & 100 & GPC \\
\hline
\emph{vowel} & 7 & 10 & 25 & 503 & 294 & 500 & $336:192$ & 10 & 100 & GPC \\
\emph{letter(a-k)} & 10 & 16 & 30 & 520 & 500 & 550 & $10:1$ & 10 & 100 & LR \\
\emph{letter(a-l)} & 10 & 16 & 30 & 570 & 500 & 550 & $10:2$ & 10 & 100 & LR \\
\emph{letter(a-m)} & 10 & 16 & 30 & 620 & 500 & 550 & $10:3$ & 10 & 100 & LR \\
\emph{letter(a-n)} & 10 & 16 & 30 & 670 & 500 & 550 & $10:4$ & 10 & 100 & LR \\
\emph{letter(a-o)} & 10 & 16 & 30 & 720 & 500 & 550 & $10:5$ & 10 & 100 & LR \\
\emph{letter(a-p)} & 10 & 16 & 30 & 770 & 500 & 550 & $10:6$ & 10 & 100 & LR \\
\emph{letter(a-q)} & 10 & 16 & 30 & 720 & 500 & 550 & $10:7$ & 10 & 100 & LR \\
\emph{letter(a-r)} & 10 & 16 & 30 & 870 & 500 & 550 & $10:8$ & 10 & 100 & LR \\
\emph{letter(a-s)} & 10 & 16 & 30 & 920 & 500 & 550 & $10:9$ & 10 & 100 & LR \\
\emph{letter(a-t)} & 10 & 16 & 30 & 970 & 500 & 550 & $10:10$ & 10 & 100 & LR \\
\emph{letter(a-u)} & 10 & 16 & 30 & 1020 & 500 & 550 & $10:11$ & 10 & 100 & LR \\
\emph{letter(a-v)} & 10 & 16 & 30 & 1070 & 500 & 550 & $10:12$ & 10 & 100 & LR \\
\emph{letter(a-w)} & 10 & 16 & 30 & 1120 & 500 & 550 & $10:13$ & 10 & 100 & LR \\
\emph{letter(a-x)} & 10 & 16 & 30 & 1170 & 500 & 550 & $10:14$ & 10 & 100 & LR \\
\emph{letter(a-y)} & 10 & 16 & 30 & 1220 & 500 & 550 & $10:15$ & 10 & 100 & LR \\
\emph{letter(a-z)} & 10 & 16 & 30 & 1270 & 500 & 550 & $10:16$ & 10 & 100 & LR \\
\hline
\emph{CIFAR10-04} & 6 & $32\times32\times3$ & 1000 & 49000 & 6000 & 20000 & $6:4$ & 500 & 3 & ResNet18 \\
\emph{CIFAR10-06} & 4 & $32\times32\times3$ & 1000 & 49000 & 4000 & 15000 & $4:6$ & 500 & 3 & ResNet18 \\
\emph{CIFAR100-04} & 60 & $32\times32\times3$ & 1000 & 49000 & 6000 & 25000 & $6:4$ & 500 & 3 & ResNet18 \\
\emph{CIFAR100-06} & 40 & $32\times32\times3$ & 1000 & 49000 & 4000 & 20000 & $4:6$ & 500 & 3 & ResNet18 \\
\hline
\end{tabular}}
\end{table}

\subsubsection{Baselines and implementations}

We introduce the baselines and how to make implementations in this section. 

For the basic learner/classifier, we adopted Gaussian Process Classifier (GPC), Logistic Regression (LR) and ResNet18 in our experiments. We didn't choose GPC in \emph{letter} dataset since the accuracy-budget curves based on GPC are not monotonically increasing. The requirement of the selection of a basic classifier is: the basic classifier can provide soft outputs, that is, provide predictive class probabilities to calculate the uncertainty of each unlabeled sample, e.g., entropy. We use the implementation of the GPC with RBF kernel\footnote{\url{https://scikit-learn.org/stable/modules/generated/sklearn.gaussian_process.GaussianProcessClassifier.html }} and LR\footnote{\url{https://scikit-learn.org/stable/modules/linear_model.html\#logistic-regression }} of scikit-learn library \citep{pedregosa2011scikit} with default settings. For ResNet18,  we employed ResNet18 \citep{he2016deep} based on PyTorch with Adam optimizer (learning rate: $1e-3$) as the basic learner in DL tasks. In \emph{CIFAR10} and \emph{CIFAR100} tasks, we set the number of training epochs as 30, the kernel size of the first convolutional layer in ResNet18 is $3 \times 3$ (consistent with PyTorch CIFAR implementation\footnote{\url{https://github.com/kuangliu/pytorch-cifar/blob/master/models/resnet.py}}). Input pre-processing steps include random crop ($\text{pad}=4$), random horizontal flip ($p = 0.5$) and normalization.

For the implementation of the classical ML baselines, we have introduced it in main paper.

For the baselines in DL tasks, including \textbf{$k$-Means}, \textbf{ENT}, \textbf{BALD}, \textbf{LPL} and \textbf{BADGE}. We implemented them based on PyTorch. About  \textbf{ENT}, \textbf{BALD}, and \textbf{RAND}, we use the implementation of DeepAL\footnote{\url{https://github.com/ej0cl6/deep-active-learning}}\citep{Huang2021deepal}. About \textbf{$k$-Means}, to facilitate the process, we use the faiss-gpu\footnote{\url{https://github.com/facebookresearch/faiss/tree/26abede81245800c026bd96a386c51430c1f4170}} \citep{johnson2019billion} version instead of scikit-learn version in \citep{Huang2021deepal}. For \textbf{LPL} \citep{yoo2019learning}, we adopted a re-implementation version\footnote{\url{https://github.com/Mephisto405/Learning-Loss-for-Active-Learning}}. For \textbf{BADGE} \citep{ash2020deep}, we adopted the original version provided the author\footnote{\url{https://github.com/JordanAsh/badge}}. We provide simple introductions of \textbf{BALD}, \textbf{LPL} and \textbf{BADGE} as follows:

\begin{compactitem}
\item Bayesian Active Learning by Disagreements (\textbf{BALD}) \citep{houlsby2011bayesian, gal2017deep}: it chooses the data points that are expected to maximize the information gained from the model parameters, i.e. the mutual information between predictions and model posterior.
\item Loss Prediction Loss (\textbf{LPL}) \citep{yoo2019learning}: it is a loss prediction strategy by attaching a small parametric module that is trained to predict the loss of unlabeled inputs with respect to the target model, by minimizing the loss prediction loss between predicted loss and target loss. \textbf{LPL} picks the top $b$ data samples with highest predicted loss. 
\item  Batch Active learning by Diverse Gradient Embeddings (\textbf{BADGE}) \citep{ash2020deep}: it first measures uncertainty as the gradient magnitude with respect to the parameters in the output layer in the first stage, it then clusters the samples by \textbf{$k$-Means++} in the second stage.
\end{compactitem}

We run all our experiments on a single Tesla V100-SXM2 GPU with 16GB memory except for running \textbf{SIMILAR} related experiments. Since  \textbf{SIMILAR} needs much memory. We run the experiments ( \textbf{SIMILAR} on down-sampled \emph{CIFAR10}) on another workstation of Tesla V100-SXM2 GPU with 94GB memory in total.

\subsubsection{Licences of datasets and methods}
\label{licences}
\paragraph{Datasets.}
We listed the licence of datasets we used in our experiments:
\begin{compactitem}
    \item EX8a and EX8b \citep{andrew2008stan}: Not listed.
    \item vowel \citep{asuncion2007uci, aggarwal2015theoretical, Dua2019}: Aucune licence fournie.
    \item letter \citep{frey1991letter, asuncion2007uci, Dua2019}: Not listed.
    \item CIFAR10 and CIFAR100 \citep{krizhevsky2009learning}: MIT Licence.  
\end{compactitem}

\paragraph{Methods.}
We listed the license of libraries we used in our experiments:
\begin{compactitem}
    \item PyTorch \citep{paszke2019pytorch}: Modified BSD.
    \item CCAL \citep{du2021contrastive}: Not listed.
    \item SCMI(DISTIL)\footnote{\url{https://decile-team-distil.readthedocs.io/en/latest/}} \citep{kothawade2021similar}: MIT Licence.
    \item BADGE \citep{ash2020deep}: Not listed.
    \item LPL \citep{yoo2019learning}: Not listed.
    \item ENT \citep{lewis1994heterogeneous}, MARGIN \citep{wang2014new}, BALD \citep{houlsby2011bayesian, gal2017deep}, use DeepAL \citep{Huang2021deepal} library implementation: MIT Licence.
    \item KMeans (faiss library \citep{johnson2019billion} implementation): MIT Licence.
    \item MAHA\footnote{\url{https://github.com/pokaxpoka/deep_Mahalanobis_detector}} \citep{lee2018simple}: Not listed.
    \item GMM, GPC, LR (scikit-learn library): BSD License.
    \item ResNet18 \citep{he2016deep}: MIT License.
    \item QBC \citep{seung1992query}, LAL \citep{konyushkova2017learning} (ALipy library\footnote{\url{https://github.com/NUAA-AL/ALiPy}} \citep{TLHalipy}): BSD 3-Clause "New" or "Revised" License.

\end{compactitem}

\subsection{Experimental results}

\subsubsection{Additional experimental results of classical ML tasks}
\paragraph{Additional results of experiments with classical ML tasks in main paper, including accuracy vs. budget curves and number of OOD samples selected vs. budge curves.} We present the complete accuracy vs. budget curves and numbers of OOD samples selected vs. budget curves during AL processes in Figure~\ref{classical_ml} and Figure~\ref{classical_ml_ood} respectively. We could observe from Figure~\ref{dataset}, Figure~\ref{classical_ml} and Figure~\ref{classical_ml_ood} that the capability of Mahalanobis distance is limited by the distinction between ID data distribution and OOD data distribution. \emph{EX8} has distinct ID/OOD data distributions (see Figure~\ref{dataset}-a), thus the Mahalanobis distance well distinguishes ID and OOD data, and reaches the optimal performance (Figure~\ref{classical_ml_ood}-a, \textbf{MAHA} has the same curve with \textbf{IDEAL-ENT}). However, for \emph{vowel} and \emph{letter} dataset, the ID/OOD data distributions are not as distinct as \emph{EX8} (see  Figure~\ref{dataset} b-r), thus the performance of \textbf{MAHA} is influenced. Furthermore, it affects the performance of our \textbf{POAL}. It makes our \textbf{POAL} behaves less perfectly on \emph{vowel} and \emph{letter} datasets than on the \emph{EX8} dataset. Similar conclusions appear in \citep{ren2019likelihood}. This is our future work, to find better ways to differentiate ID and OOD data distributions.

\paragraph{Flexibility: POAL with more sampling strategies, i.e., LAL and QBC.}
To present the flexibility of our \textbf{POAL}, e.g., is able to incorporate more AL sampling schemes, we conduct a simple experiment on classical ML tasks, we incorporate \textbf{QBC} \citep{seung1992query} and \textbf{LAL} \citep{konyushkova2017learning}. The selection of basic AL sampling strategies is according to the comparative survey \citep{zhan2021comp}, in which both \textbf{QBC} and \textbf{LAL} show competitive performance across multiple tasks and AL sampling strategies. We repeat the trials 10 times in each experiment. The results are presented in Figure~\ref{flexible}. We conduct experiments on \emph{EX8} and \emph{vowel} datasets. This experiment shows that our \textbf{POAL} is flexible to handle various AL sampling strategies.

\begin{figure*} [htb]
\centering
\subfloat[EX8]{\includegraphics[width=0.25\linewidth]{img/exp/classicalML/ex8ab_GPC_100.pdf}}
\subfloat[vowel]{\includegraphics[width=0.25\linewidth]{img/exp/classicalML/vowel_GPC_100.pdf}}
\subfloat[letter(a-k)]{\includegraphics[width=0.25\linewidth]{img/exp/classicalML/letter_ak10_LR_fixedbudget.pdf}}
\subfloat[letter(a-l)]{\includegraphics[width=0.25\linewidth]{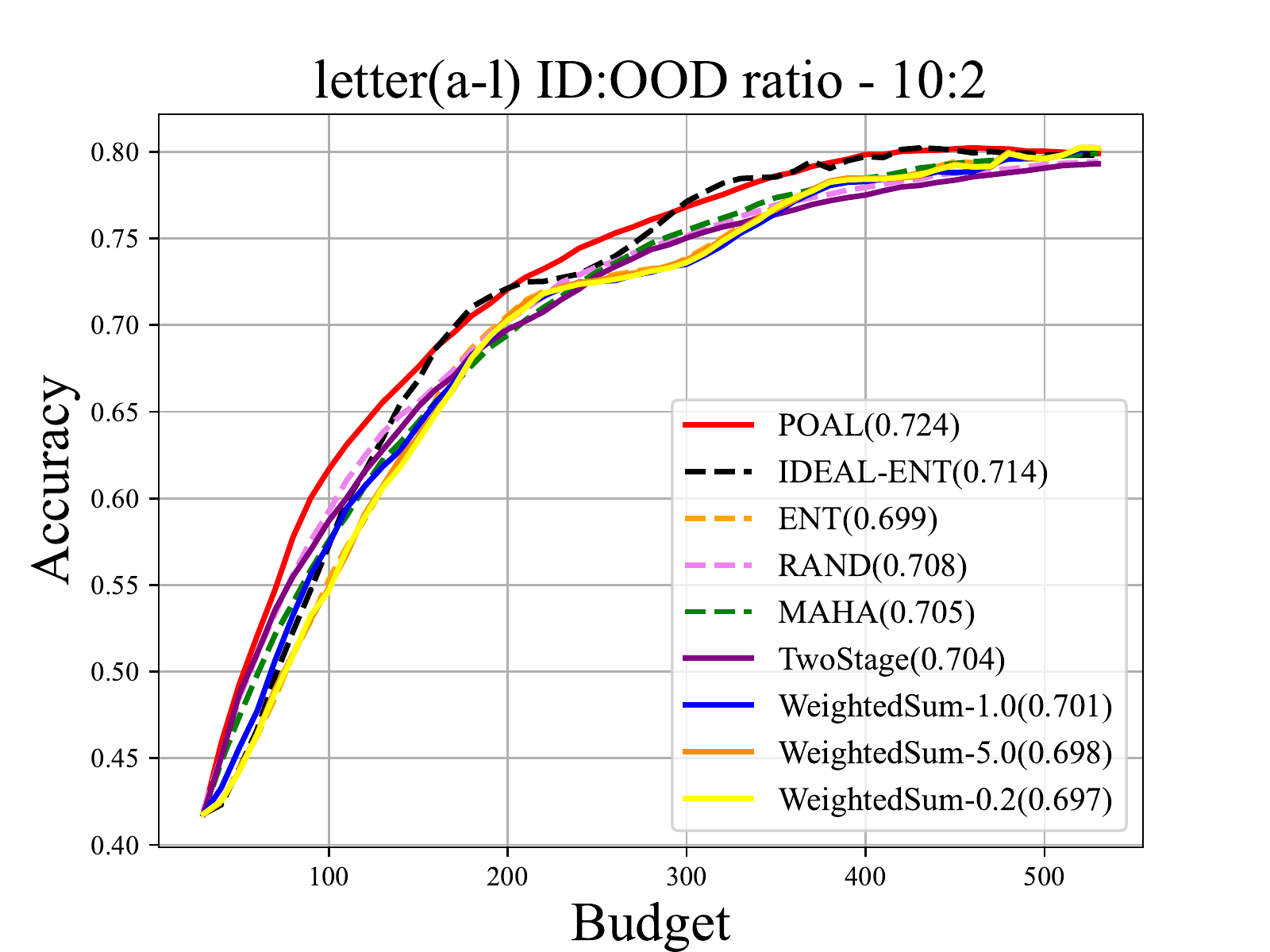}}

\subfloat[letter(a-n)]{\includegraphics[width=0.25\linewidth]{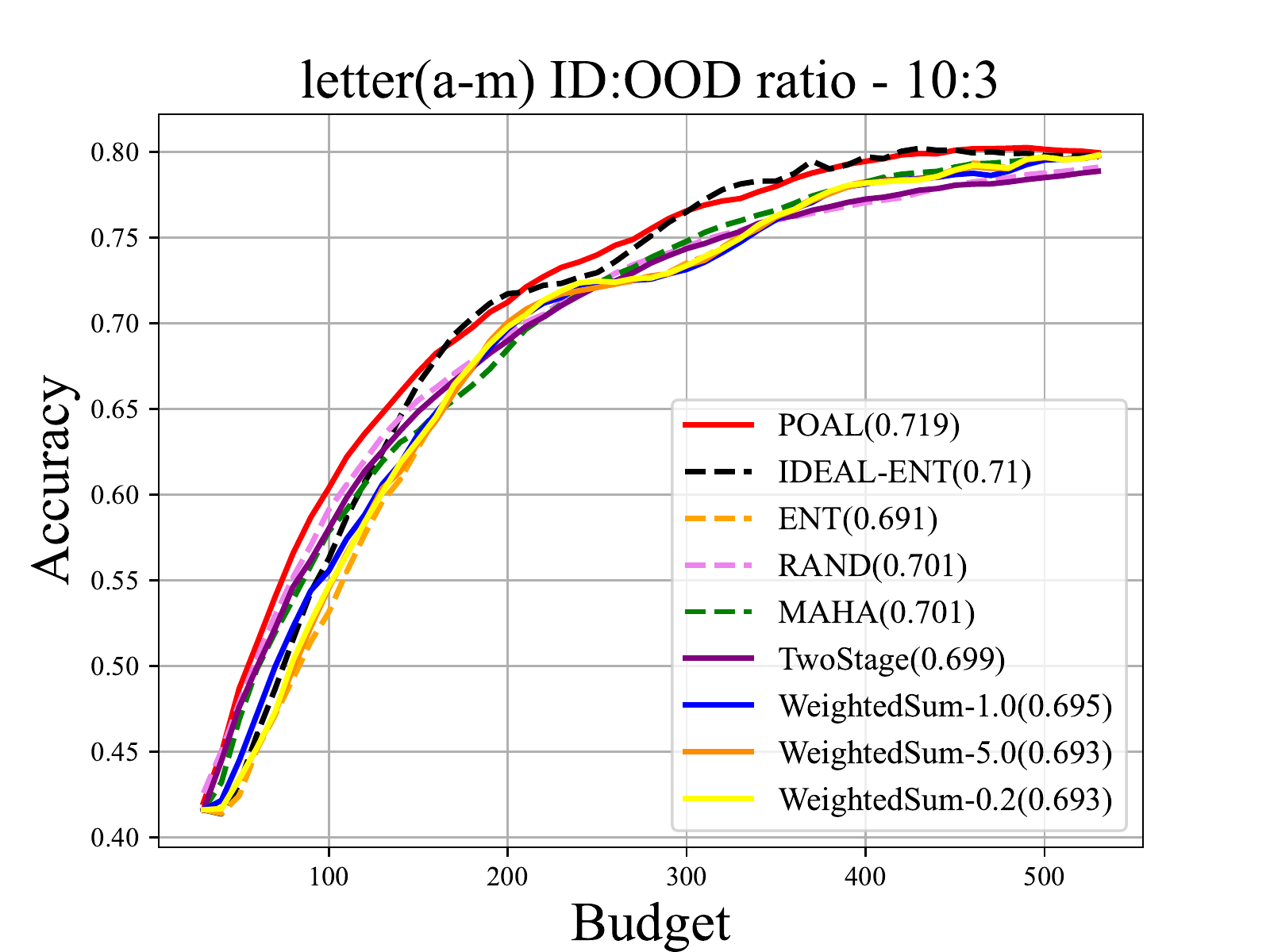}}
\subfloat[letter(a-m)]{\includegraphics[width=0.25\linewidth]{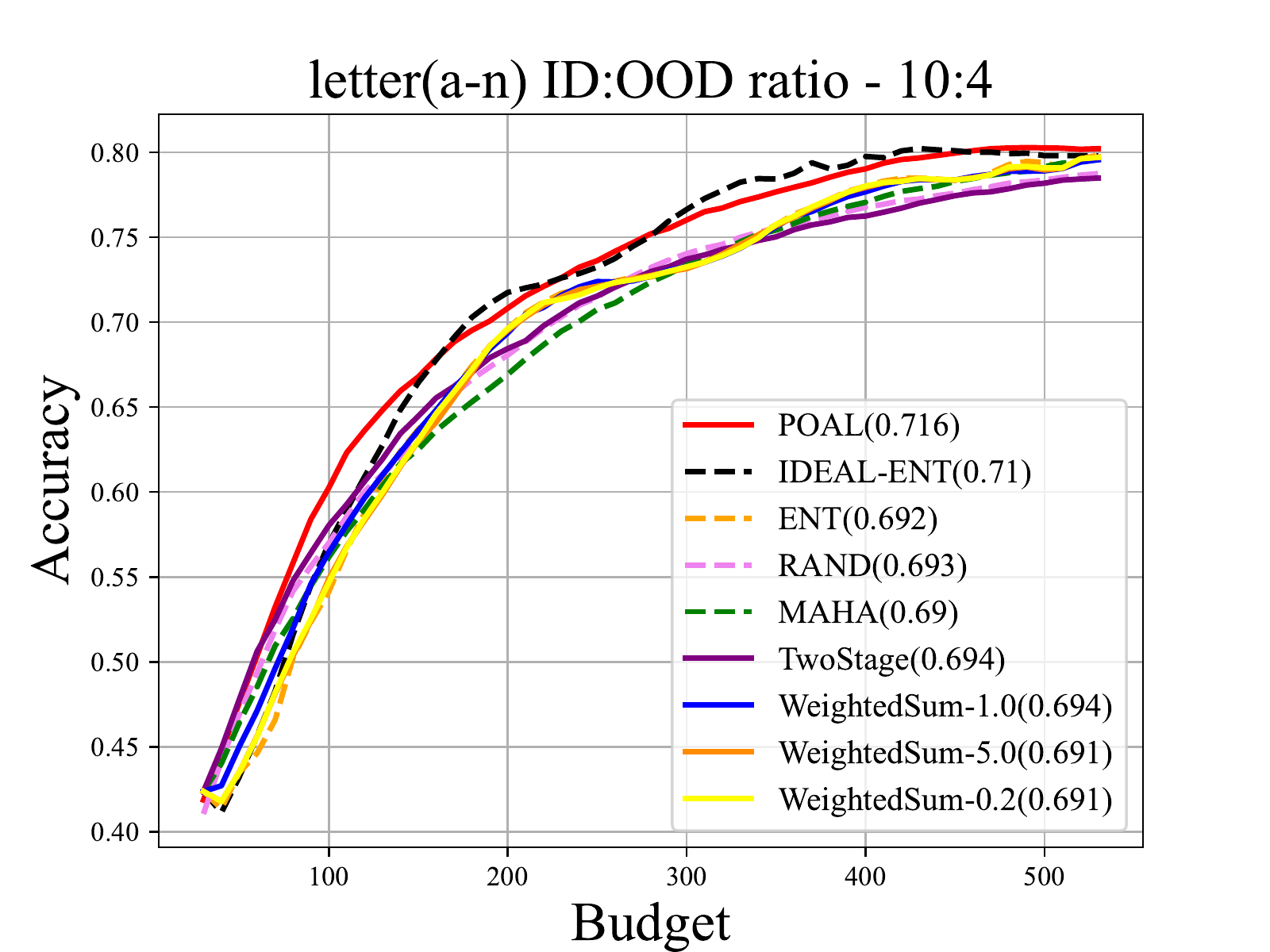}}
\subfloat[letter(a-o)]{\includegraphics[width=0.25\linewidth]{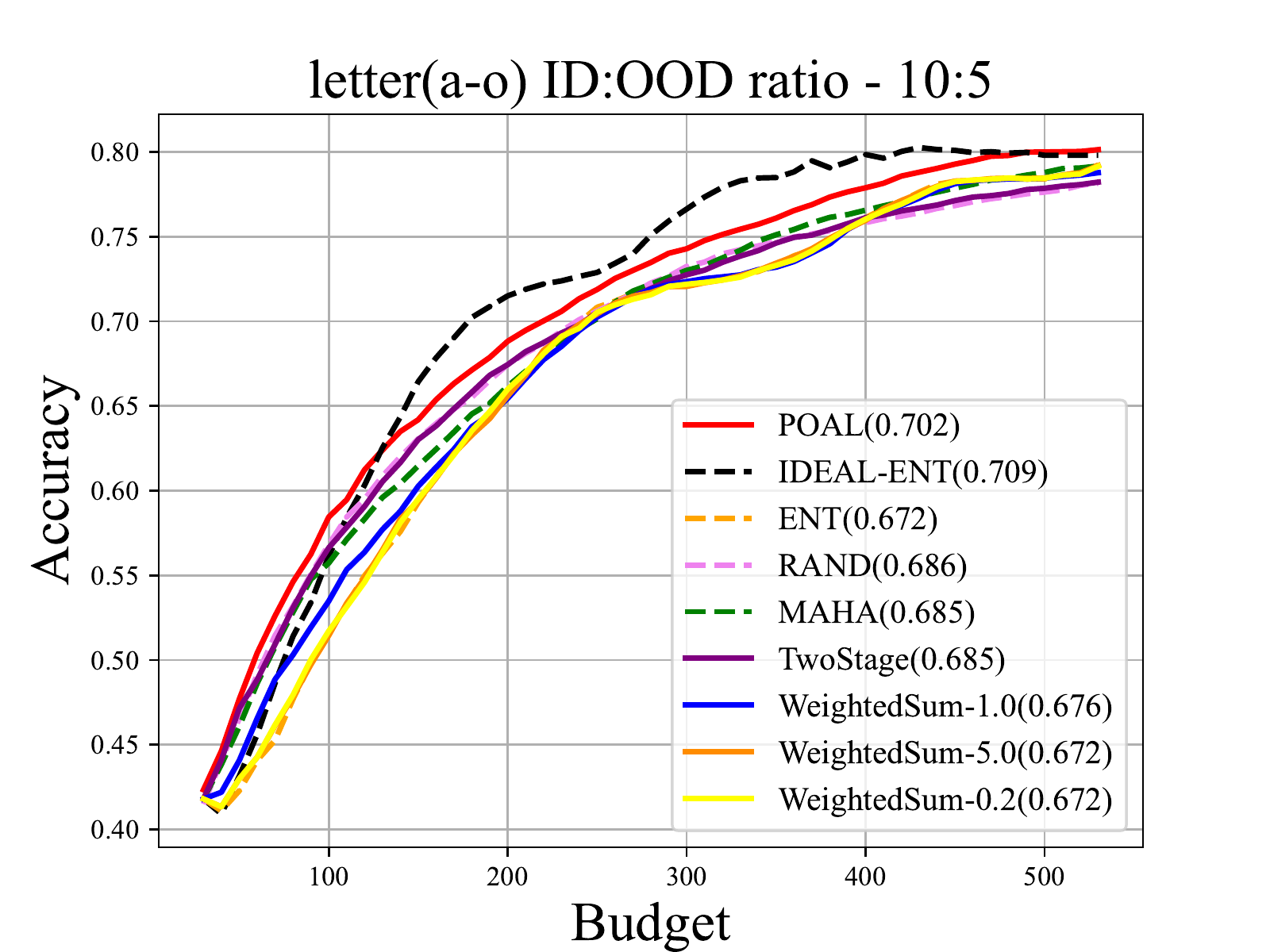}}
\subfloat[letter(a-p)]{\includegraphics[width=0.25\linewidth]{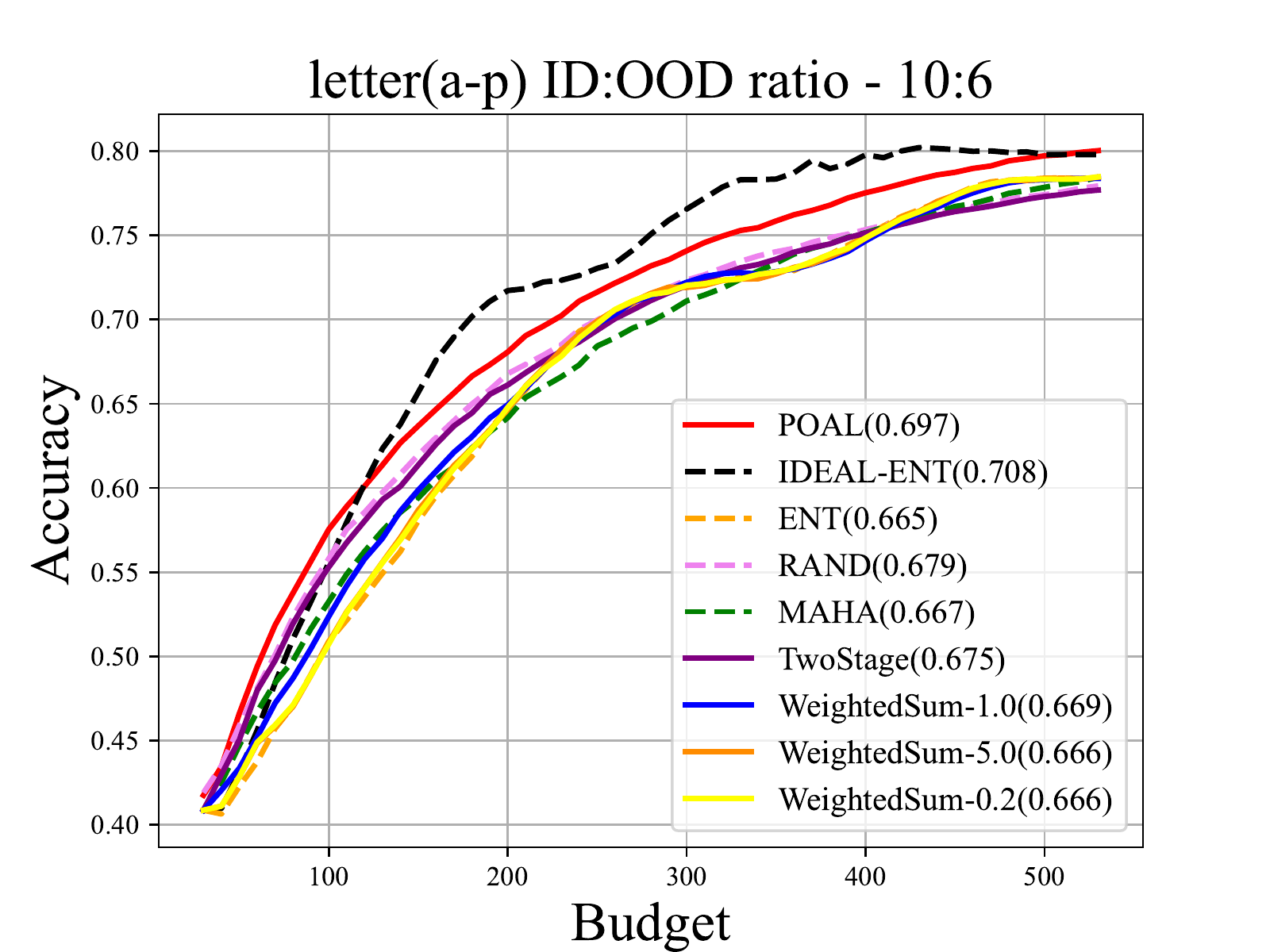}}

\subfloat[letter(a-q)]{\includegraphics[width=0.25\linewidth]{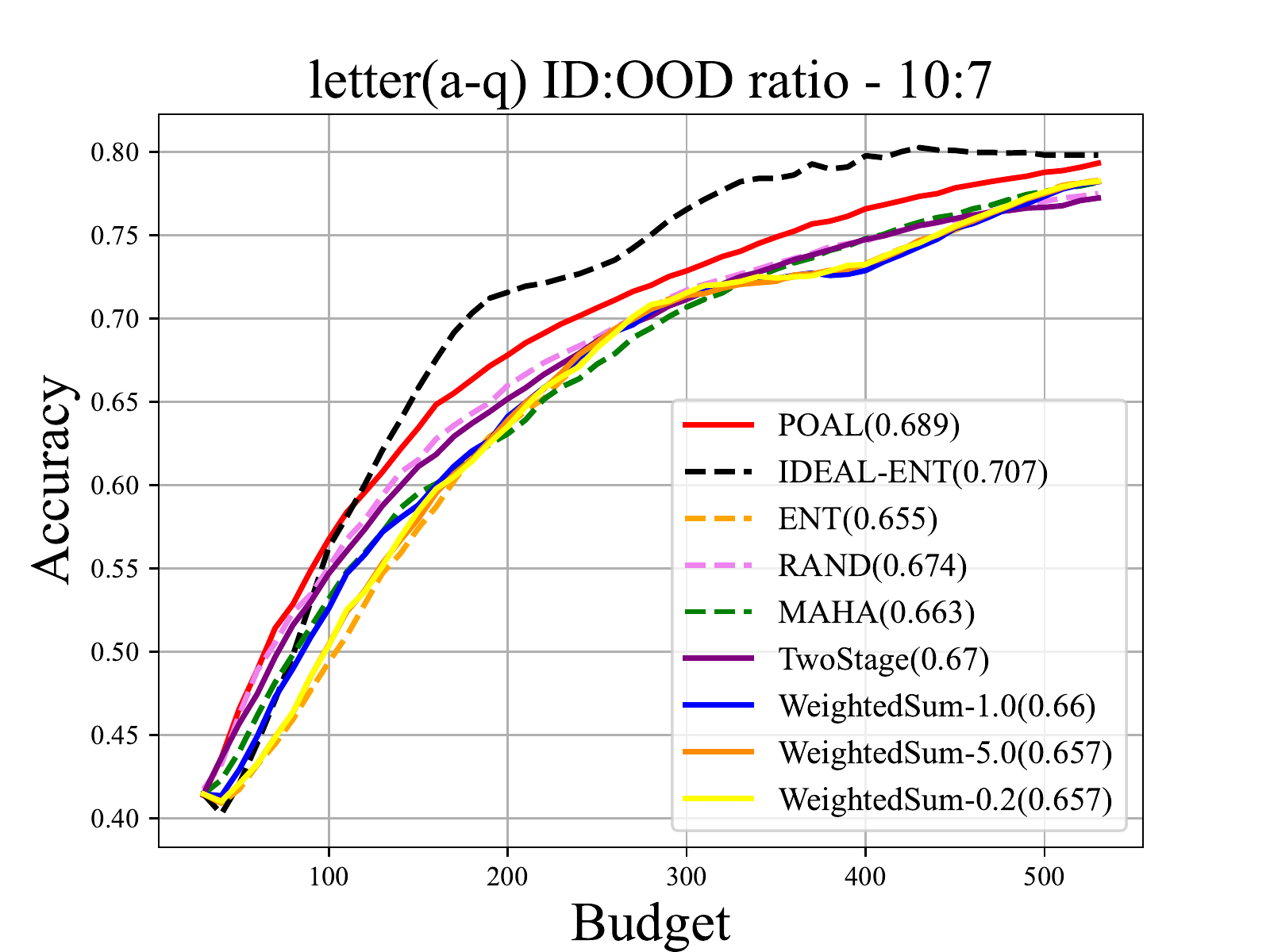}}
\subfloat[letter(a-r)]{\includegraphics[width=0.25\linewidth]{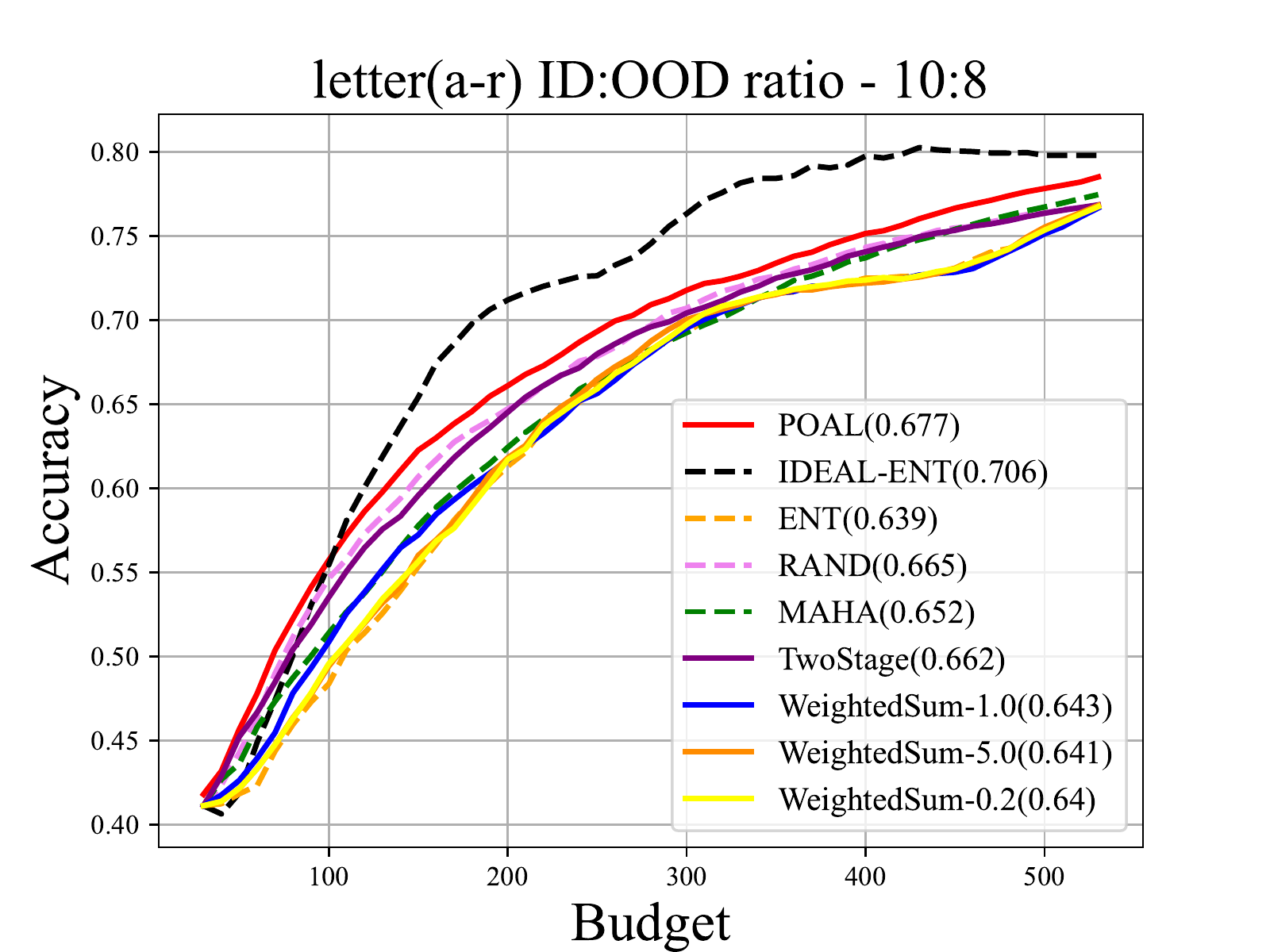}}
\subfloat[letter(a-s)]{\includegraphics[width=0.25\linewidth]{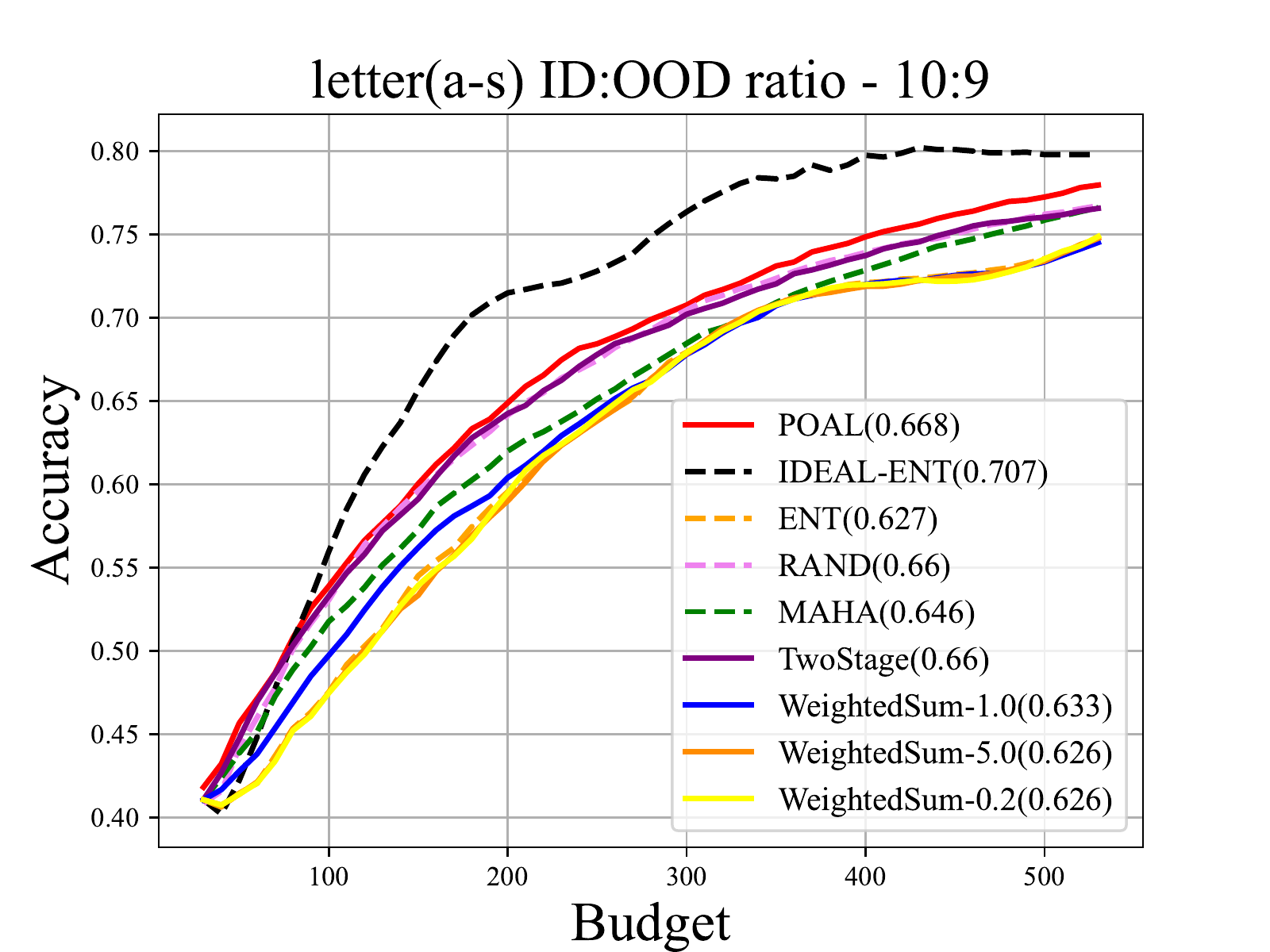}}
\subfloat[letter(a-t)]{\includegraphics[width=0.25\linewidth]{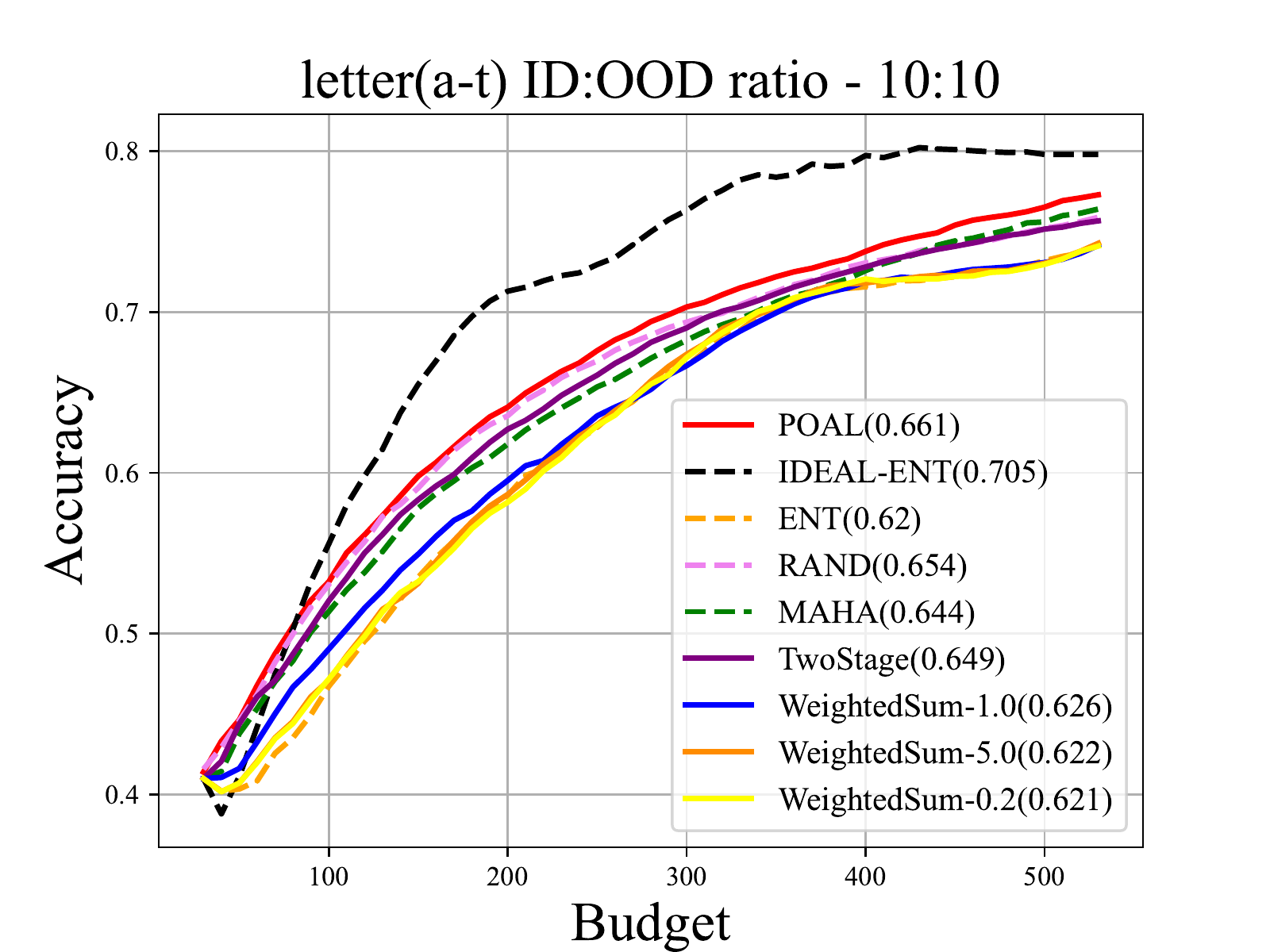}}

\subfloat[letter(a-u)]{\includegraphics[width=0.25\linewidth]{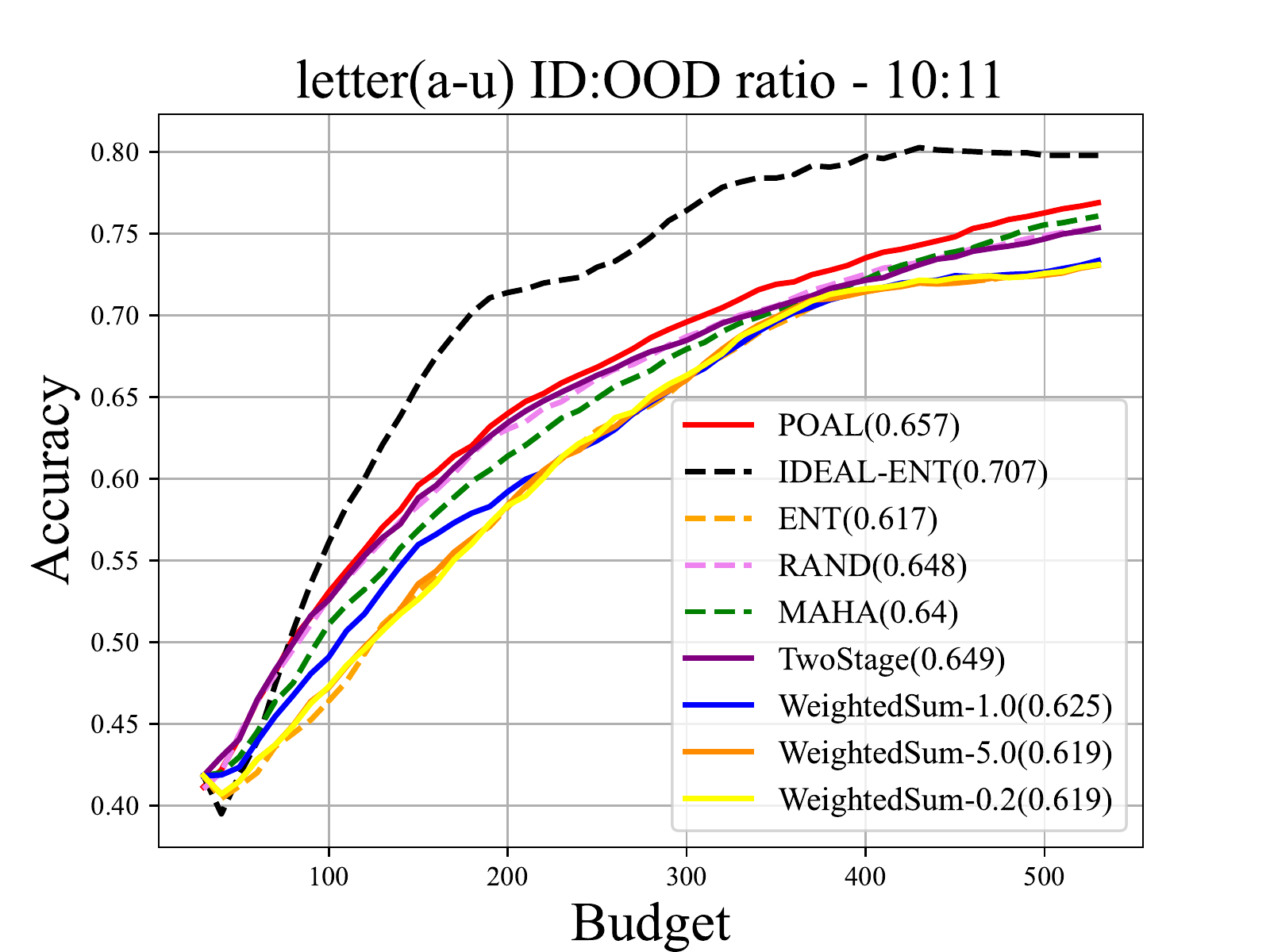}}
\subfloat[letter(a-v)]{\includegraphics[width=0.25\linewidth]{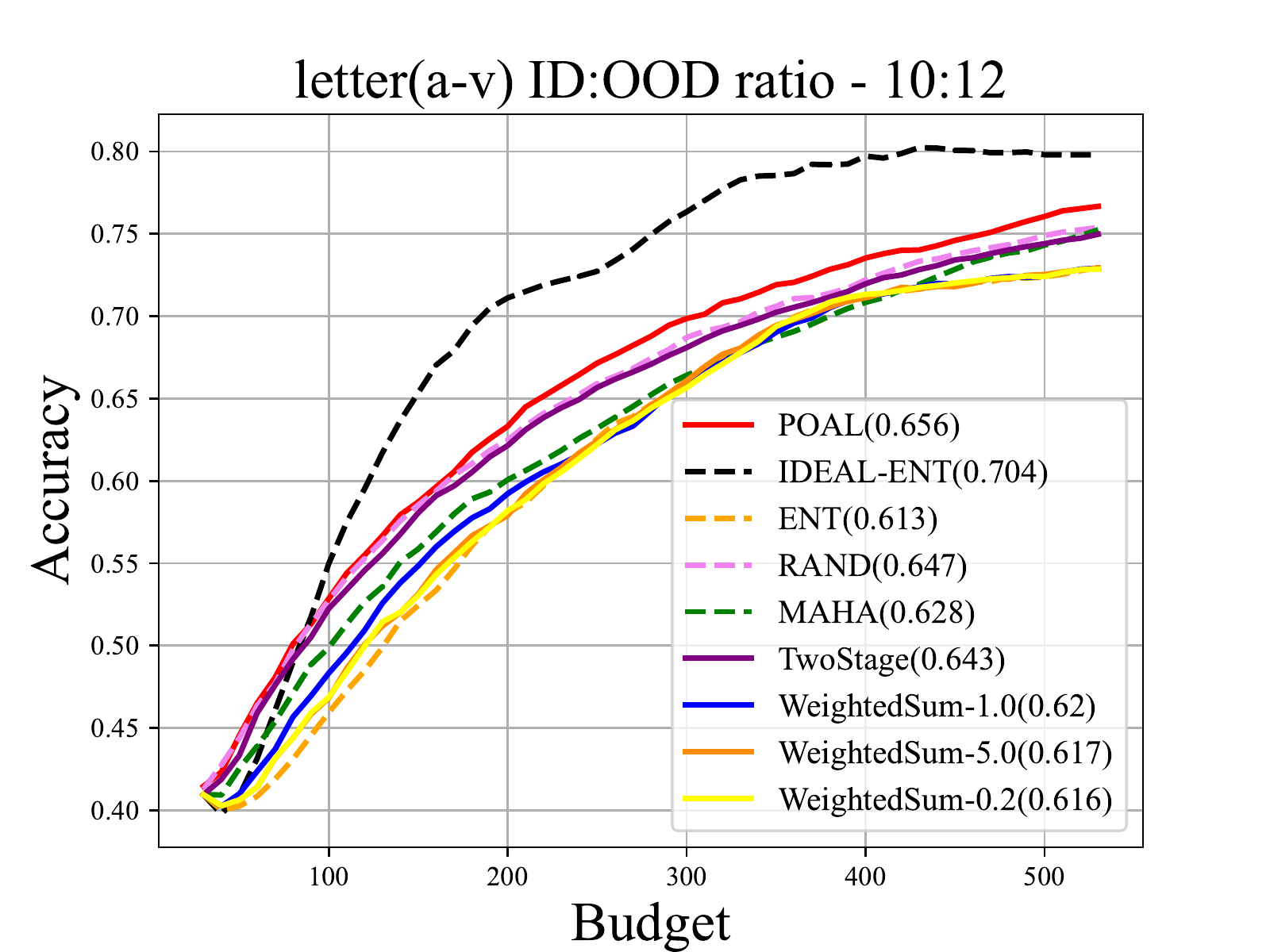}}
\subfloat[letter(a-w)]{\includegraphics[width=0.25\linewidth]{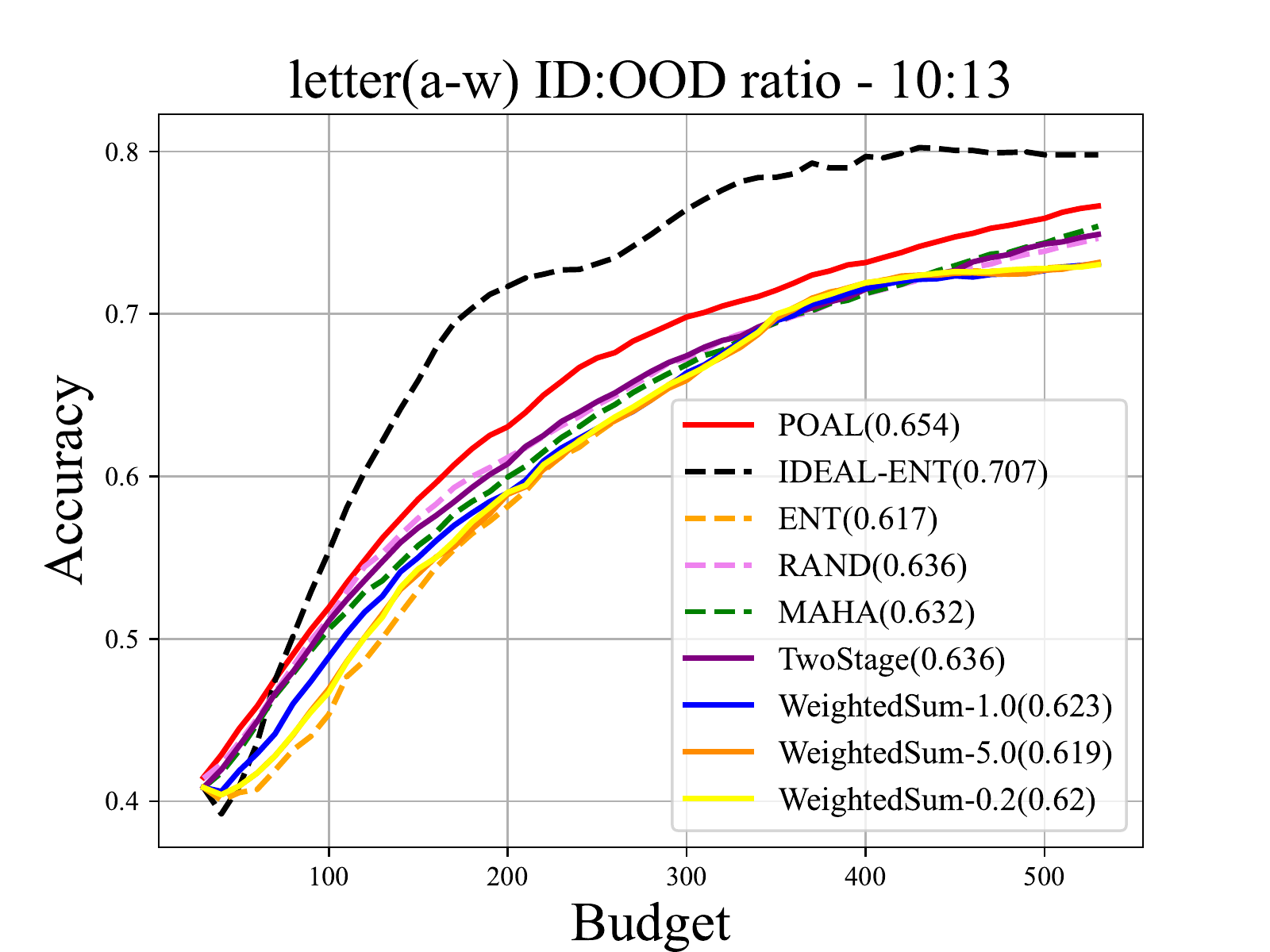}}
\subfloat[letter(a-x)]{\includegraphics[width=0.25\linewidth]{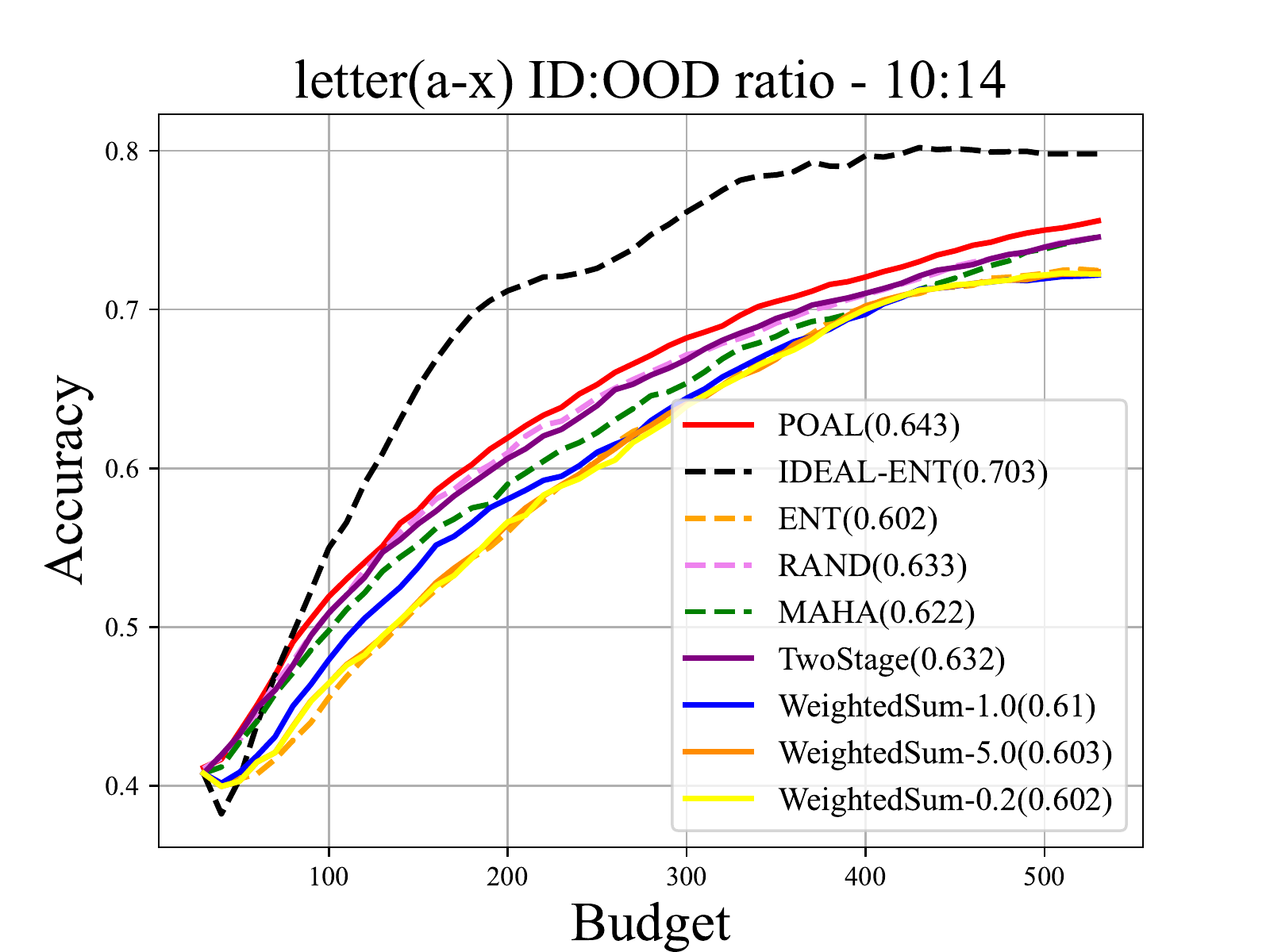}}

\subfloat[letter(a-y)]{\includegraphics[width=0.25\linewidth]{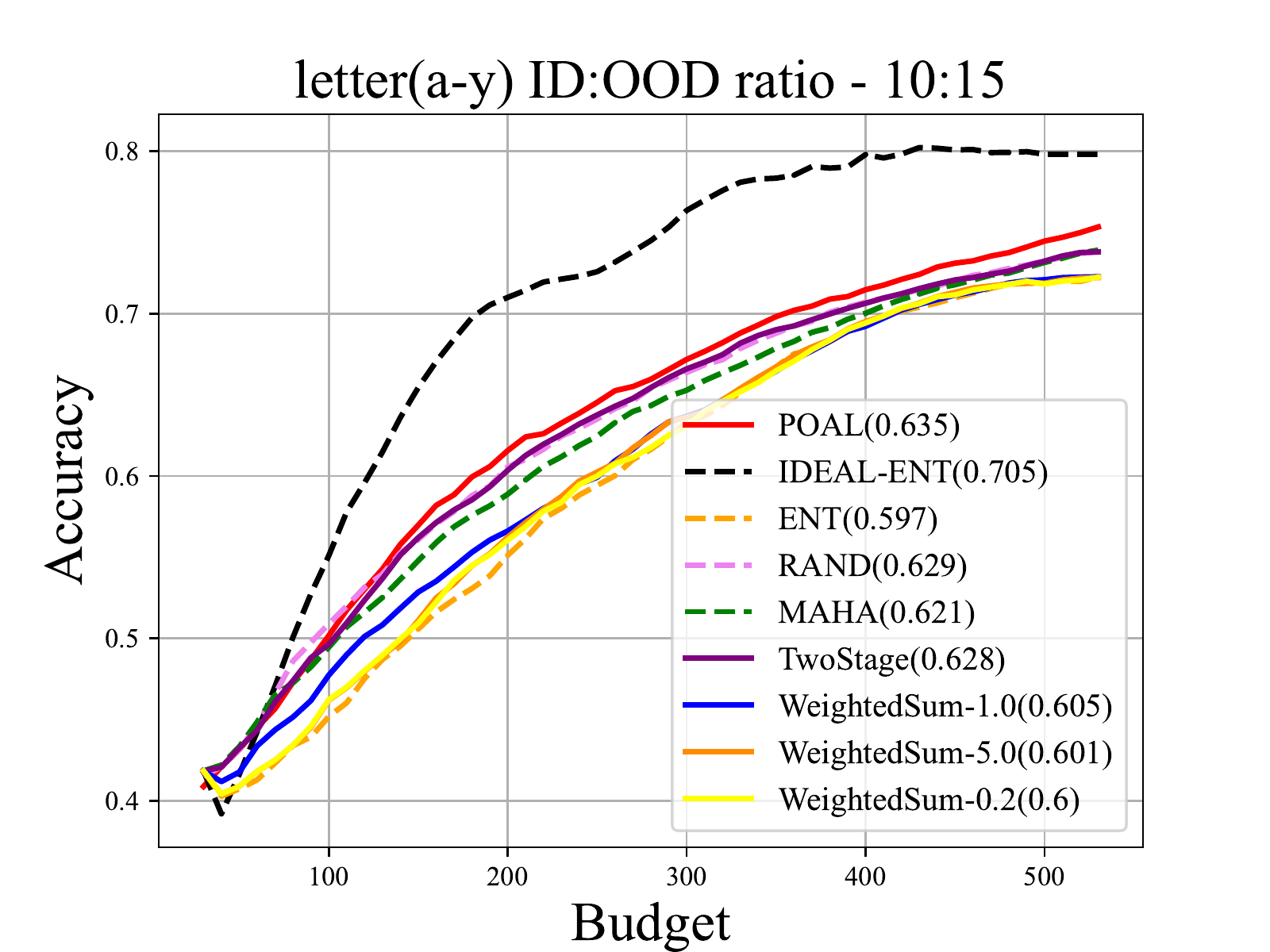}}
\subfloat[letter(a-z)]{\includegraphics[width=0.25\linewidth]{img/exp/classicalML/letter_az160_LR_fixedbudget.pdf}}
\caption{Accuracy vs. budget curves for classical ML tasks. The AUBC performances are shown in parentheses in the legend.}
\label{classical_ml}
\end{figure*}

\begin{figure*} [htb]
\centering
\subfloat[EX8]{\includegraphics[width=0.25\linewidth]{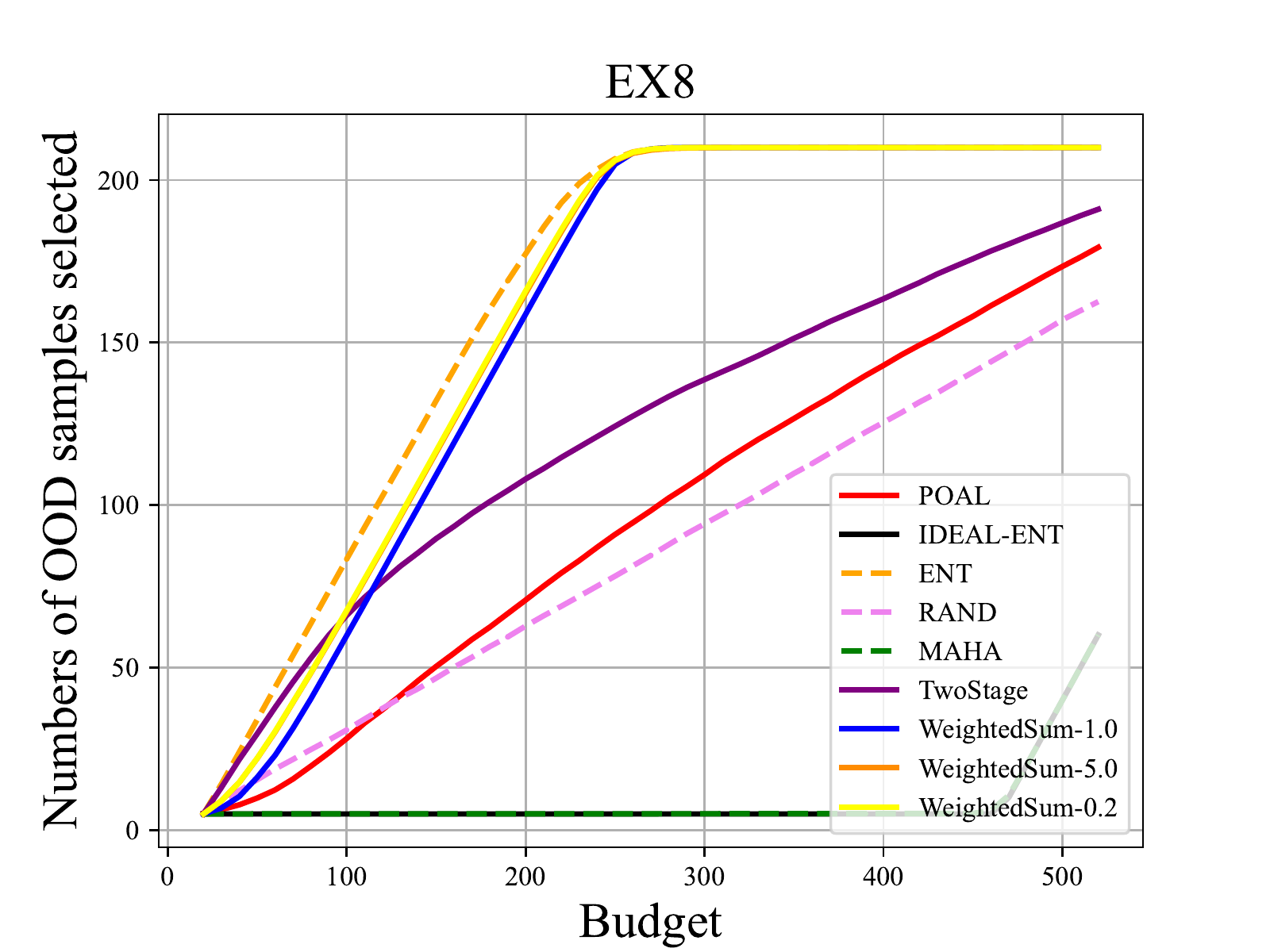}}
\subfloat[vowel]{\includegraphics[width=0.25\linewidth]{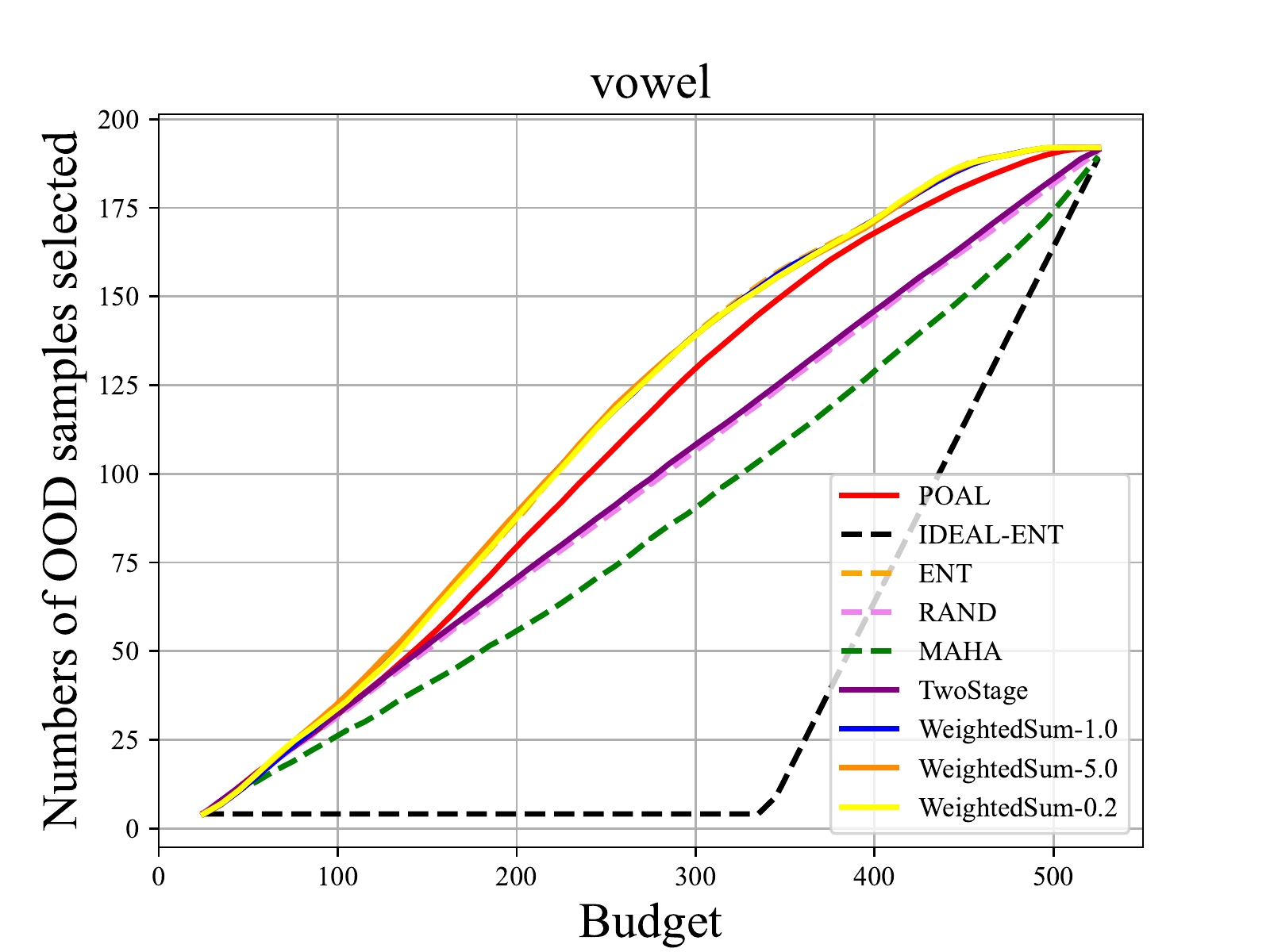}}
\subfloat[letter(a-k)]{\includegraphics[width=0.25\linewidth]{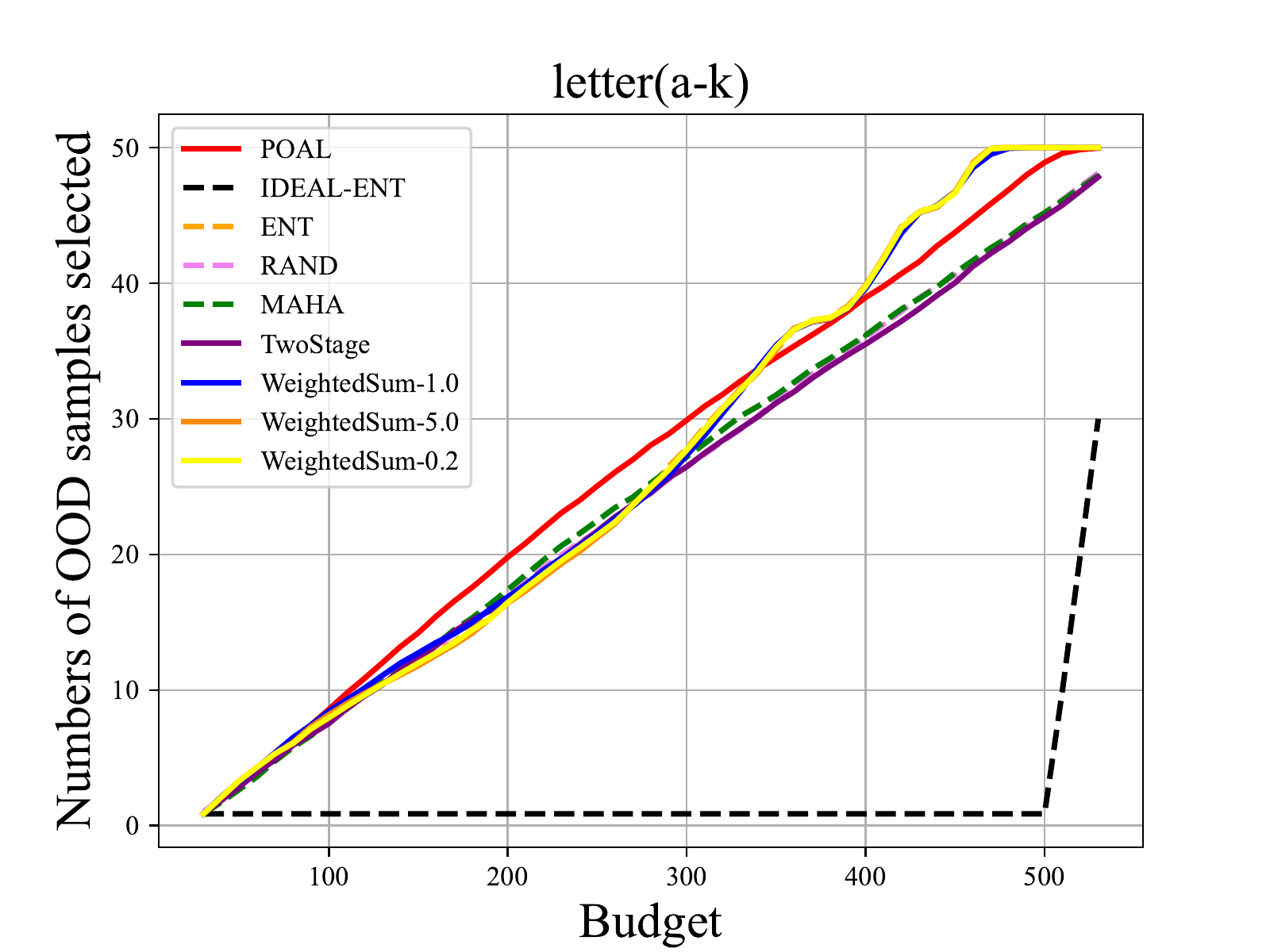}}
\subfloat[letter(a-l)]{\includegraphics[width=0.25\linewidth]{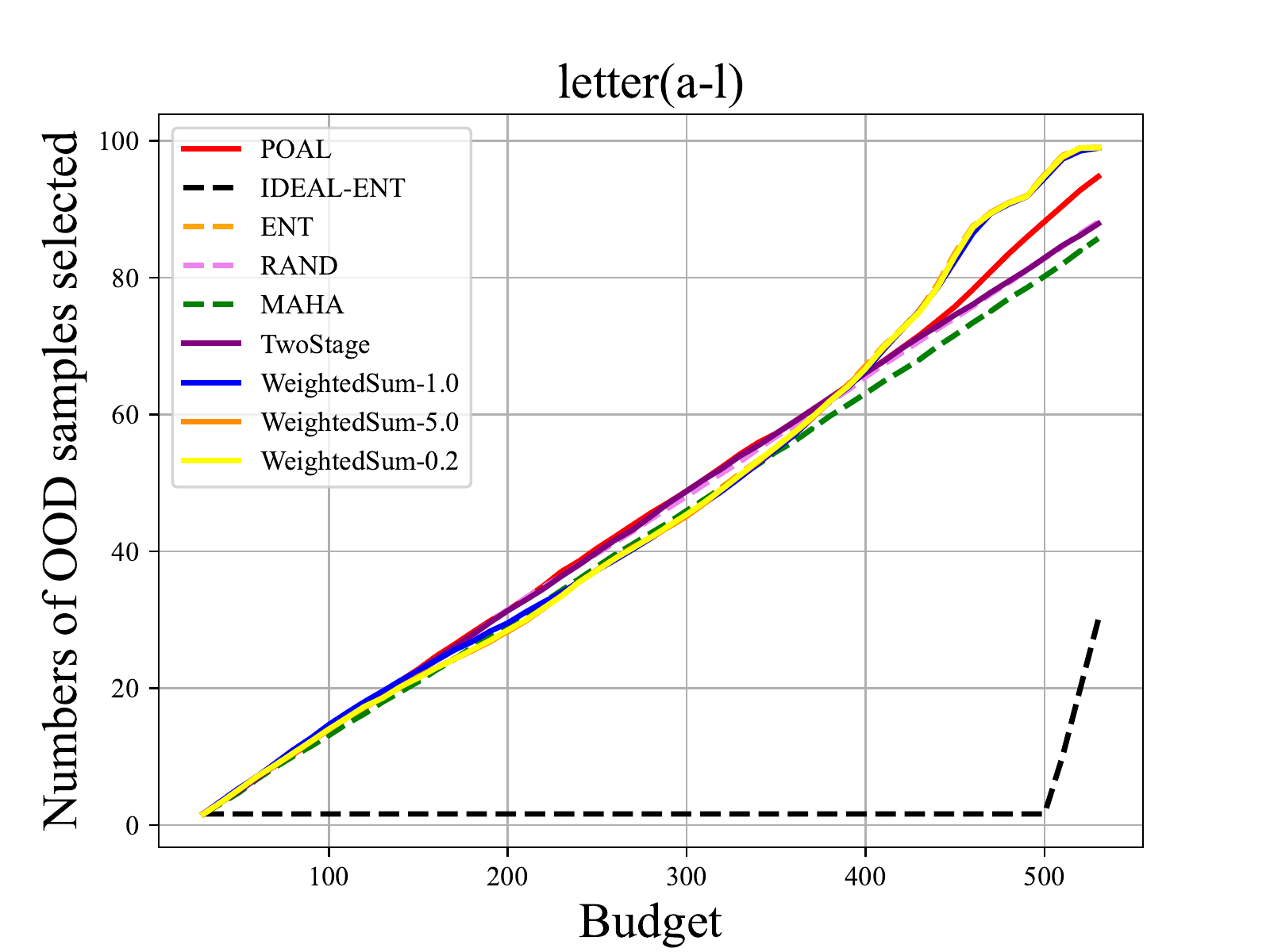}}

\subfloat[letter(a-n)]{\includegraphics[width=0.25\linewidth]{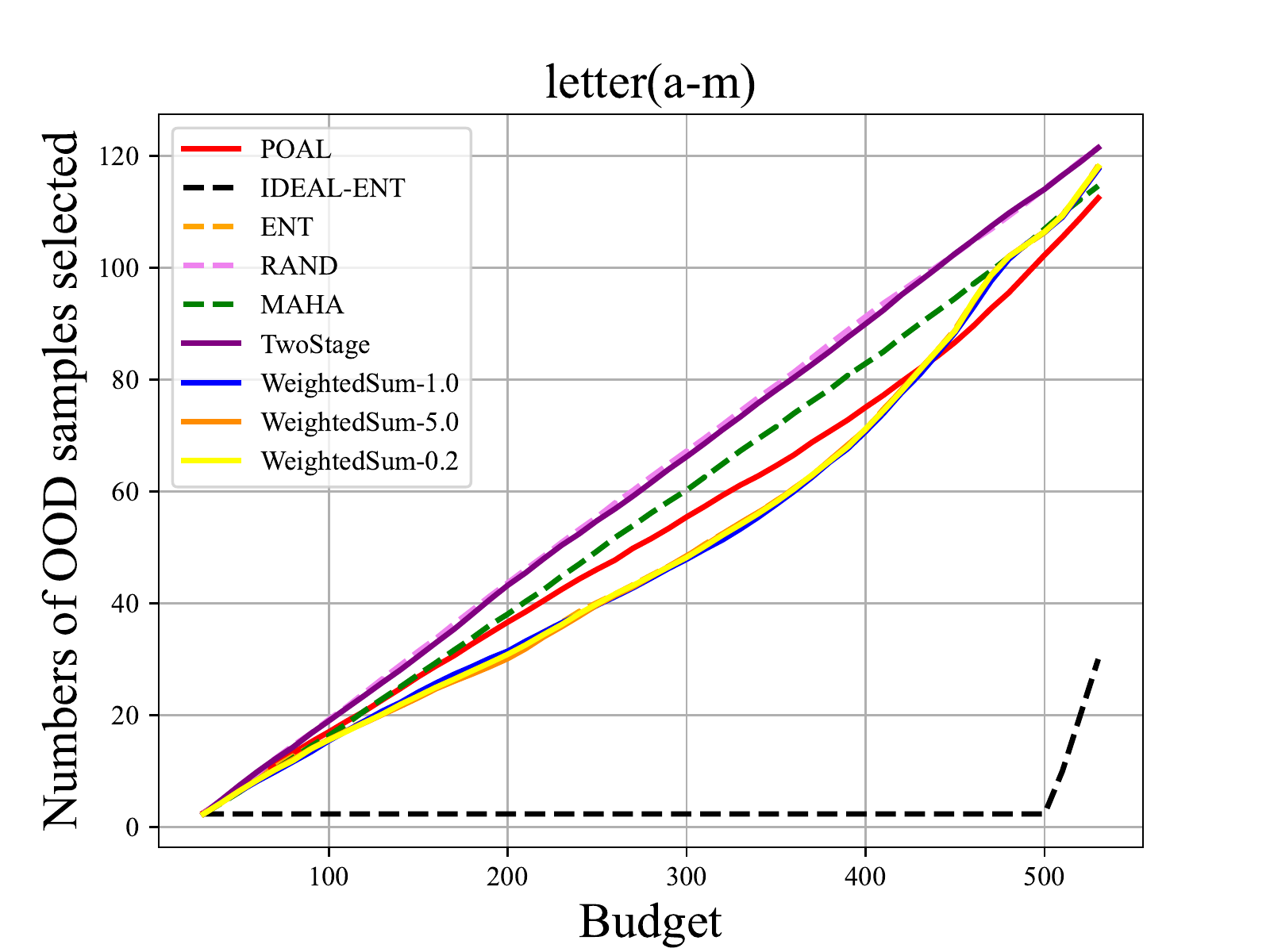}}
\subfloat[letter(a-m)]{\includegraphics[width=0.25\linewidth]{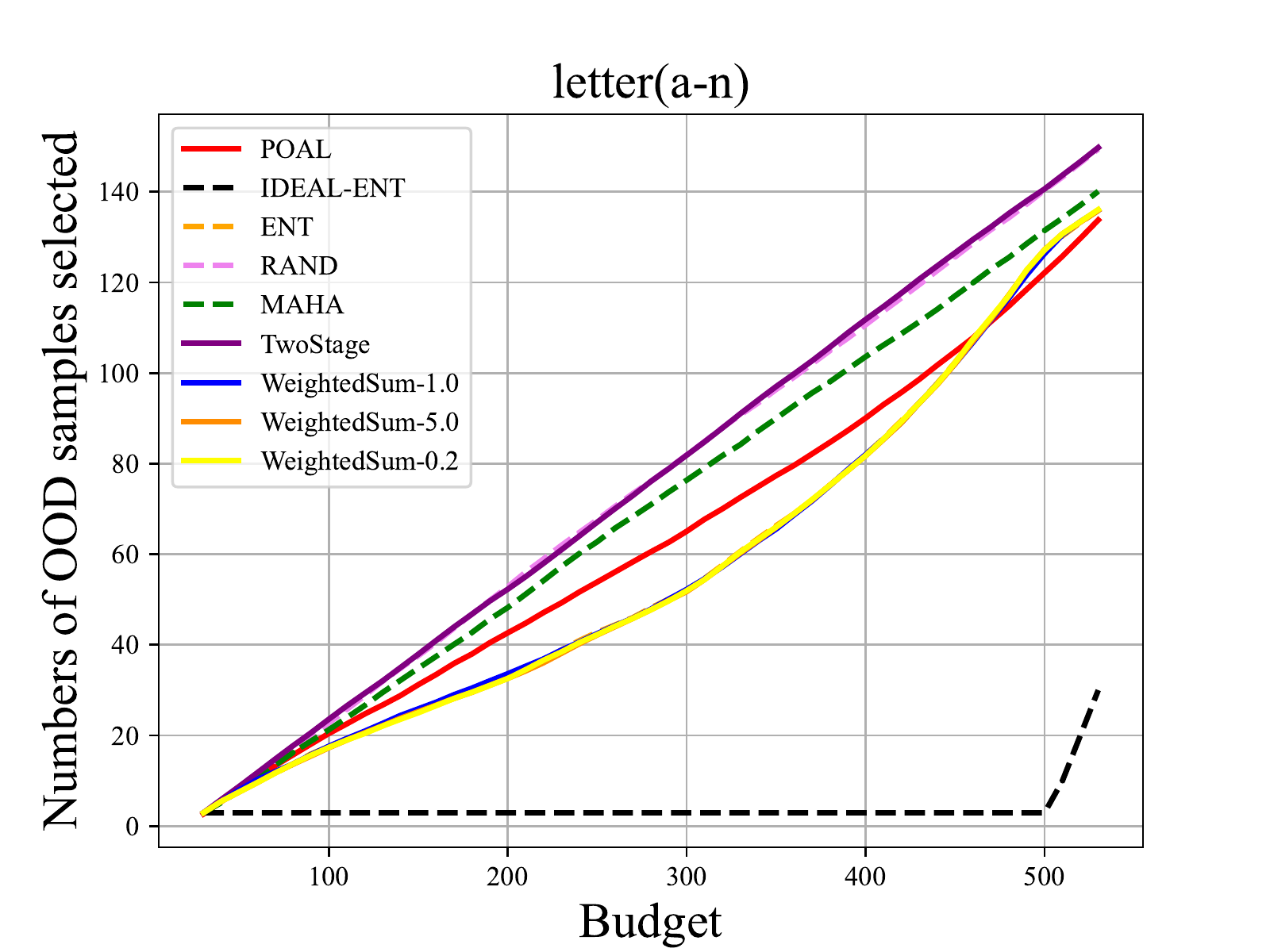}}
\subfloat[letter(a-o)]{\includegraphics[width=0.25\linewidth]{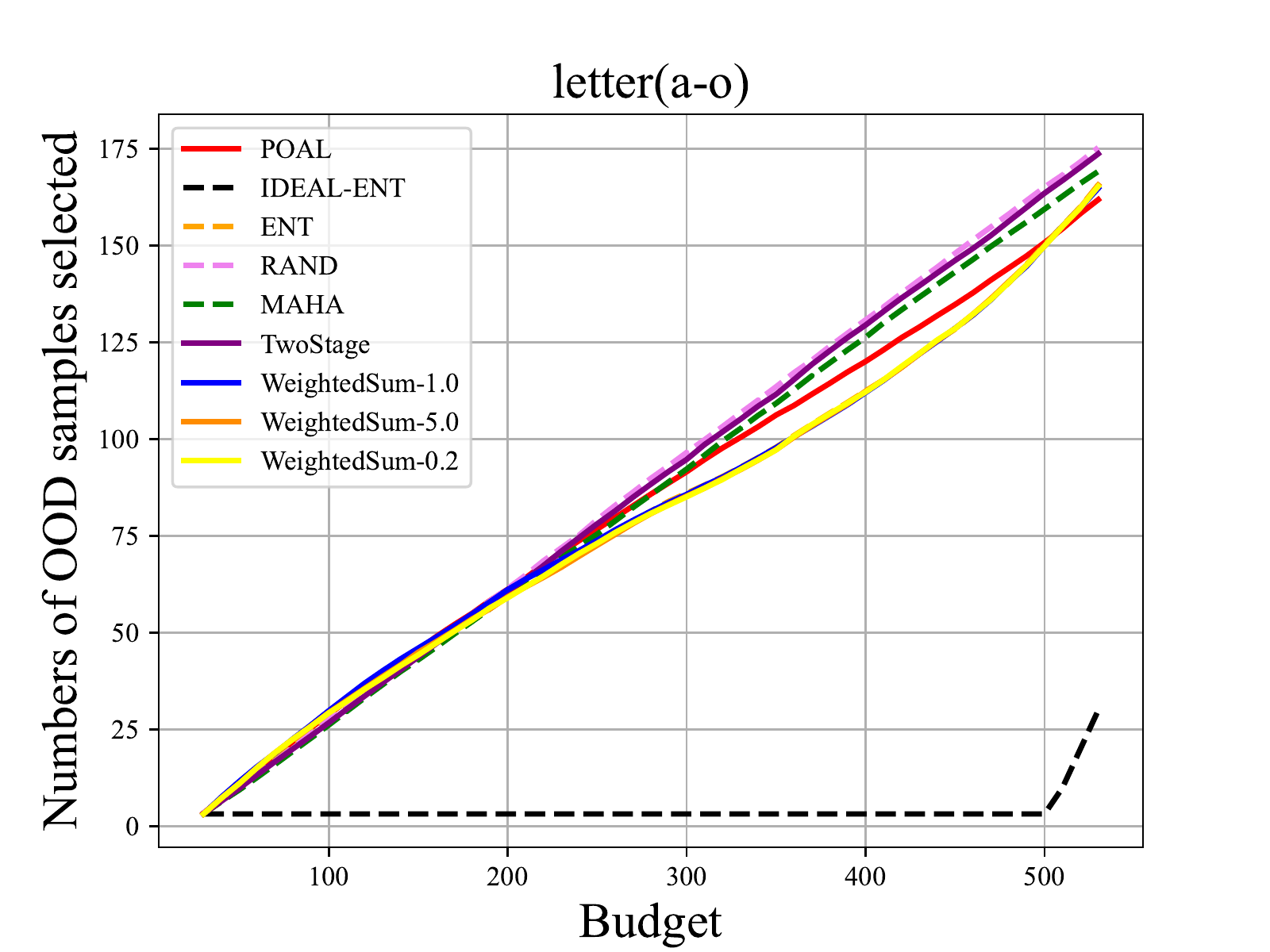}}
\subfloat[letter(a-p)]{\includegraphics[width=0.25\linewidth]{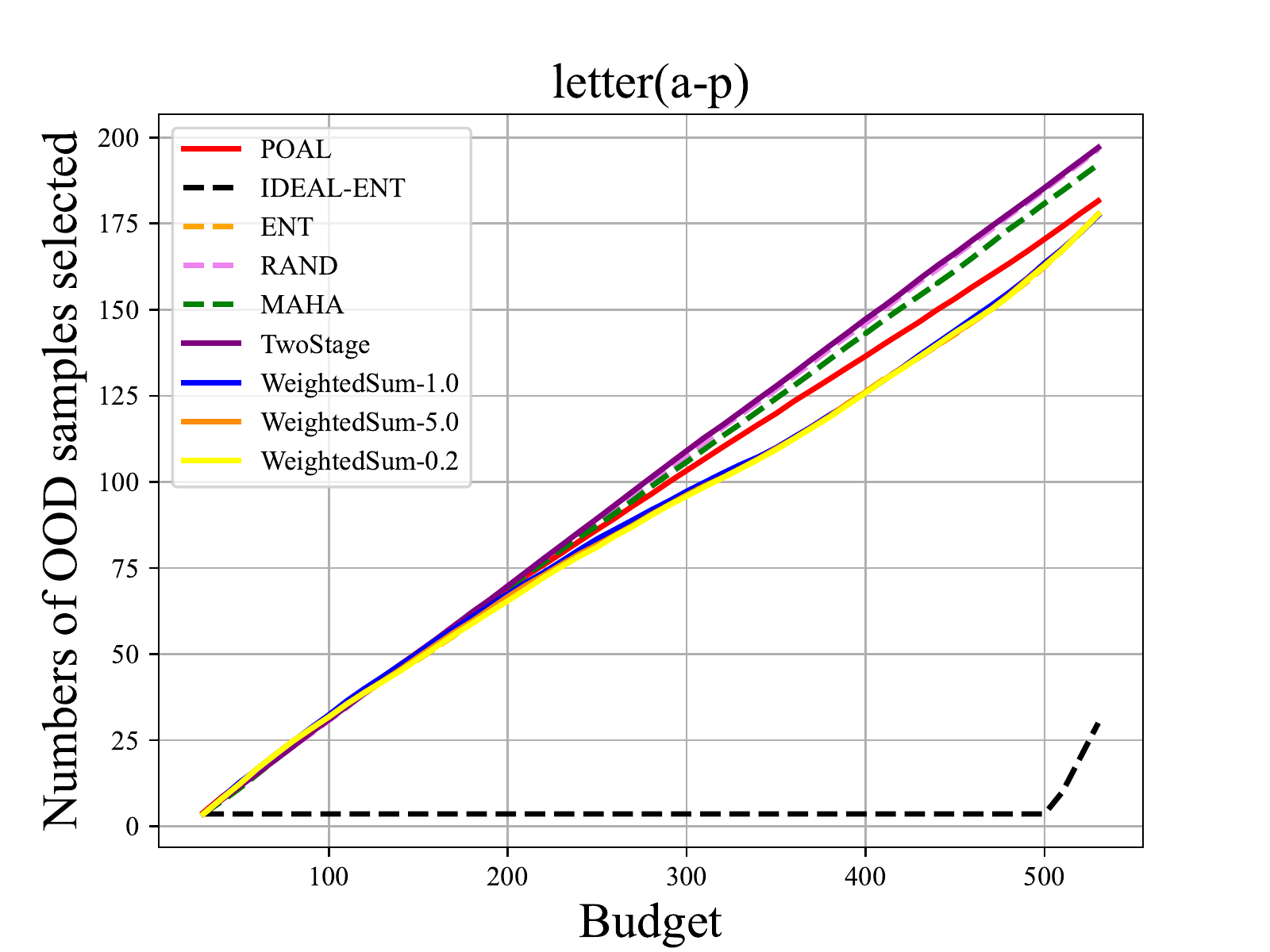}}

\subfloat[letter(a-q)]{\includegraphics[width=0.25\linewidth]{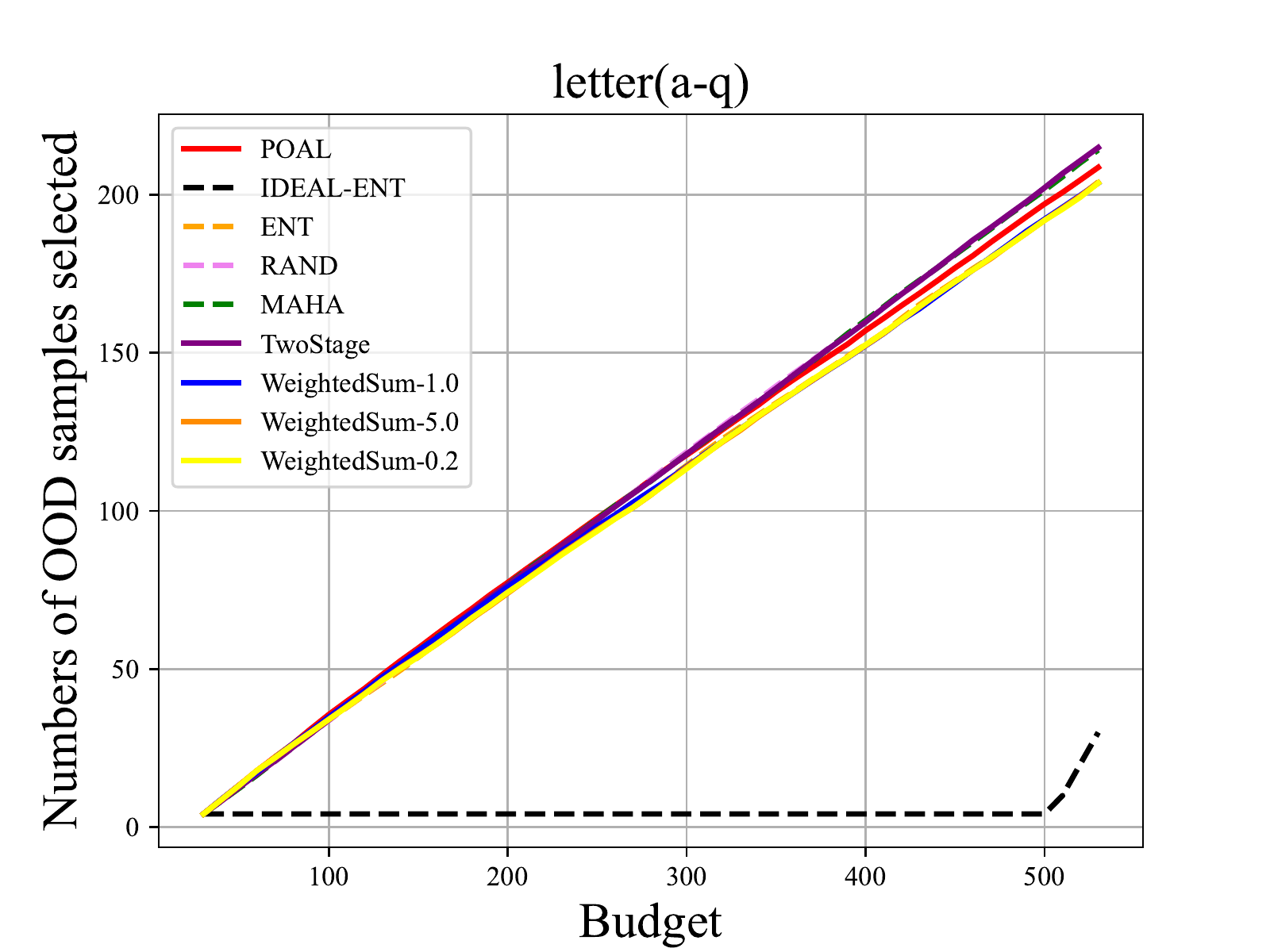}}
\subfloat[letter(a-r)]{\includegraphics[width=0.25\linewidth]{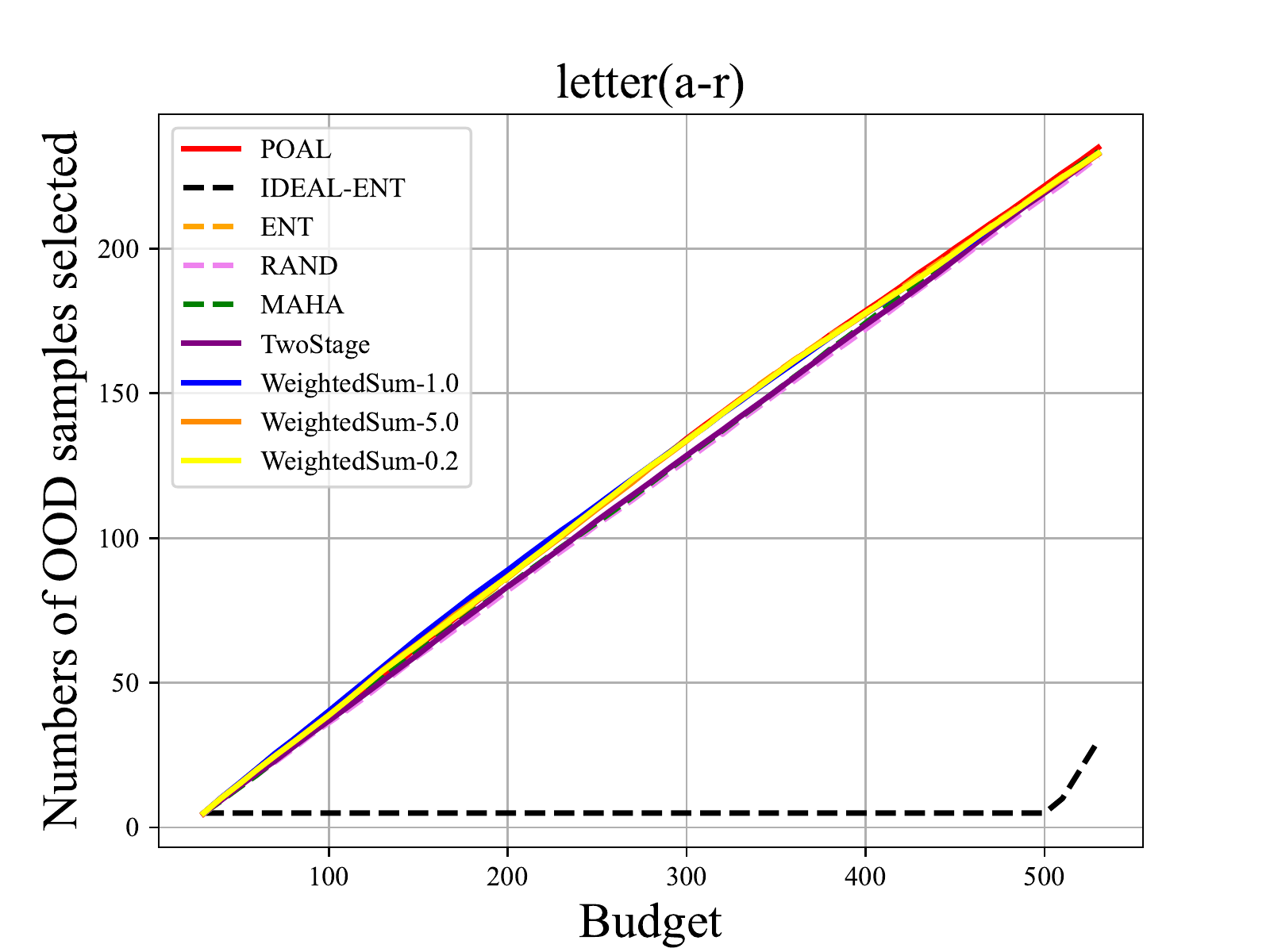}}
\subfloat[letter(a-s)]{\includegraphics[width=0.25\linewidth]{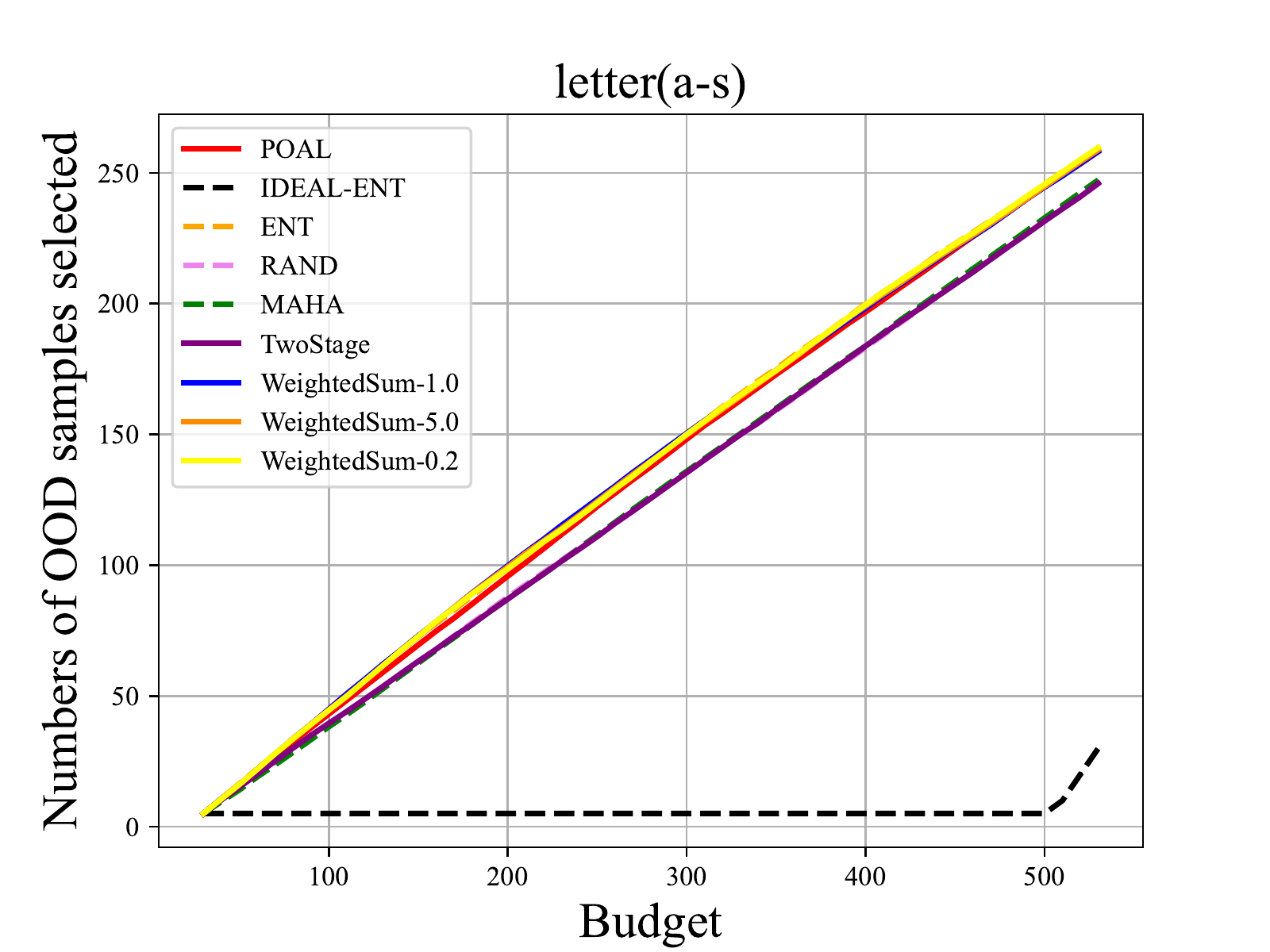}}
\subfloat[letter(a-t)]{\includegraphics[width=0.25\linewidth]{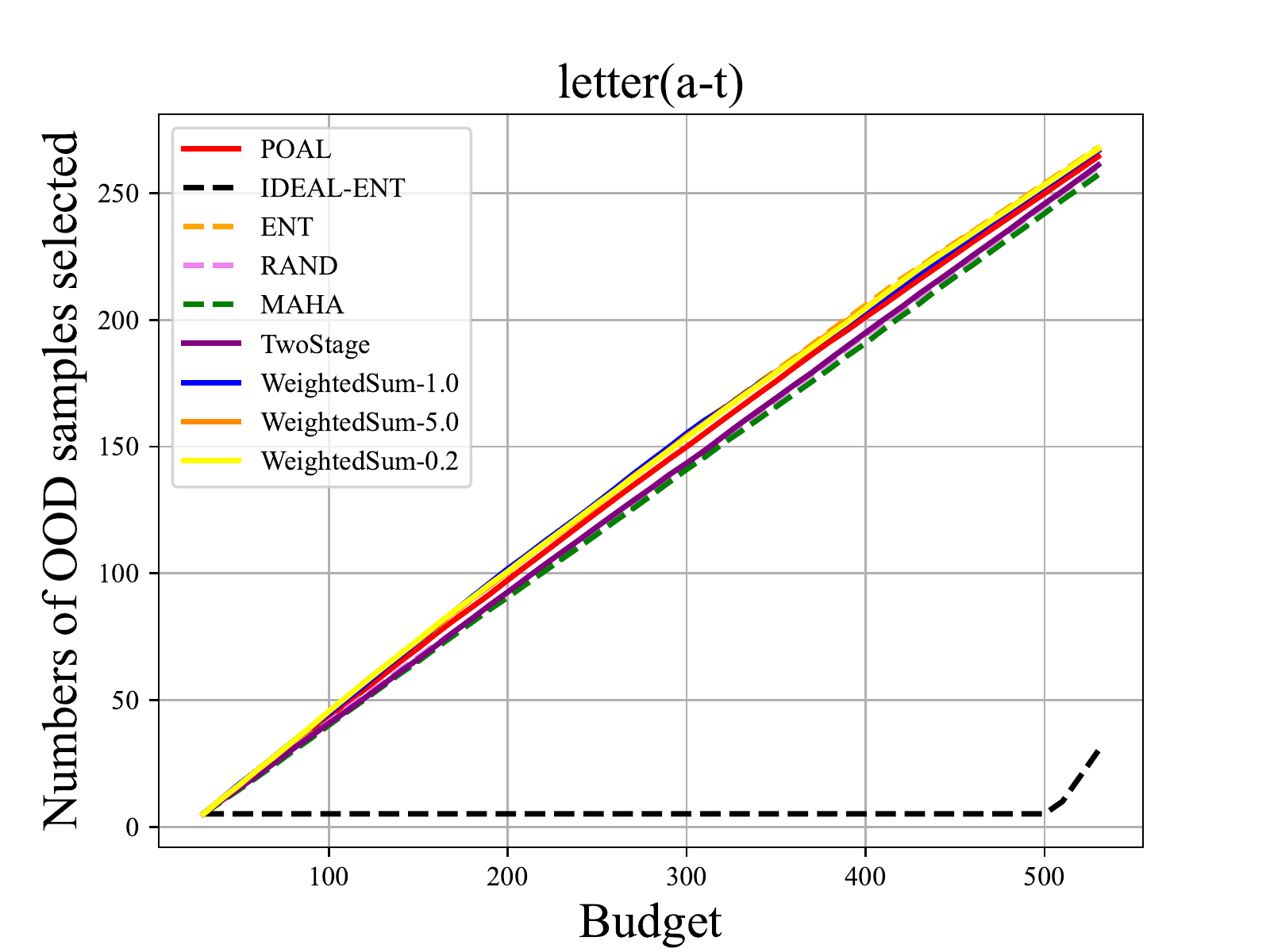}}

\subfloat[letter(a-u)]{\includegraphics[width=0.25\linewidth]{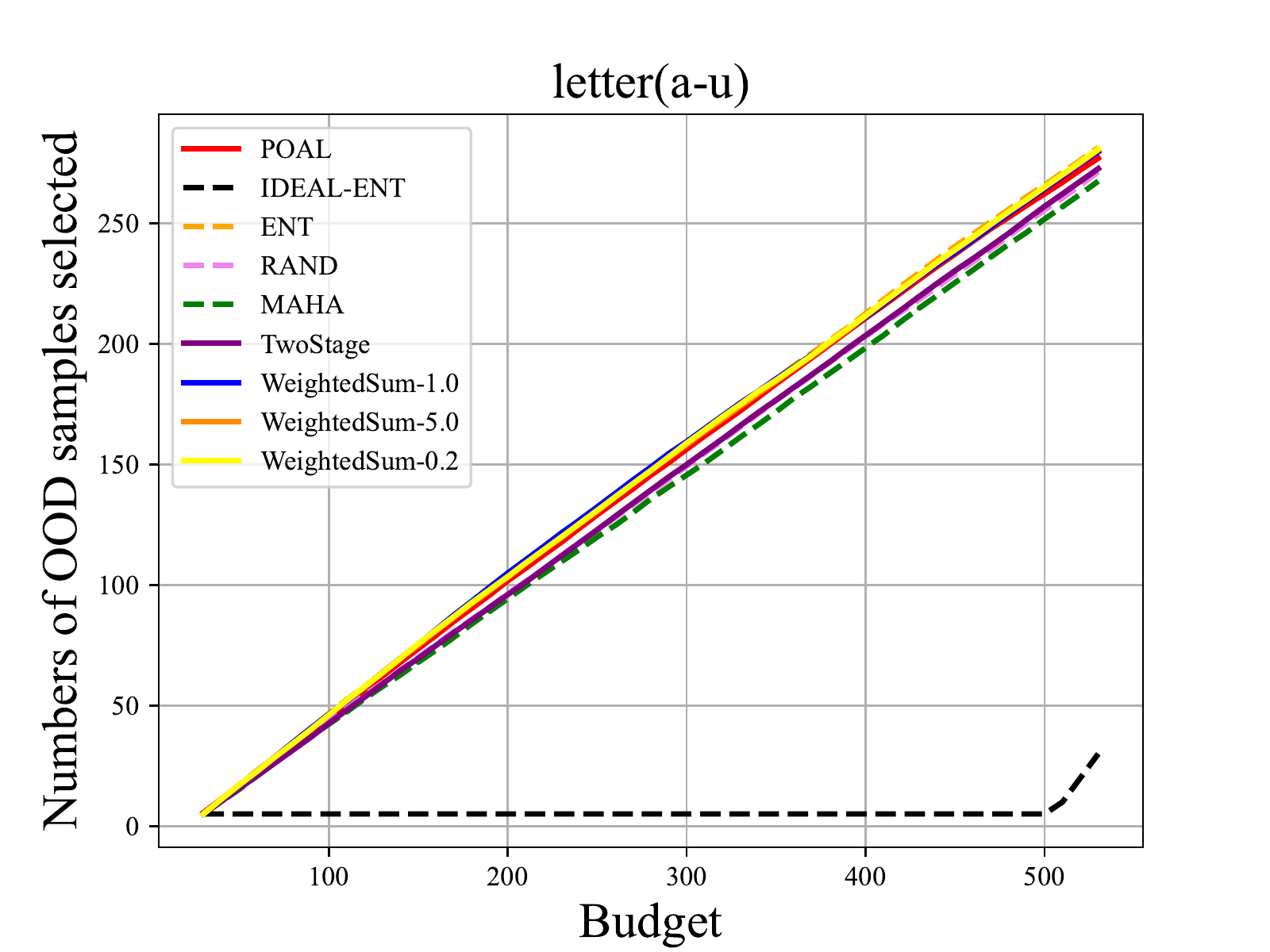}}
\subfloat[letter(a-v)]{\includegraphics[width=0.25\linewidth]{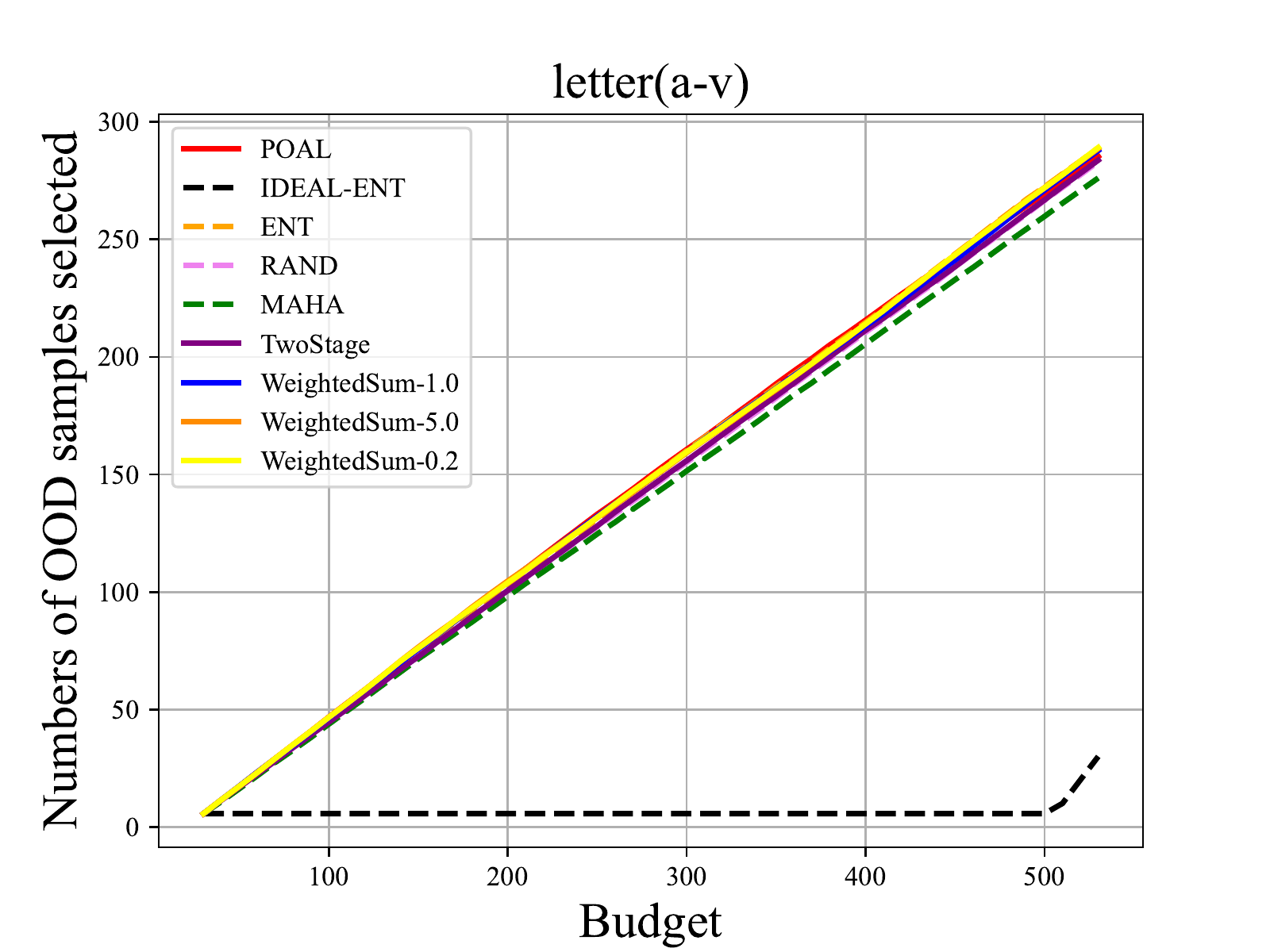}}
\subfloat[letter(a-w)]{\includegraphics[width=0.25\linewidth]{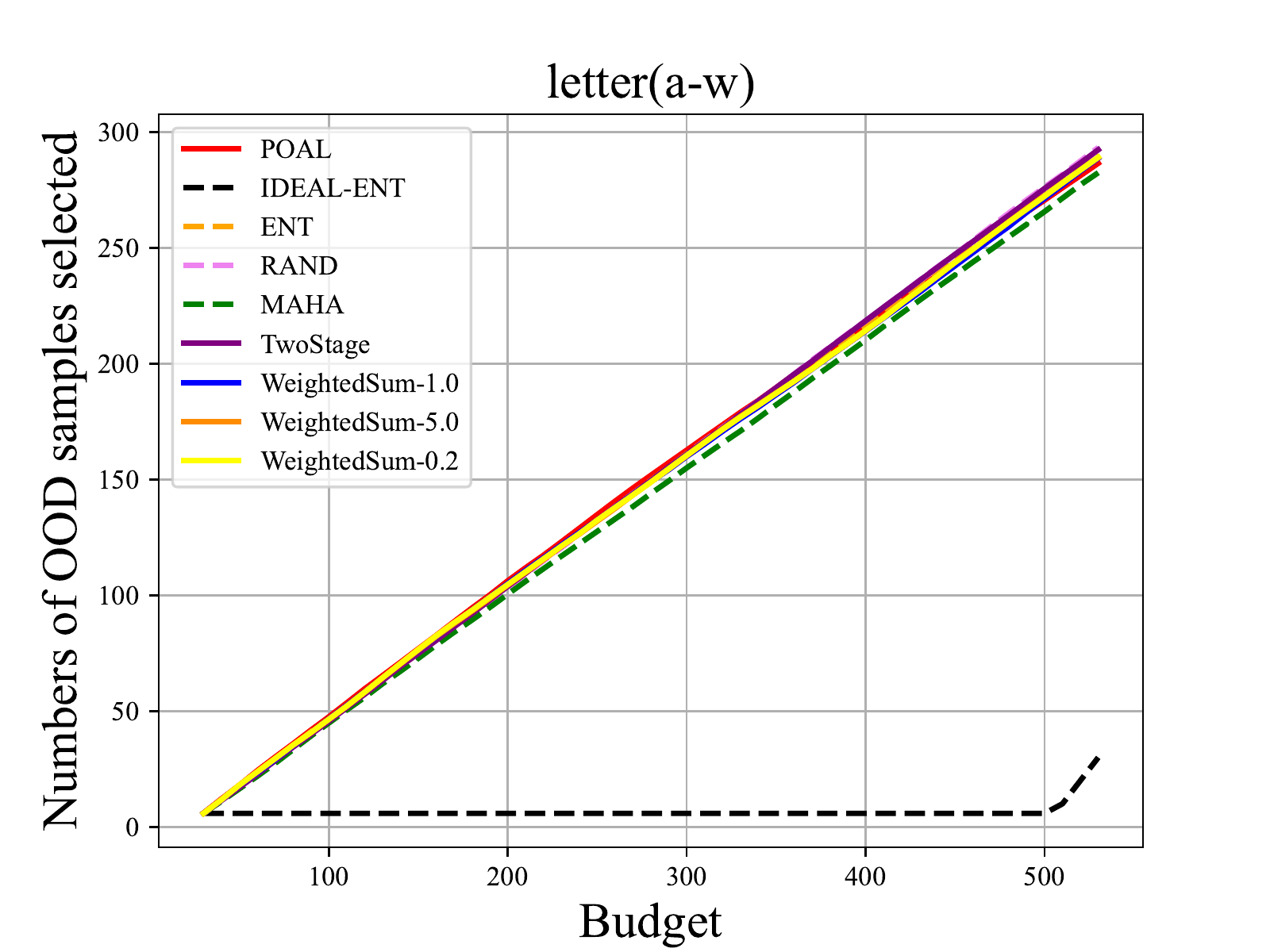}}
\subfloat[letter(a-x)]{\includegraphics[width=0.25\linewidth]{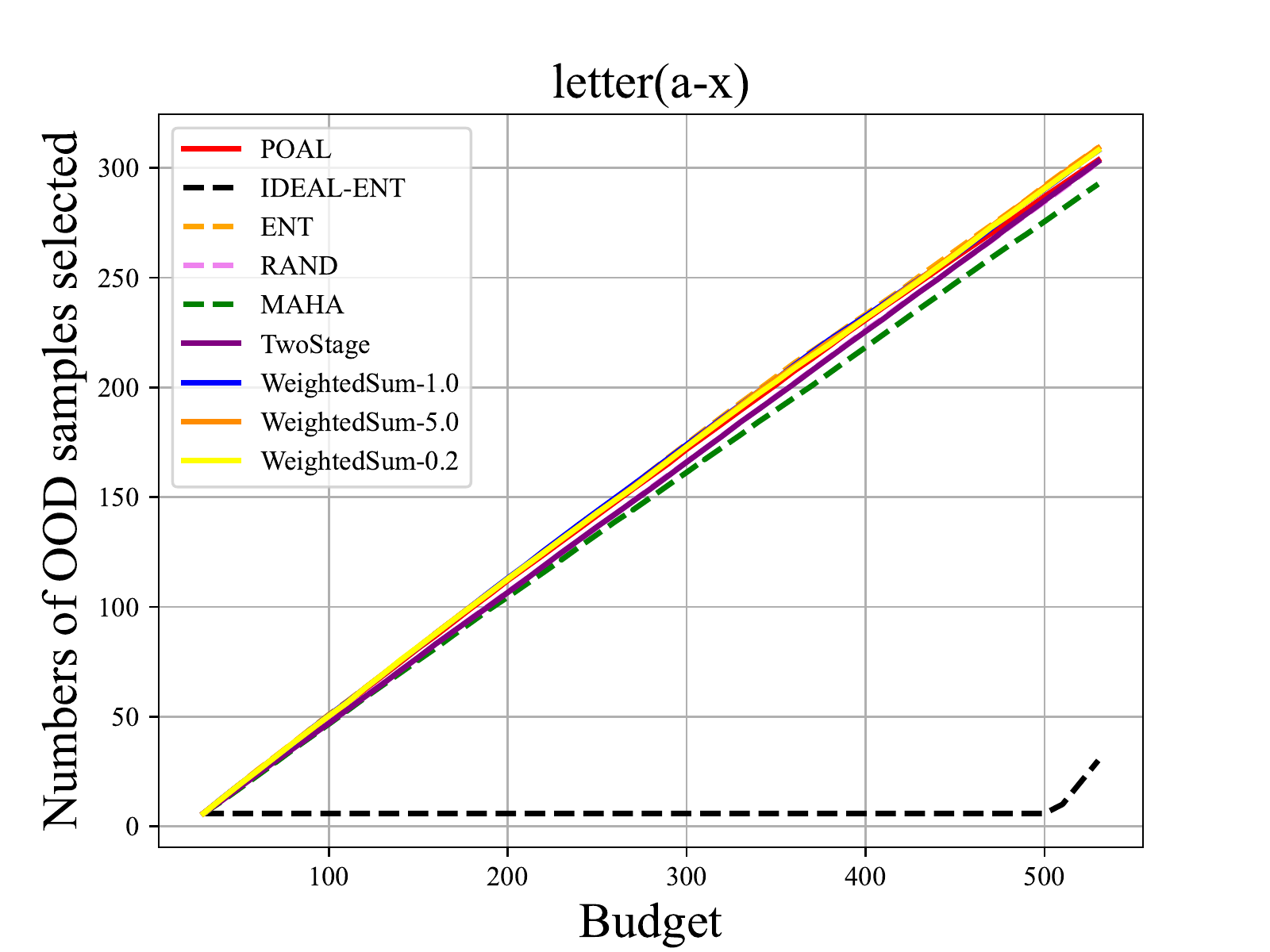}}

\subfloat[letter(a-y)]{\includegraphics[width=0.25\linewidth]{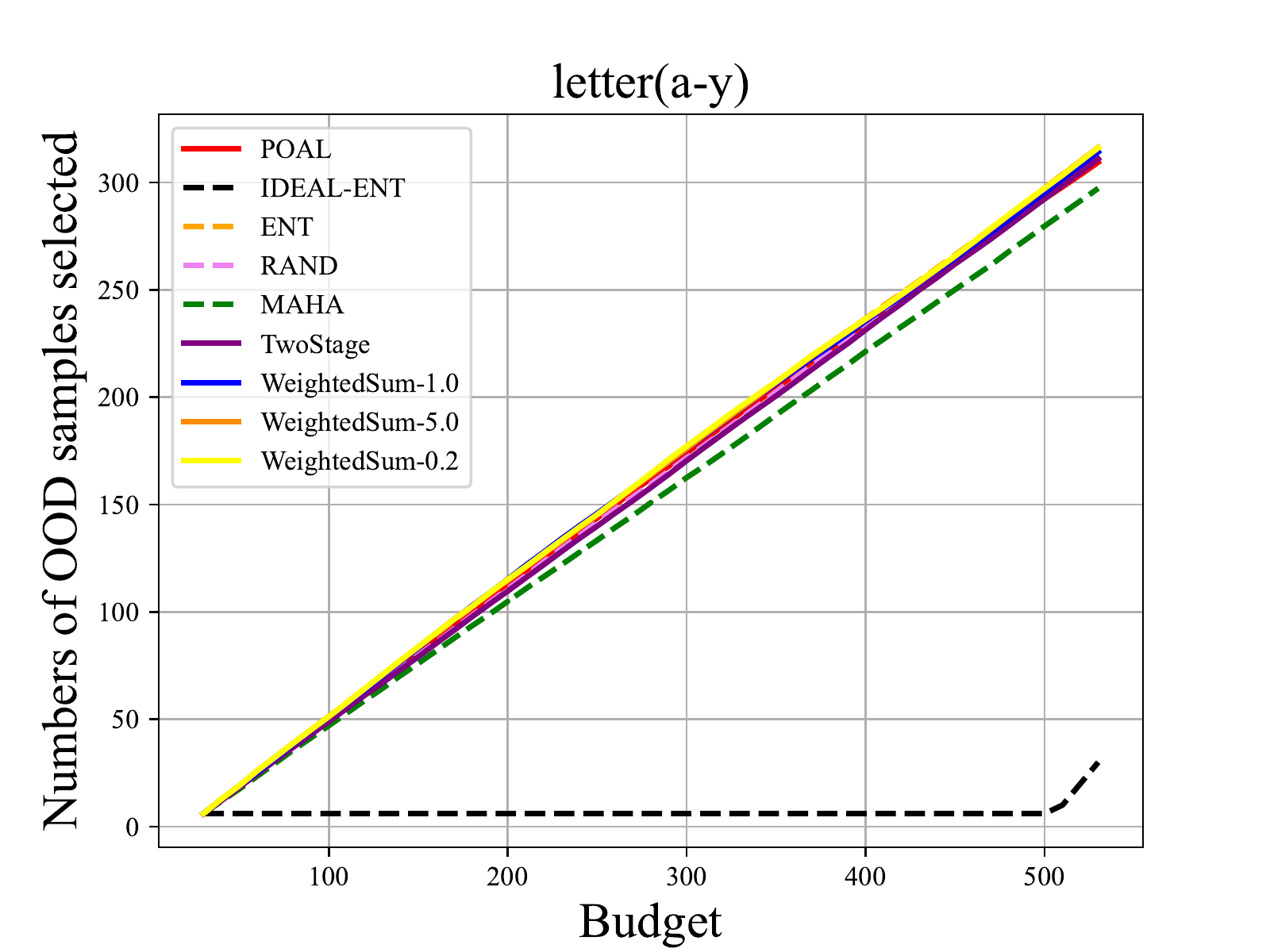}}
\subfloat[letter(a-z)]{\includegraphics[width=0.25\linewidth]{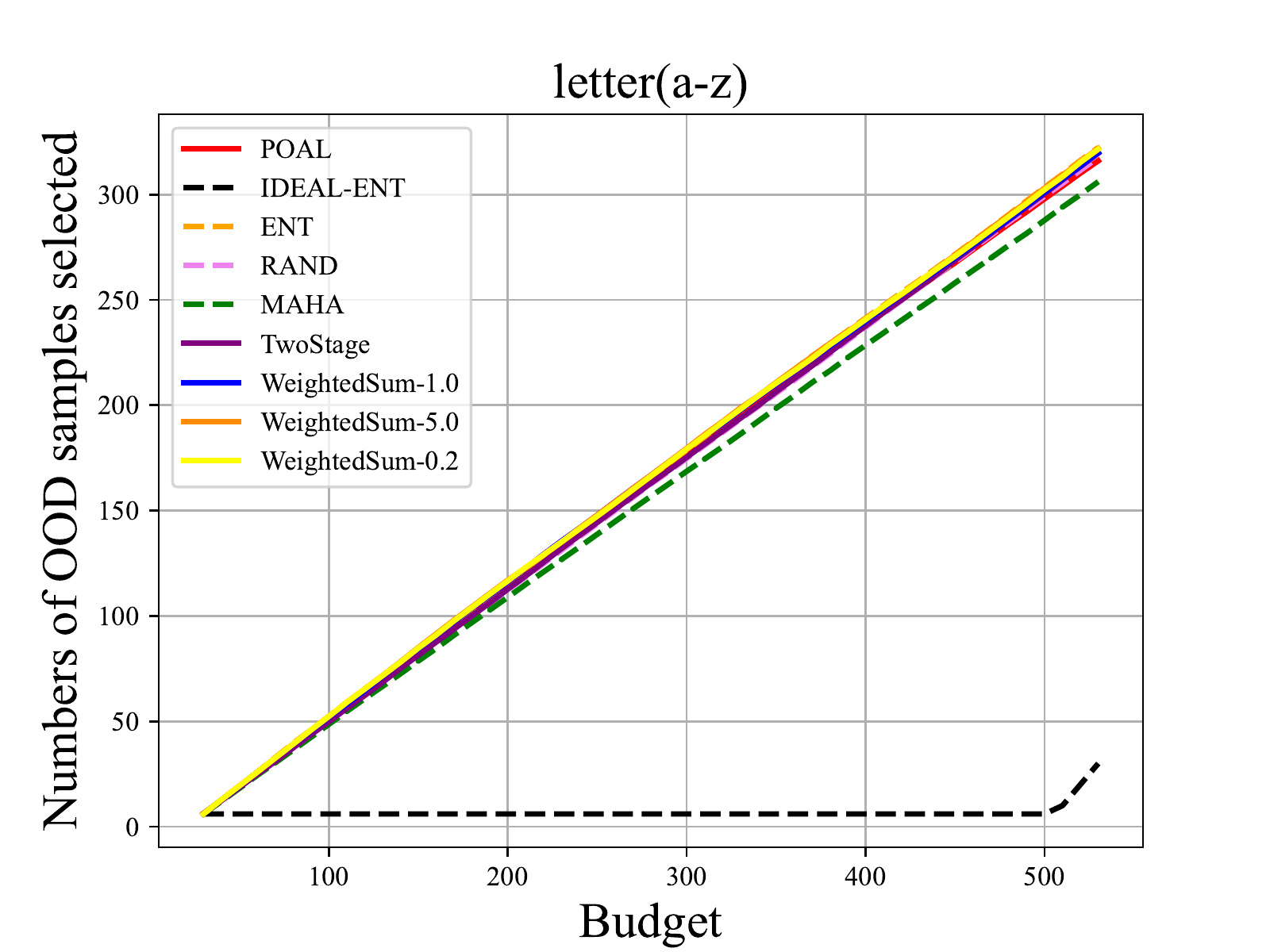}}

\caption{Numbers of OOD samples selected vs. budget curves for classical ML tasks during AL processes.}
\label{classical_ml_ood}
\end{figure*}

\subsubsection{Additional experimental results on DL tasks.}
\label{comp-ccal-similar}

\begin{figure*} [htb]
\centering
\subfloat[Accuracy vs. Budget]{\includegraphics[width=0.5\linewidth]{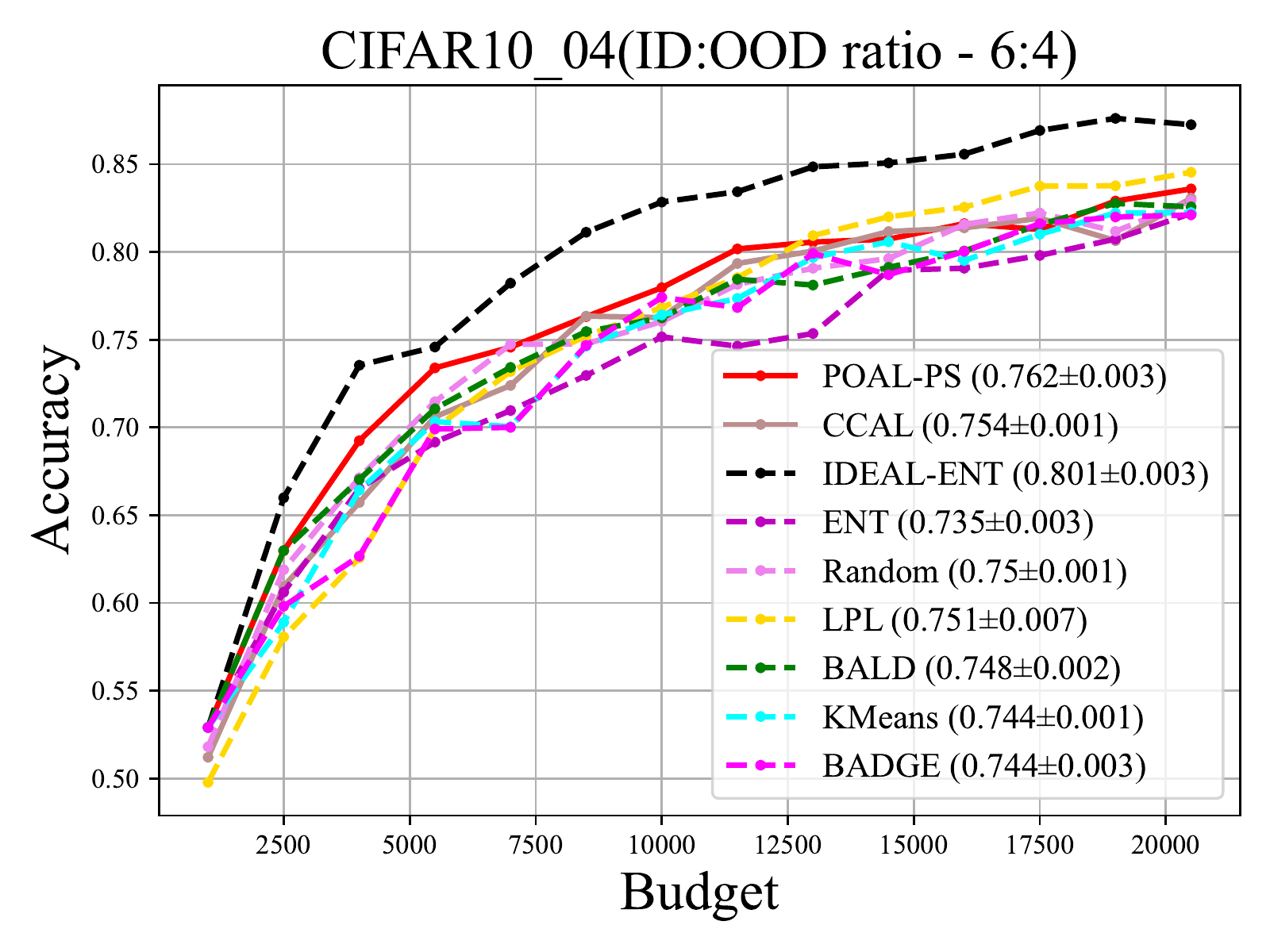}}
\subfloat[Accuracy vs. Budget]{\includegraphics[width=0.5\linewidth]{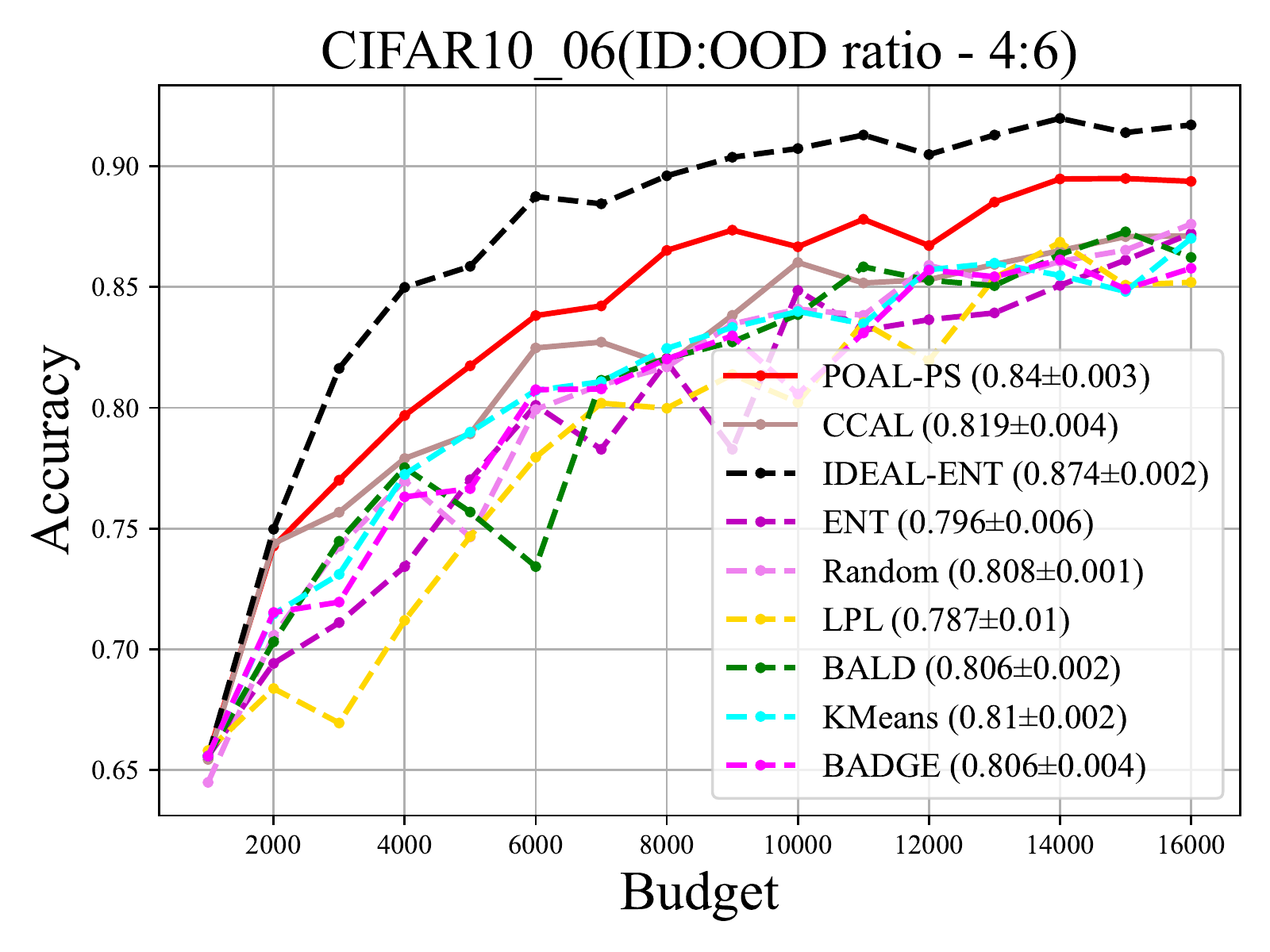}}

\subfloat[OOD samples vs. Budget]{\includegraphics[width=0.5\linewidth]{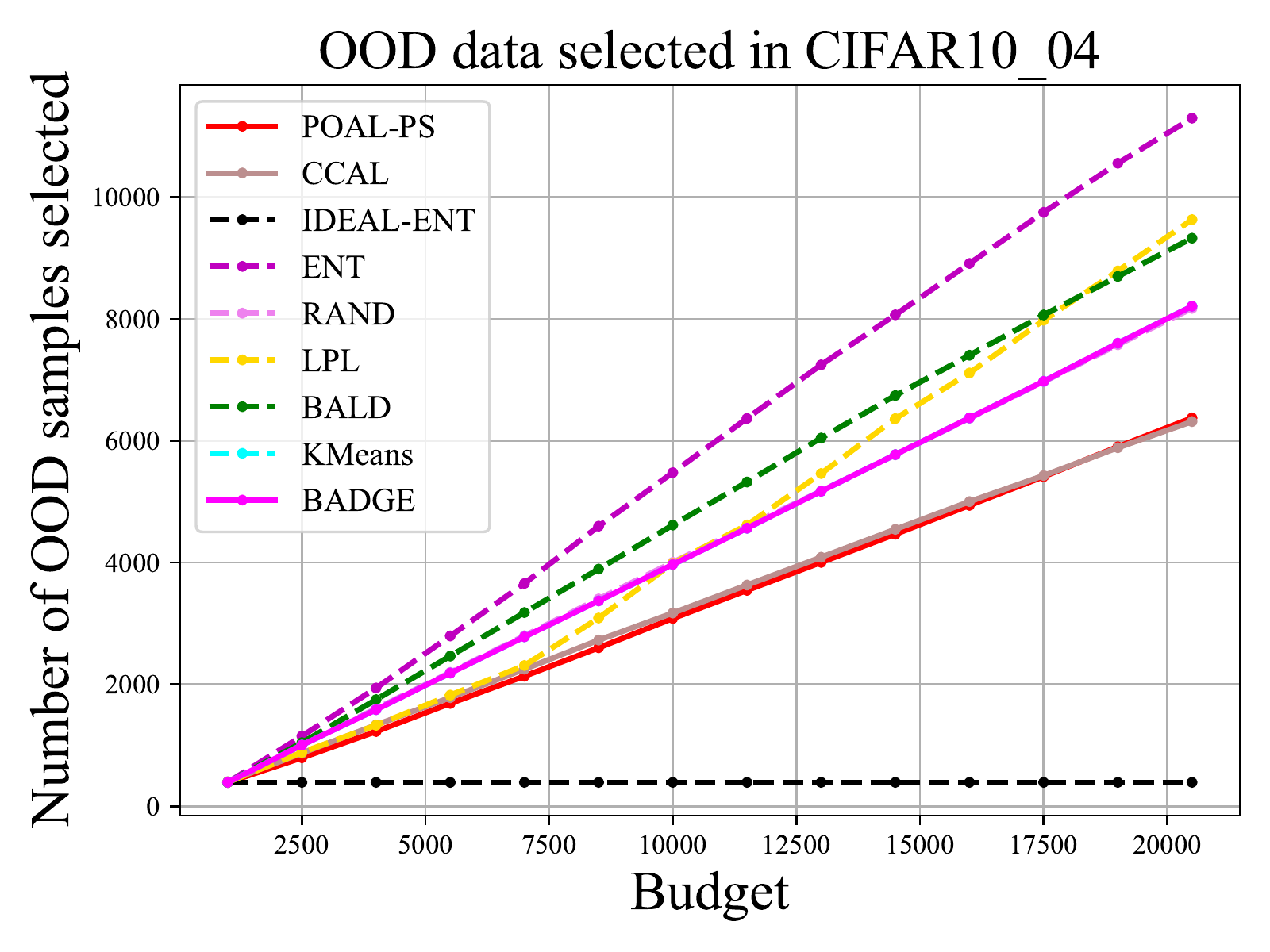}}
\subfloat[OOD samples vs. Budget]{\includegraphics[width=0.5\linewidth]{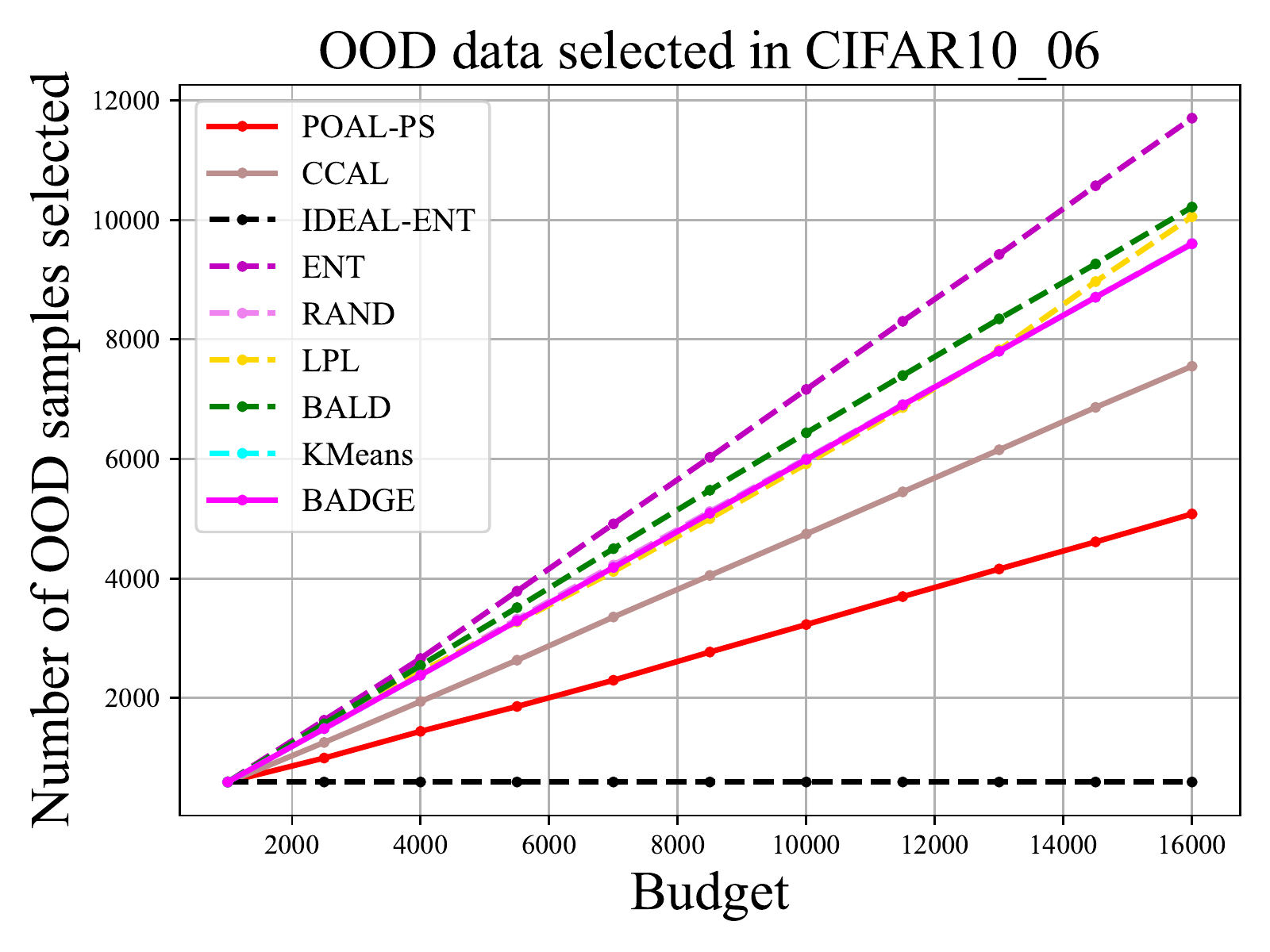}}
\caption{The comparative experiments between our \textbf{POAL-PS} and \textbf{CCAL} on \emph{CIFAR10} dataset (also include \textbf{BADGE} model for comparison).}
\label{ccal}
\end{figure*}

\begin{figure*} [htb]
\centering
\subfloat[EX8: Accuracy vs. Budget]{\includegraphics[width=0.5\linewidth]{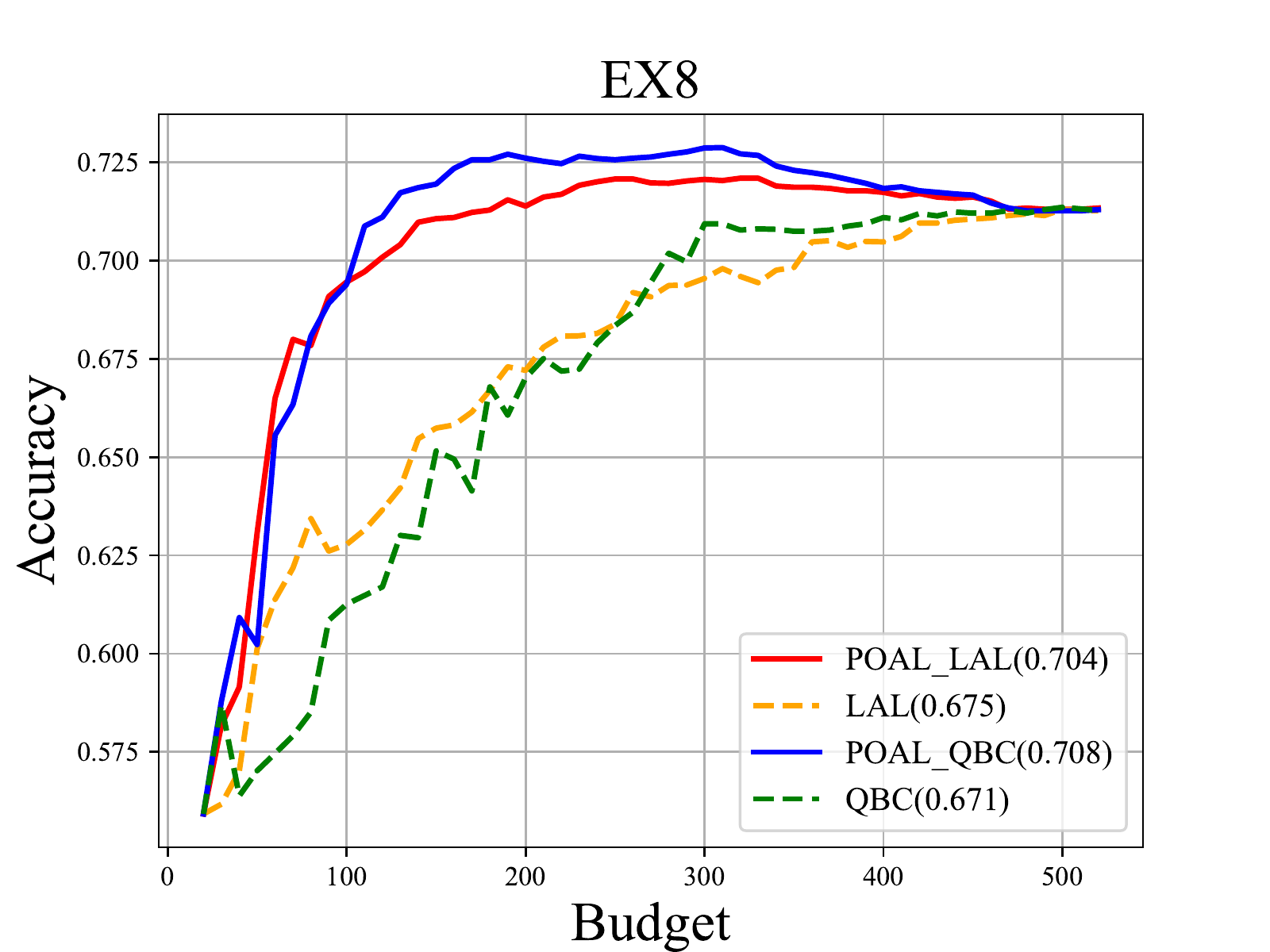}}
\subfloat[Vowel: Accuracy vs. Budget]{\includegraphics[width=0.5\linewidth]{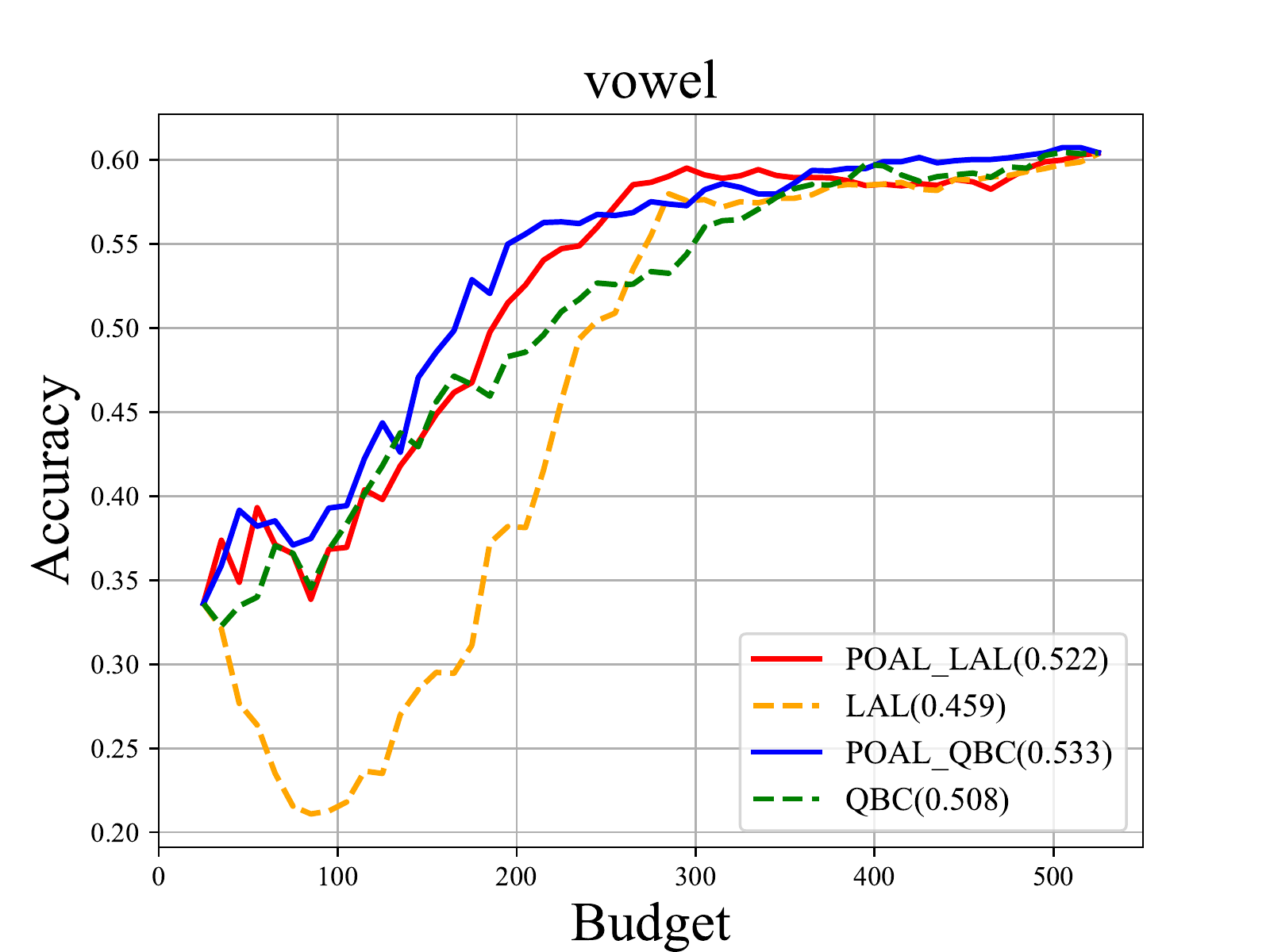}}

\subfloat[EX8: OOD num vs. Budget]{\includegraphics[width=0.5\linewidth]{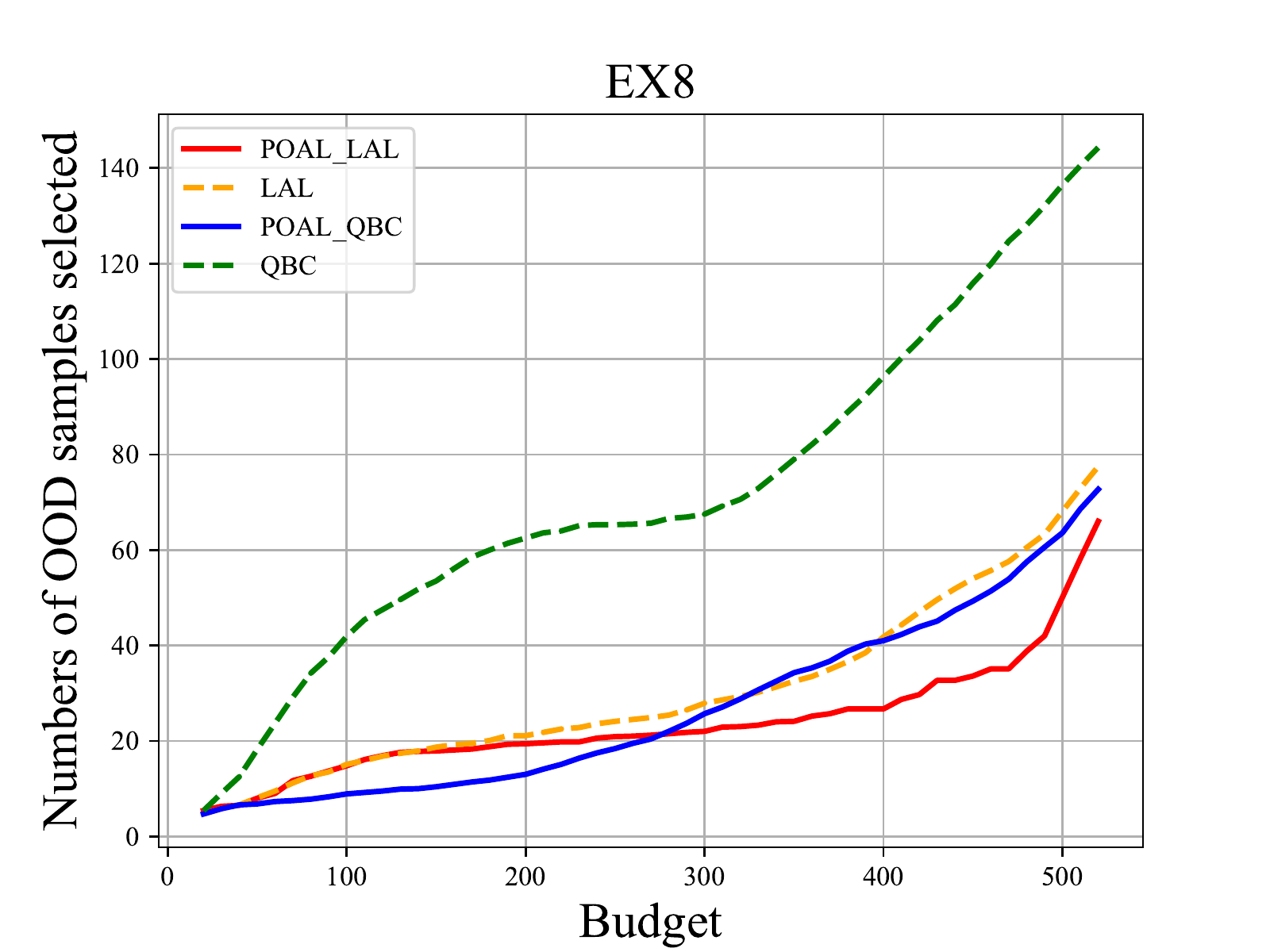}}
\subfloat[Vowel: OOD num vs. Budget]{\includegraphics[width=0.5\linewidth]{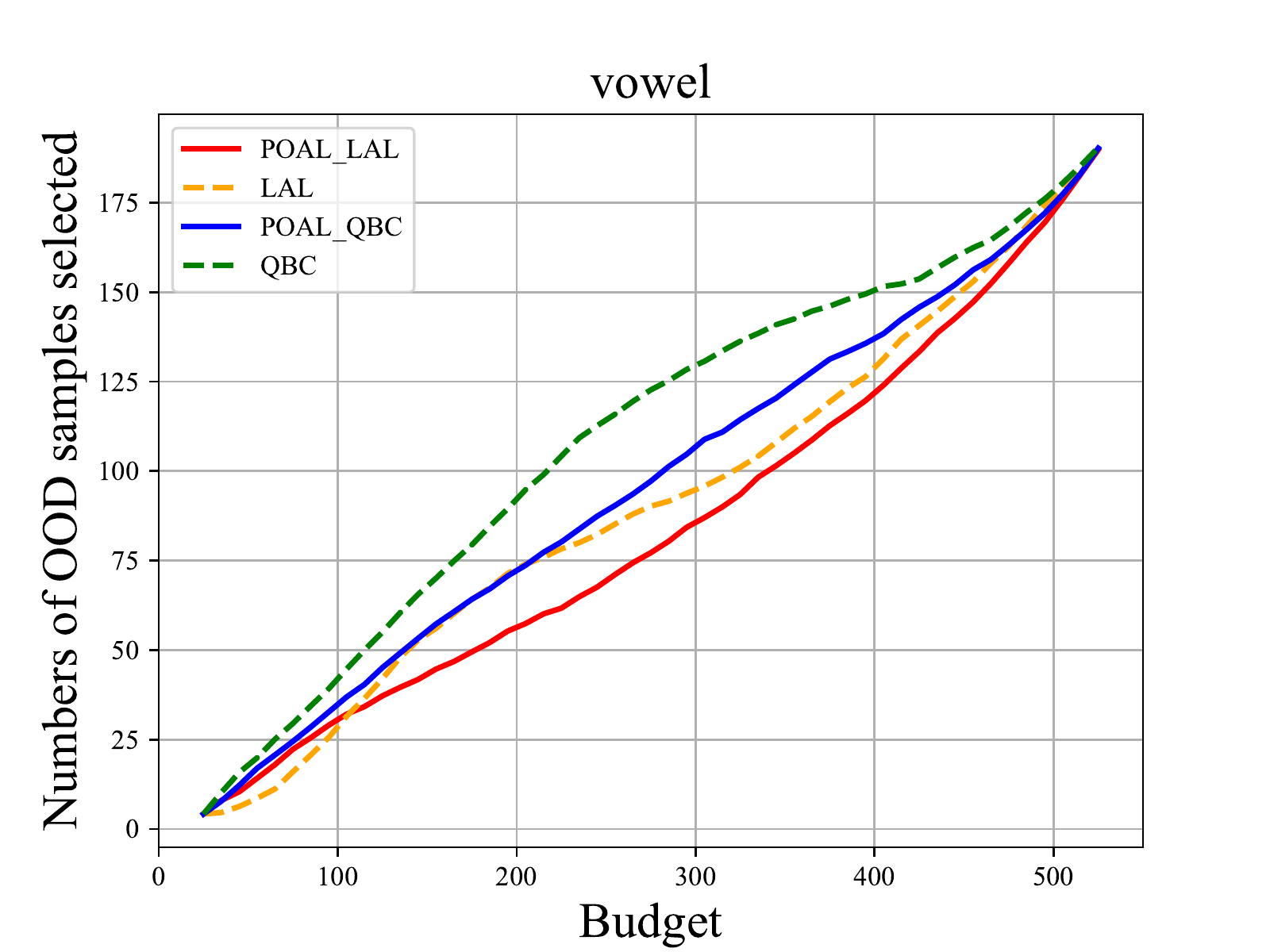}}
\caption{Flexibility: \textbf{POAL} incorporates various AL sampling strategies, including \textbf{LAL} and \textbf{QBC}.}
\label{flexible}
\end{figure*}

\paragraph{POAL vs. CCAL.}
We described the detailed information of \textbf{CCAL} \citep{du2021contrastive} in Section~\ref{related-work-ccal}. We here compare the \textbf{POAL} and \textbf{CCAL} on \emph{CIFAR10-04} and \emph{CIFAR10-06} datasets. Together with the same experimental design in our experiments. We utilize the open-source code implementation of \textbf{CCAL}\footnote{\url{https://github.com/RUC-DWBI-ML/CCAL/}}. We train SimCLR \citep{chen2020simple}, the semantic/distinctive feature extractor, which is provided by \textbf{CCAL}'s source code. We train the two feature extractors with 700 epochs and batch size of 32 on a single V100 GPU. The comparative results are shown in Figure~\ref{ccal}. We have moderately  better performance than \textbf{CCAL} on \emph{CIFAR10-04} dataset, i.e., our \textbf{POAL} has 0.762 AUBC performance and \textbf{CCAL} is 0.754. Note that in Figure~\ref{ccal}-c, \textbf{POAL} has similar efficiency with \textbf{CCAL} on preventing selecting OOD data samples on \emph{CIFAR10-04} dataset. In \emph{CIFAR10-06} task, as the OOD ratio increases, our \textbf{POAL} outperforms \textbf{CCAL}, i.e., our \textbf{POAL} has 0.84 AUBC performance while \textbf{CCAL} is only 0.819. And \textbf{POAL} selects less OOD samples than \textbf{CCAL}, as shown in Figure~\ref{ccal}-d.
Both \textbf{POAL} and \textbf{CCAL} perform better than normal AL sampling schemes (i.e., \textbf{ENT}, \textbf{LPL}, \textbf{BALD}, \textbf{KMeans}, \textbf{BADGE} and \textbf{Random}) on both \emph{CIFAR10-04} and \emph{CIFAR10-06} datasets.

\paragraph{POAL vs. SIMILAR.}
Since \textbf{SIMILAR} (SCMI - FLCMI as acquisition function, see Section~\ref{related-work-ccal}) is both time and memory consuming on large scale datasets, it could not run experiments with the same settings as our experiments. In the original paper \citep{kothawade2021similar}, they down-sample \emph{CIFAR10} training set from $50,000$ to $15,600$, they set class 0 to 8 as ID classes and the 9 to 10 as OOD classes. They first randomly selected $200 \times 8 = 1,600$ ID data samples as initial labeled set, then they randomly selected $500 \times 8 = 4,000$ ID samples and together with the remaining $5,000 \times 2 = 10,000$ OOD samples as unlabeled data pool (size is $4,000 + 10,000 = 14, 000$). We followed the implementation details provided in \citep{kothawade2021similar} and the open-source code\footnote{\url{https://github.com/decile-team/distil}}. The implementation details are: We use ResNet18 \citep{he2016deep} with the same implementations as them in our experiments. Additionally we use a SGD optimizer with an initial learning rate of 0.01, the momentum of 0.9, and a weight decay of $5e-4$. The learning rate is decayed using cosine annealing for each epoch. The batch size is 250, the total budget is $2250$. \citep{kothawade2021similar}. However, we cannot acquire the random seed and times of repeated experiments, thus we don't know how they split/down-sampled the dataset. We can not completely reproduce the original experimental environment in \citep{kothawade2021similar}. In our experiments, we set the random seed as 4666 and repeat the experiment 3 times. We use the source code of FLCMI submodular function\footnote{\url{https://github.com/decile-team/distil/blob/main/distil/active_learning_strategies/scmi.py}}. 

\begin{figure*} [htb]
\centering
\subfloat[Accuracy vs. Budget]{\includegraphics[width=0.5\linewidth]{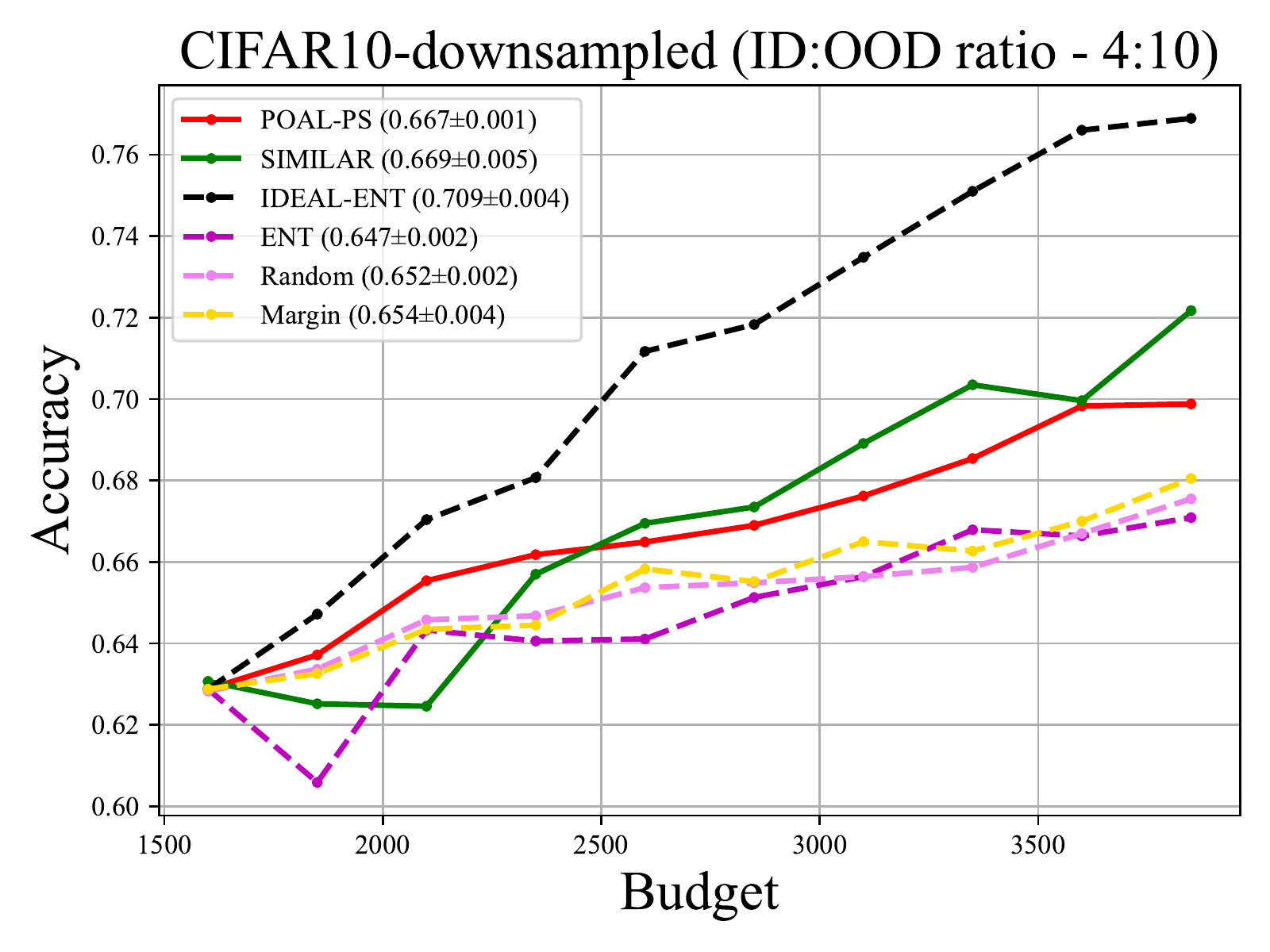}}
\subfloat[OOD num vs. Budget]{\includegraphics[width=0.5\linewidth]{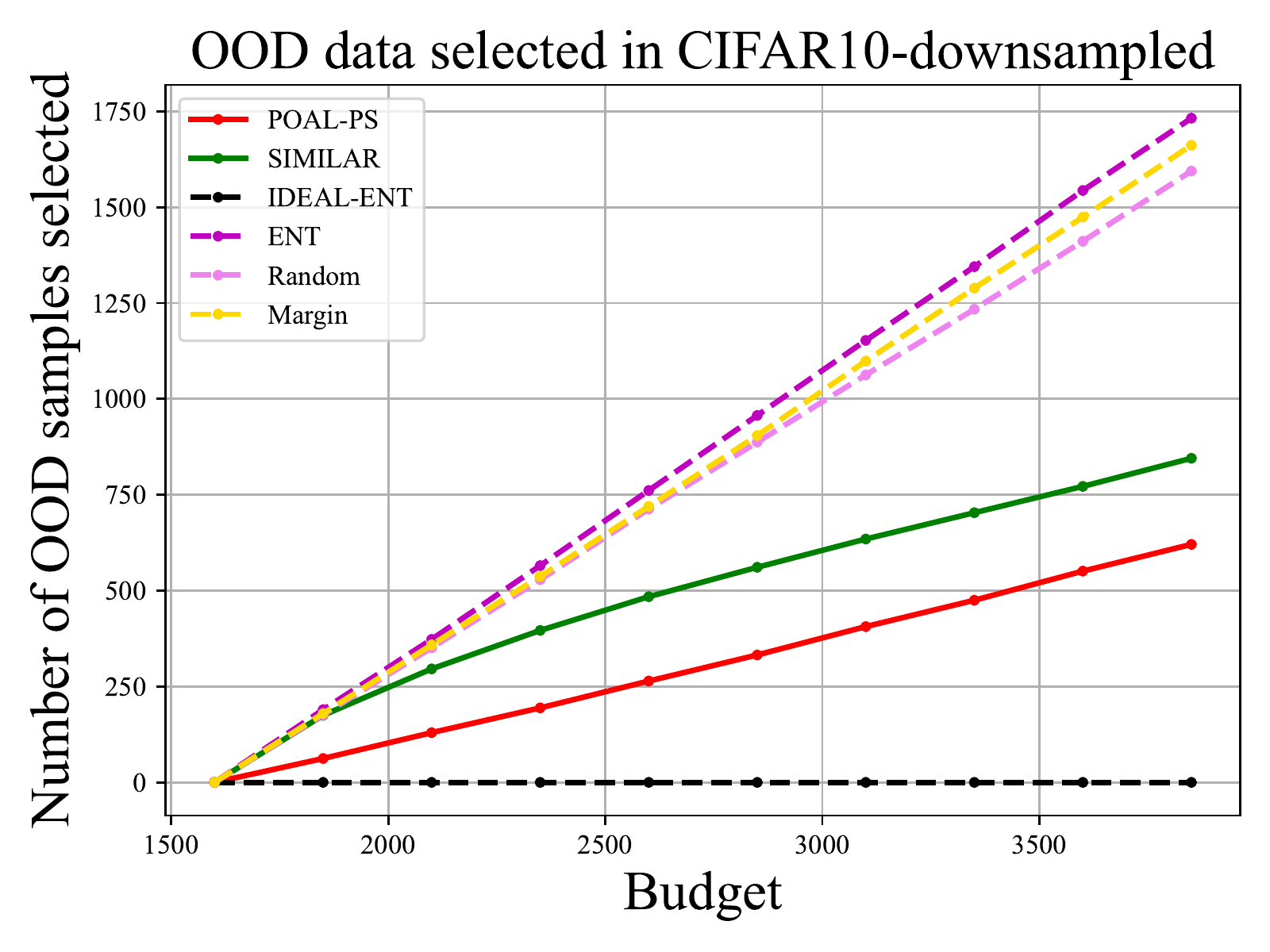}}
\caption{The comparative experiments between our model and \textbf{SIMILAR} on down-sampled \emph{CIFAR10} dataset.}
\label{sim}
\end{figure*}

The comparison experiments are shown in Figure~\ref{sim}. Besides our \textbf{POAL} and \textbf{SIMILAR}, we also provide more baselines as reference, i.e., \textbf{IDEAL-ENT}, \textbf{ENT}, \textbf{Margin} \citep{wang2014new} and \textbf{Random}. As shown in Figure~\ref{sim}, both \textbf{SIMILAR} and \textbf{POAL} have better performance than normal AL sampling strategies. From the aspect AUBC evaluation metric, our model has comparable performance with \textbf{SIMILAR}, 0.667 vs 0.669. But we have lower standard deviation value than \textbf{SIMILAR}, that is, our \textbf{POAL} is more stable. From the aspect of Accuracy vs. Budget curves,  in early stages (e.g., Budget < 2,500), our \textbf{POAL} is better than \textbf{SIMILAR} and in latter stages \textbf{SIMILAR} exceeds \textbf{POAL}. The reason is, in \textbf{SIMILAR}, they calculate the similarities between ID labeled set and unlabeled pool and calculate the dissimilarities between OOD labeled set and unlabeled pool. In early stages, the OOD data is insufficient, thus \textbf{SIMILAR} would select more OOD samples, as shown in Figure~\ref{sim}-b. From this experiment, we find that \textbf{SIMILAR} only performs well when we have enough information of both ID and OOD data samples, which results in more OOD data sample selection. Our \textbf{POAL} only considers the distance between unlabeled samples and ID labeled samples, so we are more efficient on preventing OOD sample selection. Additionally, our \textbf{POAL} is more widely adopted, since we could be adopted on large-scale datasets. But \textbf{SIMILAR} is limited by the computation condition. Our method is also more time efficient than \textbf{SIMILAR}, as shown in Table~\ref{time-sim}. \textbf{SIMILAR} runs five times longer than our \textbf{POAL}.

\begin{table}[htb]
\centering
\small
\caption{The mean and standard deviation of running time (unit: second) the comparative experiments with repeated 3 trials between our \textbf{POAL} and \textbf{SIMILAR}.}
\begin{tabular}{l|c|c|c|c|c|c}
 Method & POAL-PS & SIMILAR & IDEAL-ENT & ENT & Random & Margin \\
 \hline
Time & 6419.0 (109.9) & 32837.7 (897.7) & 2225.0 (12.6) & 1690.0 (15.3) & 1507.7 (5.8) & 1573.7 (9.8) \\
\end{tabular}
\label{time-sim}
\end{table}

\paragraph{Hyper-parameter sensitivity in POAL with early stopping conditions.}
In our main paper and Algorithm~\ref{mcpoal}, we have a hyper-parameter $s_w$ (sliding window size) to control the early stopping condition. Large $s_w$ refers to more strict early stop condition and vice versa. If there is no significant change within $s_w \times p_{\text{inv}}$ iterations/populations, then we can stop early. In our experiments, we set $s_w = 20$. In this section, we conduct an ablation study to see if strict/more strict condition would influence the final performance, as shown in Table~\ref{ablation-study}. We found that $s_w = 10$ (less strict) does not affect the model performance (there is no significant difference of AUBC performance between $s_w = 20$ and $s_w = 10$). Less strict early stopping also requires less running time, e.g., in \emph{CIFAR100-04} dataset, we have 66880.7 seconds average running time with $a_w = 20$ and 4932.3  seconds average running time with $s_w = 10$.

\begin{table}[htb]
\centering
\caption{The hyper-parameter sensitivity of sliding window size $s_w$, including mean and standard deviation (SD) of AUBC performance and running time.}
\begin{tabular}{c|c|c}
\hline
\emph{CIFAR10-04} & AUBC & Time (seconds) \\
\hline
 \textbf{POAL} with $s_w = 20$ & 0.7620 (0.0033) & 62082.7 (7371.9) \\
 \textbf{POAL} with $s_w = 10$ & 0.7613 (0.0024) & 56107.0 (3364.2) \\
\hline
\hline
\emph{CIFAR10-06} & AUBC & Time \\
\hline
 \textbf{POAL} with $s_w = 20$ & 0.8400 (0.0029) & 34946.0 (4860.2) \\
 \textbf{POAL} with $s_w = 10$ & 0.8403 (0.0029) & 47362.0 (188.7) \\
\hline
\hline
\emph{CIFAR100-04} & AUBC & Time \\
\hline
 \textbf{POAL} with $s_w = 20$ & 0.4807 (0.0009) & 66880.7 (1583.5) \\
 \textbf{POAL} with $s_w = 10$ & 0.4797 (0.0017) & 4932.3 (1891.2) \\
\hline
\hline
\emph{CIFAR100-06} & AUBC & Time \\
\hline
 \textbf{POAL} with $s_w = 20$ & 0.5253 (0.0005) & 62762.0 (4277.4) \\
 \textbf{POAL} with $s_w = 10$ & 0.5270 (0.0008) & 55955.3 (5162.4) \\
\hline
\end{tabular}
\label{ablation-study}
\end{table}

\paragraph{Summarizing: comparison of running time and AUBC performance across all models and DL tasks.}
To observe the efficiency of each AL sampling strategies in our experiments, we summarized the overall performance across all models and datasets we adopted in our experiments, including the mean and standard deviation (SD) of AUBC performance and running time. We record the running time from the start of the AL process to the output of the final basic learner. From Table~\ref{overall}, we could observed that our model outperforms all baselines (except for ideal model -- \textbf{IDEAL-ENT}) in terms of AUBC performance. For running time, compared with normal AL sampling strategies, both \textbf{POAL} and \textbf{CCAL} are incomparable. Among \textbf{POAL}, \textbf{CCAL} and \textbf{SIMILAR}, which are designed specific to AL under OOD data scenarios, the running cost of our \textbf{POAL} is affordable. Especially, as mentioned in previous section of hyper-parameter sensitivity, the running timing cost of our \textbf{POAL} can be reduced by loosing the early stop conditions (less $s_w$). 

\begin{table}[htb]
\centering
\caption{Overall comparison across all AL sampling strategies of all DL tasks, including mean and standard deviation (SD) of AUBC performance and running time. $-$ refers to the corresponding experiment is not completed after running many days ($> 9$ days).}
\begin{tabular}{l|ll|ll}
\hline
 & \multicolumn{2}{c|}{\emph{CIFAR10-04}} & \multicolumn{2}{c}{\emph{CIFAR10-06}} \\
 \hline
 & AUBC & Time (seconds)& AUBC & Time (seconds) \\
\hline
\textbf{RAND} & 0.7500 (0.0014) & 10805.3 (148.0) & 0.8080 (0.0008) & 5217.7 (348.0) \\
\textbf{IDEAL-ENT} & 0.8007 (0.0029) & 14578.0 (838.2) & 0.8737 (0.0019) & 9398.0 (696.4) \\
\hdashline
\textbf{POAL} ($s_w = 20$) & \textbf{0.7620 (0.0033)} & 62082.7 (7371.9) & 0.8400 (0.0029) & 34946.0 (4860.2) \\
\textbf{POAL} ($s_w = 10$) & 0.7613 (0.0024) & 56107.0 (3364.2) & \textbf{0.8403 (0.0029)} & 47362.0 (188.7) \\
\textbf{CCAL} & 0.7543 (0.0009) & 76129.3 (5853.7) & 0.8190 (0.0037) & 35715.3 (328.5) \\
\textbf{ENT} & 0.7350 (0.0029) & 10155.3 (872.3) & 0.7960 (0.0057) & 5485.7 (655.4) \\
\textbf{LPL} & 0.7510 (0.0071) & 12384.7 (1555.9) & 0.7873 (0.0103) & 3783.3 (326.9) \\
\textbf{BADGE} & 0.7437 (0.0029) & 41646.0 (12723.5) & 0.8063 (0.0037) & 17539.3 (628.0) \\
\textbf{KMeans} & 0.7440 (0.0014) & 18785.7 (2571.3) & 0.8100 (0.0022) & 8627.3 (407.2) \\
\textbf{BALD} & 0.7480 (0.0022) & 10478.3 (679.7) & 0.8060 (0.0022) & 4158.3 (431.1) \\
\textbf{TwoStage} & 0.7300 (0.0029) & 8342.3 (1059.4) & 0.7970 (0.0028) & 5524.7 (821.3) \\
\textbf{WeightedSum-1.0} & 0.7337 (0.0019) & 9489.7 (454.5) & 0.8033 (0.0033) & 2006.3 (13.8) \\
\textbf{WeightedSum-5.0} & 0.7327 (0.0029) & 11892.3 (1733.9) & 0.8060 (0.0045) & 7998.3 (81.6) \\
\textbf{WeightedSum-0.2} & 0.7340 (0.0045) & 8249.3 (2019.2) & 0.8063 (0.0012) & 5751.0 (795.3) \\
\hline
\hline
 & \multicolumn{2}{c|}{\emph{CIFAR100-04}} & \multicolumn{2}{c}{\emph{CIFAR100-06}} \\
 \hline
 & AUBC & Time (seconds) & AUBC & Time (seconds) \\
\hline
\textbf{RAND} & 0.4560 (0.0016) & 11563.3 (985.5) & 0.4453 (0.0026) & 11075.3 (1198.2) \\
\textbf{IDEAL-ENT} & 0.5250 (0.0008) & 17736.7 (1694.7) & 0.5707 (0.0017) & 19098.0 (1458.8) \\
\hdashline
\textbf{POAL} ($s_w = 20$) & \textbf{0.4807 (0.0009)} & 66880.7 (1583.5) & 0.5253 (0.0005) & 62762.0 (4277.4) \\
\textbf{POAL} ($s_w = 10$) & 0.4797 (0.0017) & 4932.3 (1891.2) & \textbf{0.5270 (0.0008)} & 55955.3 (5162.4) \\
\textbf{CCAL} & $0.4400 (0.0008)$ & $21253.0 (2171.9)$ & $0.4467 (0.00244)$ & $65303 (11676.6)$  \\
\textbf{ENT} & 0.4267 (0.0034) & 11202.3 (2484.7) & 0.4100 (0.0036) & 10940.0 (1318.4) \\
\textbf{LPL} & 0.4140 (0.0037) & 17797.0 (4093.8) & 0.4087 (0.0042) & 5633.0 (252.3) \\
\textbf{BADGE} & 0.4530 (0.0000) & 58877.7 (13876.3) & 0.4430 (0.0008) & 58046.3 (17296.9) \\
\textbf{KMeans} & 0.4527 (0.0005) & 34944.3 (4121.3) & 0.4413 (0.0021) & 17465.0 (3570.3) \\
\textbf{BALD} & 0.4467 (0.0040) & 14699.0 (2675.9) & 0.4313 (0.0061) & 4574.3 (140.5) \\
\textbf{TwoStage} & 0.4347 (0.0021) & 11484.3 (4360.1) & 0.4137 (0.0017) & 3234.3 (91.1) \\
\textbf{WeightedSum-1.0} & 0.4267 (0.0025) & 12722.0 (4164.8) & 0.4103 (0.0005) & 9455.3 (468.5) \\
\textbf{WeightedSum-5.0} & 0.4287 (0.0039) & 14822.3 (6249.7) & 0.4073 (0.0050) & 11309.3 (1303.4) \\
\textbf{WeightedSum-0.2} & 0.4290 (0.0022) & 11736.7 (4394.3) & 0.4097 (0.0033) & 8865.7 (1230.1) \\
\hline
\end{tabular}
\label{overall}
\end{table}

\subsection{Related Work}
\subsubsection{Comparison between POAL and CCAL, SIMILAR}
\label{related-work-ccal}
In this section, we discussed the related work that involved AL under OOD data scenarios. As we mentioned in main paper, there is little related work of AL under OOD data scenarios. To the best of our knowledge, there are two published paper/work that discussed AL under OOD data scenarios: Contrastive Coding for Active Learning (\textbf{CCAL}) \citep{du2021contrastive} and Submodular Information Measures Based Active Learning In Realistic Scenarios (\textbf{SIMILAR}) \citep{kothawade2021similar}. We would introduce these two papers in detail and compared them with our work.

\citet{du2021contrastive} concerns class distribution mismatch problem in AL, their goal is to select most informative samples with matched categories. \textbf{CCAL} is the first work related to AL for class distribution mismatch, it proposed a contrastive coding based method, which extracts semantic and distinctive features by contrastive learning. Semantic feature refers to category-level features, that can be exploited to filter invalid samples with mismatched categories. Distinctive features describe individual level. It is a AL task-specific feature, and used to select the most representative and informative class-matched unlabeled data samples. CCAL achieves good performance on well-studied datasets, e.g., CIFAR10 and CIFAR100 datasets. \textbf{CCAL} utilized self-supervised model like SimCLR and CSI, it is a very good idea since it well extracts semantic and distinctive features, compared with normal feature representations like the output of penultimate layer of neural networks. However, it also brings limitations, that training a self-supervised model has both high computational and timing cost, especially on large-scale data sets. Based on the release code of \textbf{CCAL}\footnote{\url{https://github.com/RUC-DWBI-ML/CCAL}}, we take 3 days to train a distinctive feature extraction model (with 700 epochs and batch size 32, on CIFAR10 dataset) on a V100 GPU. Additionally, \textbf{CCAL} adopted weighted-sum optimization to combine semantic and distinctive scores, whose acquisiton function is as follows:
\begin{equation*}
    \alpha_{\textbf{CCAL}} = \tanh{(kS_{\text{semantic}(\mathbf{x}) - t})} + S_{\text{distinctive}}(\mathbf{x}),
\end{equation*}
where the threshold $t$ is for narrowing the semantic scores of samples selectively, and $k$ controls the slope of the tanh function. $\alpha_{\textbf{CCAL}}$ brings two hyper-parameters for balancing semantic score and distinctive score. As discussed in Section 4.3 in \citep{du2021contrastive}, the choice of the two hyper-parameters would influence the final performance. The final shortcoming comes from the calculation of distinctive score, to obtain representativeness information. \textbf{CCAL} needs pair-wise comparison among the whole data pool. It is very time and memory consuming on large-scale data sets. 

\textbf{SIMILAR} is an active learning framework using previously proposed submodular information measures (SIM) as acquisition functions. Apparently, \citep{kothawade2021similar} is an application of SIM \citep{iyer2021submodular, kaushal2021prism}. \citet{kothawade2021similar} proposed many variants of SIM to deal with various realistic scenarios, e.g., data imbalance, rare-class, OOD, etc. AL under OOD data scenario is only a sub-task of their work. They adopted a submodular conditional mutual information (SCMI) function that best matches AL under OOD data scenario tasks, that is, Facility Location Conditional Mutual Information (FLCMI), and the function is:
\begin{equation*}
    \mathcal{I}_f(A, Q|P) = \sum_{i \in V} \max(\min(\max_{j \in A} s_{ij}, \eta\max_{j \in Q}s_{ij}) - \nu\max_{j \in P}s_{ij},0), 
\end{equation*}
where they use the currently labeled OOD points as the conditioning set $P$, and the currently labeled in-distribution (ID) points as the query set $Q$. $A$ is unlabeled data set. It jointly models the similarity between $A$ and $Q$ and their dissimilarity with $P$. The advantages and disadvantages of \textbf{SIMILAR} are both clear. The highlight of \textbf{SIMILAR} is, the proposed methods have strong theoretical supports \citep{iyer2021submodular, kaushal2021prism}. However, the proposed FLCMI is both time and memory consuming, it could not handle large-scale datasets. For instance, we cannot run experiments on full CIFAR10 datset with one V100 GPU with 32GB memory, it would report memory error. Actually, in \citep{kothawade2021similar}, they conduct OOD related experiments on downsampled CIFAR10 dataset, they downsample the dataset to size $15.6K$ (the size of whole CIFAR10 dataset for training is $50K$).

To sum up, the existing researches that concern AL under OOD data scenarios remain the following problems to be overcome: 1) time/memory cost on large-scale datasets (both appear in \textbf{CCAL} and \textbf{SIMILAR}); 2) additional trade-off hyper-parameters for balancing different (even conflict) criteria need hand tuning or tuned by extra validation set. To address these issues, we proposed our \textbf{POAL} framework, we solved the first problem, to save time/memory cost by pre-selection and early stopping techniques, we are able to reduce the searching space to meet any researcher's computational conditions/requirements. We solved the trade-off hyper-parameter problem by applying pareto optimization for self-adaptive balancing various (even conflict) objectives. Therefore, our proposed framework can more easily be adapted to various tasks.

The empirical comparison between our \textbf{POAL} and \textbf{CCAL}, \textbf{SIMILAR} are in previous section~\ref{comp-ccal-similar}.

\subsubsection{Differences between POAL and POSS}

In this part, we focus on the differences between our proposed \textbf{POAL} and Pareto Optimization for Subset Selection (\textbf{POSS}) \citep{qian2015subset}. \textbf{POSS} is originally applied to the subset selection problem, which is defined as follows:
\begin{equation*}
    \mathop{\arg\min}_{S \subseteq V} f(S) \quad s.t. \quad |S| \le k
\end{equation*}
where $V=\{X_{1},...,X_{n}\}$ is a set of variables, $f$ is a criterion function and $k$ is a positive integer. The subset selection problem aims to select a subset $S \subseteq V$ such that $f$ is optimized with the constraint $|S| \le k$, where $|\cdot|$ denotes the size of a set. The subset selection problem is NP-hard in general \citep{davis1997adaptive}.

\textbf{POSS} solves the subset selection by separating the problem into two objectives, namely optimizing the criterion function and minimizing the subset size. Usually the two objectives are conflicting. Thus the problem is transformed to a multi-objective optimization problem:
\begin{equation*}
    \min_{S \subseteq V} \quad (f(S), |S|)
\end{equation*}
Due to different trade-offs of objectives, multi-objective optimization algorithms need to find a Pareto set $\mathcal{P}$ containing Pareto optimal solutions. Specifically, \textbf{POSS} first initializes $\mathcal{P}$ with a random solution, and then selects a solution from $\mathcal{P}$ to generate a new solution by flipping each bit with probability $1/n$ ($n$ is the number of variables). The new solution will be added it to $\mathcal{P}$ if it is not strictly dominated by any solution in $\mathcal{P}$, and the solutions in $\mathcal{P}$ that are dominated by the new solution will also be excluded. \textbf{POSS} repeats this iteration for $T$ times, and is proved to achieve the optimal approximation guarantee of $(1-1/e)$ with $E[T]\le 2ek^{2}n$ expected running time.

Our \textbf{POAL} differs from POSS in three aspects. Firstly, although POSS is in the form of bi-objective optimization, it actually supports only one criterion function. But our  \textbf{POAL} needs to solve two different criterion functions, which is a much harder problem in fact. Secondly, POSS solves the problem with the constraint $|S| \le k$, while our \textbf{POAL} needs fixed-size subset solutions, which means the search space is different. The last difference lies in the generation of new solutions in the algorithm. \textbf{POSS} generates new solutions by flipping the bits of the solutions from the previous iterations. This operation may change the subset size, violating our setting of fixed-size solutions. So we adopt Monte-Carlo approach to generate fixed-size solutions that are not related to the solutions from the previous iterations.

Given these differences, we cannot utilize the theoretical proof of $T$ in \textbf{POSS}. Nevertheless, we noticed in previous research \citep{qian2017optimizing} that the Pareto set empirically converges much faster than $E[T]$, i.e., see Figure 2 in \citep{qian2017optimizing}. Therefore, we propose an \emph{early-stopping} technique to terminate our Monte-Carlo \textbf{POAL} algorithm.

\subsection{Limitations and negative societal impacts}
\paragraph{Limitations.}
As mentioned in Section 5 in main paper, the limitations of our work comes from two aspects: (1) the initiations of Mahalanobis distance based ID confidence score calculation; and (2) the Monte-Carlo sampling scheme in our \textbf{POAL} for seeking non-dominated fixed-sized subset solutions is not high-efficiency. In future work we will try more suitable feature representations that are able to construct more distinct ID-/OOD- data distributions, more suitable feature representations like semantic and distinctive feature representations in \citep{du2021contrastive}. For the concerns of the efficiency of Monte-Carlo sampling scheme, although many sampling schemes like (adaptive) importance sampling and metropolis-hastings sampling would be more efficient than Monte-Carlo sampling, however, these methods might suffer from initialization problem. Monte-Carlo sampling has no such problem, In future work, we would try to propose more efficient sampling methods for finding non-dominated subset solutions.

\paragraph{Societal impacts.}
The negative societal impacts of our work include using \textbf{POAL} with pre-selecting strategies (\textbf{POAL-PS}) on large-scale datasets would induce bias problems. Since Pre-Selection technique would firstly filter some unlabeled data samples in one AL iteration, and thus have no probability to be selected. Our work also has positive societal impacts. Our framework can be easily adopted to various data scenarios, due to its simplicity and compatibility. We have no limitation to applied downstream tasks. Our framework is flexible, it is able to incorporate various AL sampling strategies, various OOD detection methods, different scales of target data and various personal requirements (e.g., maximum budget, batch size).

\end{document}